\documentclass{article}
  
\usepackage{comment,url,algorithm,algorithmic,graphicx,subcaption,relsize}
\usepackage{amssymb,amsfonts,amsmath,amsthm,amscd,dsfont,mathrsfs,mathtools,nicefrac}
\usepackage{float,psfrag,epsfig,color,xcolor,url,hyperref}
\usepackage{epstopdf,bbm,mathtools,enumitem}
\usepackage[toc,page]{appendix}
\usepackage[mathscr]{euscript}

\usepackage[top=1in, bottom=1in, left=1in, right=1in]{geometry}

\def\ddefloop#1{\ifx\ddefloop#1\else\ddef{#1}\expandafter\ddefloop\fi}

\def\ddef#1{\expandafter\def\csname bb#1\endcsname{\ensuremath{\mathbb{#1}}}}
\ddefloop ABCDEFGHIJKLMNOPQRSTUVWXYZ\ddefloop

\def\ddef#1{\expandafter\def\csname c#1\endcsname{\ensuremath{\mathcal{#1}}}}
\ddefloop ABCDEFGHIJKLMNOPQRSTUVWXYZ\ddefloop

\def\ddef#1{\expandafter\def\csname v#1\endcsname{\ensuremath{\boldsymbol{#1}}}}
\ddefloop ABCDEFGHIJKLMNOPQRSTUVWXYZabcdefghijklmnopqrstuvwxyz\ddefloop

\def\ddef#1{\expandafter\def\csname v#1\endcsname{\ensuremath{\boldsymbol{\csname #1\endcsname}}}}
\ddefloop {alpha}{beta}{gamma}{delta}{epsilon}{varepsilon}{zeta}{eta}{theta}{vartheta}{iota}{kappa}{lambda}{mu}{nu}{xi}{pi}{varpi}{rho}{varrho}{sigma}{varsigma}{tau}{upsilon}{phi}{varphi}{chi}{psi}{omega}{Gamma}{Delta}{Theta}{Lambda}{Xi}{Pi}{Sigma}{varSigma}{Upsilon}{Phi}{Psi}{Omega}{ell}\ddefloop

\def\balign#1\ealign{\begin{align}#1\end{align}}
\def\baligns#1\ealigns{\begin{align*}#1\end{align*}}
\def\balignat#1\ealign{\begin{alignat}#1\end{alignat}}
\def\balignats#1\ealigns{\begin{alignat*}#1\end{alignat*}}
\def\bitemize#1\eitemize{\begin{itemize}#1\end{itemize}}
\def\benumerate#1\eenumerate{\begin{enumerate}#1\end{enumerate}}

\newenvironment{talign*}
 {\csname align*\endcsname}
 {\endalign}
\newenvironment{talign}
 {\csname align\endcsname}
 {\endalign}

\def\balignst#1\ealignst{\begin{talign*}#1\end{talign*}}
\def\balignt#1\ealignt{\begin{talign}#1\end{talign}}

\let\originalleft\left
\let\originalright\right
\renewcommand{\left}{\mathopen{}\mathclose\bgroup\originalleft}
\renewcommand{\right}{\aftergroup\egroup\originalright}

\def\tinycitep*#1{{\tiny\citep*{#1}}}
\def\tinycitealt*#1{{\tiny\citealt*{#1}}}
\def\tinycite*#1{{\tiny\cite*{#1}}}
\def\smallcitep*#1{{\scriptsize\citep*{#1}}}
\def\smallcitealt*#1{{\scriptsize\citealt*{#1}}}
\def\smallcite*#1{{\scriptsize\cite*{#1}}}

\def\mbb#1{\mathbb{#1}}

\theoremstyle{plain}  
\newtheorem*{remark}{\textbf{Remark}}

\def\<{\left\langle} 
\def\>{\right\rangle}

\def\E{\mbb{E}} 

\DeclareSymbolFont{rsfs}{U}{rsfs}{m}{n}
\DeclareSymbolFontAlphabet{\mathscrsfs}{rsfs}

\providecommand{\argmin}{\mathop\mathrm{arg min}}

\ifdefined\nonewproofenvironments\else
\ifdefined\ispres\else
\newtheorem{theo}{Theorem}

\newtheorem{coro}[theo]{Corollary}
\newtheorem{defi}[theo]{Definition}

\newtheorem{prop}[theo]{Proposition}

\newenvironment{proof-sketch}{\noindent\textbf{Proof Sketch}
  \hspace*{1em}}{\qed\bigskip\\}
\newenvironment{proof-idea}{\noindent\textbf{Proof Idea}
  \hspace*{1em}}{\qed\bigskip\\}
\newenvironment{proof-of-lemma}[1][{}]{\noindent\textbf{Proof of Lemma {#1}}
  \hspace*{1em}}{\qed\\}
\newenvironment{proof-of-theorem}[1][{}]{\noindent\textbf{Proof of Theorem {#1}}
  \hspace*{1em}}{\qed\\}
\newenvironment{proof-attempt}{\noindent\textbf{Proof Attempt}
  \hspace*{1em}}{\qed\bigskip\\}

\fi

\newtheorem{assumption}{Assumption}

\fi
\makeatletter
\@addtoreset{equation}{section}
\makeatother

\hypersetup{
  colorlinks,
  linkcolor={red!50!black},
  citecolor={blue!50!black},
  urlcolor={blue!80!black}
}

\renewcommand{\Pr}[1]{\mathbb{P}\left( #1 \right)}

\usepackage{caption}
\usepackage{authblk}

\usepackage{wrapfig}
\usepackage{cancel}

\newcommand{\hvb}{\hat{\boldsymbol{\beta}}}

\newcommand\trace[1]{\operatorname{tr}\left(#1\right)} 
\newcommand\diag[1]{\operatorname{diag}\left(#1\right)}

\newcommand\dif{\operatorname{d}}

\newcommand\opt{\text{opt}}

\def\rv{\text{R}_\text{v}}
\def\rb{\text{R}_\text{b}}

\def\ipe{\stackrel{\text{p}}{\rightarrow}}
\def\ase{\stackrel{\text{a.s.}}{\rightarrow}}
\def\aseq{\stackrel{\text{a.s.}}{=}}
\def\n{\nonumber}

\renewcommand\t{{\ensuremath{\scriptscriptstyle{\top}}}}

\title{On the Optimal Weighted $\ell_2$ Regularization in\\ Overparameterized Linear Regression}

\author[1]{\thanks{Equal contribution; alphabetical ordering. Email: \texttt{dennywu@cs.toronto.edu}, \texttt{jixu@cs.columbia.edu}.}Denny Wu}
\author[2]{Ji Xu}

\affil[1]{University of Toronto and Vector Institute.}
\affil[2]{Columbia University.}

\begin{document}

\maketitle

\vspace{-0.6cm}
\begin{abstract}
We consider the linear model $\vy=\vX\vbeta_{\star}+\vepsilon$ with $\vX\in \mathbb{R}^{n\times p}$ in the overparameterized regime $p>n$. We estimate $\vbeta_{\star}$ via generalized (weighted) ridge regression:  $\hat{\vbeta}_{\lambda}=\left(\vX^{\t}\vX+\lambda\vSigma_w\right)^{\dagger}\vX^{\t}\vy$, where $\vSigma_w$ is the weighting matrix. 
Under a random design setting with general data covariance $\vSigma_x$ and anisotropic prior on the true coefficients  $\bbE\vbeta_{\star}\vbeta_{\star}^{\t}=\vSigma_\beta$, we provide an exact characterization of the prediction risk $\mathbb{E}(y-\vx^{\t}\hat{\vbeta}_{\lambda})^2$ in the proportional asymptotic limit $p/n\rightarrow \gamma \in (1,\infty)$. Our general setup leads to a number of interesting findings. We outline precise conditions that decide the sign of the optimal setting $\lambda_{\opt}$ for the ridge parameter $\lambda$ and confirm the implicit $\ell_2$ regularization effect of overparameterization, which theoretically justifies the surprising empirical observation that $\lambda_{\opt}$ can be \textit{negative} in the overparameterized regime. We also characterize the double descent phenomenon for principal component regression (PCR) when $\vX$ and $\vbeta_{\star}$ are both anisotropic. Finally, we determine the optimal weighting matrix $\vSigma_w$ for both the ridgeless ($\lambda\to 0$) and optimally regularized ($\lambda = \lambda_{\opt}$) case, and demonstrate the advantage of the weighted objective over standard ridge regression and PCR.     
\end{abstract}

\section{Introduction}
\label{sec:intro}
In this work we consider learning the target signal $\vbeta_\star$ in the following linear regression model: 
\[
y_i = \vx_i^{\t}\vbeta_{\star} +\epsilon_i,\quad i=1,2,\ldots,n
\]
where each feature vector $\vx_i\in\bbR^{p}$ and noise $\epsilon_i\in\bbR$ are drawn i.i.d.~from the two independent random variables $\tilde{\vx}$ and $\tilde{\epsilon}$ satisfying $\bbE\tilde{\epsilon} = 0$, $\bbE\tilde{\epsilon}^2=\tilde{\sigma}^2$, $\tilde{\vx}=\vSigma_x^{1/2}\vz/\sqrt{n}$, and the components of $\vz$ are i.i.d.~random variables with zero mean, unit variance, and bounded 12th absolute central moment. To estimate $\vbeta_{\star}$ from $(\vx_i,y_i)$, we consider the following 
generalized ridge regression estimator:
\begin{eqnarray}
\hvb_{\lambda}&=&\left(\vX^{\t}\vX+\lambda\vSigma_w\right)^{\dag}\vX^{\t}\vy,\label{eq:problem}
\end{eqnarray}
in which $\vX\in\bbR^{n\times p}$ is the feature matrix, $\vy$ is vector of the observations, $\vSigma_w$ is a positive definite weighting matrix, and the symbol $^\dag$ denotes the Moore-Penrose pseudo-inverse. When $\lambda\geq 0$, $\hvb_{\lambda}$ minimizes the squared loss plus a weighted $\ell_2$ regularization: $\min_{\vbeta} \sum_{i=1}^n(y_i - \vx_i^{\t}\vbeta)^2 + \lambda\vbeta^\t\vSigma_w\vbeta$. Note that $\vSigma_w = \vI_d$ reduces the objective to standard ridge regression.

While the standard ridge regression estimator is relatively well-understood in the data-abundant regime ($n>p$), several interesting properties have been recently discovered in high dimensions, especially when $p>n$. For instance, the double descent phenomenon suggests that overparameterization may not result in overfitting due to the \textit{implicit regularization} of the least squares estimator \cite{hastie2019surprises,bartlett2019benign}. This implicit regularization also relates to the surprising empirical finding that the optimal ridge parameter $\lambda$ can be negative in the overparameterized regime \cite{kobak2020optimal}.

Motivated by the observations above, we characterize the estimator $\hvb_\lambda$ in the proportional asymptotic limit: $p/n\rightarrow \gamma \in (1,\infty)$\footnote{Some of our results also apply to the underparameterized regime ($\gamma<1$), as we explicitly highlight in the sequel.} as $n,p\rightarrow \infty$. We generalize the previous random effects hypothesis and place the following prior on the true coefficients (independent of $\tilde{\vx}$ and $\tilde{\epsilon}$): $\bbE\vbeta_{\star}\vbeta_{\star}^{\t} = \vSigma_{\vbeta}$. Note that this assumption allows us to analyze both \textit{random} and \textit{deterministic} $\vbeta_*$. 
Our goal is to study the prediction risk of $\hvb_{\lambda}$:  $\bbE_{\tilde{x},\tilde{\epsilon},\vbeta_{\star}}(\tilde{y}-\tilde{\vx}^{\t}\hvb_{\lambda})^2$,
where $\tilde{y}=\tilde{\vx}^{\t}\vbeta_{\star}+\tilde{\epsilon}$\footnote{When $\vbeta_{\star}$ is deterministic, $\bbE_{\tilde{x},\tilde{\epsilon},\vbeta_{\star}}(\tilde{y}-\tilde{\vx}^{\t}\hvb_{\lambda})^2$ reduces to the prediction risk for one fixed $\vbeta_{\star}$.
}. Compared to previous high-dimensional analysis of ridge regression \cite{dobriban2018high}, our setup is generalized in two important aspects: 

\textbf{Anisotropic $\vSigma_x$ and $\vSigma_{\beta}$.} 
Our analysis deals with general anisotropic prior $\vSigma_{\beta}$ and data covariance $\vSigma_x$, in contrast to previous works which assume either isotropic features or signal (e.g.,~ \cite{dobriban2018high,hastie2019surprises,xu2019number}). Note that the isotropic assumption on the signal or features implies that each component is roughly of the same magnitude, which may not hold true in practice. 
For instance, it has been theoretically shown that the optimal ridge penalty is always non-negative when either the signal $\vSigma_{\beta}$ \cite[Theorem 2.1]{dobriban2018high} or the features $\vSigma_x$ \cite[Theorem 5]{hastie2019surprises} is isotropic. 
On the other hand, empirical results demonstrated that in the overparameterized regime, the optimal ridge for real-world data can be negative \cite{kobak2020optimal}. While this observation cannot be captured by previous works, our less restrictive assumptions lead to a concise description of when this phenomenon occurs.
 
\textbf{Weighted $\ell_2$ Regularization.}
We consider generalized ridge regression instead of simple isotropic shrinkage. While the generalized formulation has also been studied (e.g.,~\cite{hoerl1970ridge,casella1980minimax}), to the best of our knowledge, no existing work computes the exact risk in the overparameterized proportional asymptotic regime and characterizes the corresponding optimal $\vSigma_w$. Our setting is also inspired by recent observations in deep learning that weighted $\ell_2$ regularization often achieves better generalization compare to isotropic weight decay \cite{loshchilov2017decoupled,zhang2018three}.
Our theoretical analysis illustrates the benefit of weighted $\ell_2$ regularization.
  
Under the general setup~\eqref{eq:problem}, the contributions of this work can be summarized as (see Figure~\ref{fig:illustration}):
\vspace{-0.8mm}
\begin{itemize}[leftmargin=*,topsep=1.2mm]   
    \item \textbf{Exact Asymptotic Risk.} In Section \ref{sec:riskcal} we derive the prediction risk $R(\lambda)$ of our estimator \eqref{eq:problem} in its bias-variance decomposition (see Figure \ref{fig:risk}). We also characterize the risk of principal component regression (PCR) and confirm the double descent phenomenon under more general setting than \cite{xu2019number}. 
\end{itemize}
\vspace{-3.6mm}
\parbox{0.67\linewidth}{
\begin{itemize}[leftmargin=*,itemsep=0.8mm] 
    \item \textbf{``Negative Ridge'' Phenomenon.} In Section \ref{sec:sgnopt}, we analyze the optimal regularization strength $\lambda_{\opt}$ under different $\vSigma_w$, and provide precise conditions under which the optimal $\lambda_{\opt}$ is negative. In brief, we show that in the overparameterized regime, $\lambda_{\opt}$ is negative when the SNR is large and the large directions of $\vSigma_x$ and $\vSigma_\beta$ are aligned (see Figure \ref{fig:sign_of_parameter}), and vice versa. 
    In contrast, in the underparameterized regime ($p\!<\!n$), the optimal $\ell_2$ regularization is always non-negative. We also discuss the risk monotonicity of optimally regularized ridge regression under general data covariance and isotropic $\vbeta_*$. 
    \item \textbf{Optimal Weighting Matrix $\vSigma_w$.} In Section \ref{sec:optimal_w}, we decide the optimal  $\vSigma_w$ for both the optimally regularized ridge estimator ($\lambda \!=\! \lambda_{\opt}$) and the ridgeless limit ($\lambda \!\rightarrow\! 0$). In the ridgeless limit, based on the bias-variance decomposition, we show that in certain cases the optimal $\vSigma_w$ should interpolate between $\vSigma_x$, which minimizes the variance, and $\vSigma_{\beta}^{-1}$, which minimizes the bias (for more general setting see Theorem \ref{thm:optimalw_ridgeless}). 
    Whereas for the optimal ridge regression, in many settings the optimal $\vSigma_w$ is simply  $\vSigma_{\beta}^{-1}$ (see Figure \ref{fig:optimal_w}), which is independent of the eigenvalues of $\vSigma_x$ and the SNR (Theorem \ref{thm:optimalw_optimal} also presents more general cases). We demonstrate the advantage of weighted $\ell_2$ regularization over standard ridge regression and PCR, and also propose a heuristic choice of $\vSigma_w$ when information of the signal $\vbeta_\star$ is not present.
\end{itemize}
} 
\parbox{0.02\linewidth}{\hspace{0.1cm}}
\parbox{0.3\linewidth}{
{ 
\vspace{-3mm}
\begin{minipage}[t]{1\linewidth}
\centering
{\includegraphics[width=1.03\linewidth]{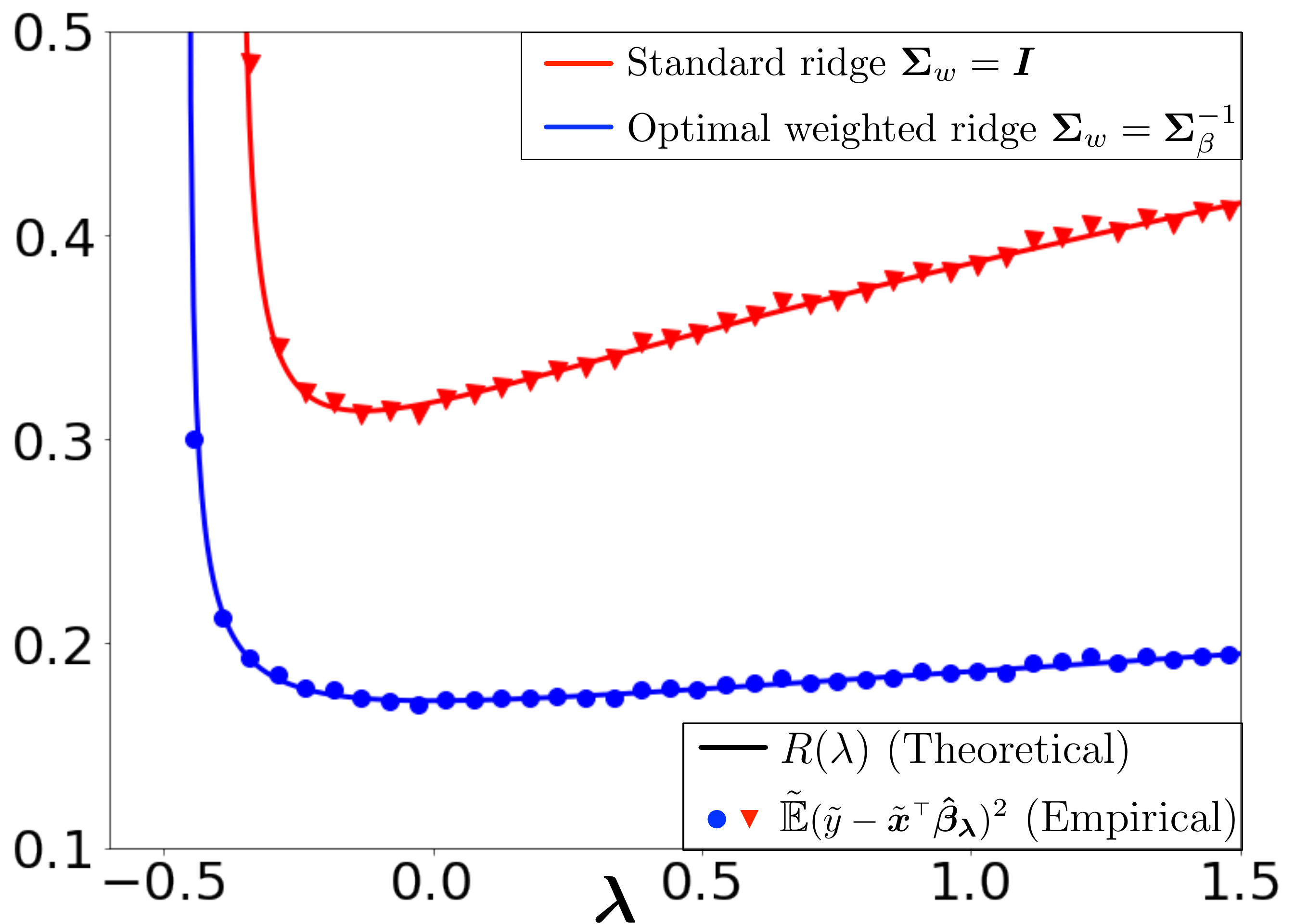}} \\ \vspace{-0.10cm}
\captionof{figure}{\small illustration of the ``negative ridge'' phenomenon and the advantage of weighted $\ell_2$ regularization under ``aligned'' $\vSigma_x$ and $\vSigma_{\beta}$. We set $\gamma=2$ and $\tilde{\sigma}^2=0$. Red: standard ridge regression ($\vSigma_w\!=\!\vI$); note that the lowest prediction risk is achieved when $\lambda<0$. Blue: optimally weighted ridge regression ($\vSigma_w\!=\!\vSigma_{\beta}^{-1}$), which achieves lower risk compared to the standard isotropic shrinkage.}
\label{fig:illustration}
\end{minipage}
}
}
 
\textbf{Notations:} We denote $\tilde{\bbE}$ as taking expectation over $\vbeta_{\star}, \tilde{x},\tilde{\epsilon}$. Let $\vd_x$, $\vd_{\beta}$, $\vd_w$ be the vectors of the eigenvalues of $\vSigma_x, \vSigma_{\beta}$ and $\vSigma_w$ respectively. We use $\bbI_{\cS}$ as the indicator function of set $\cS$. We write $\xi=\bbE (\tilde{\vx}^{\t}\vbeta_{\star})^2 / (\gamma\tilde{\sigma}^2)$ as the signal-to-noise ratio (SNR) of the problem. 
\vspace{-0.5mm}

\section{Related Works}
\vspace{-0.5mm}
\label{sec:related}
{
\textbf{Asymptotics of Ridge Regression.} 
The prediction risk of standard ridge regression ($\vSigma_w = \vI_d$) in the proportional asymptotics has been widely studied. When the data is isotropic, precise characterization can be obtained from random matrix theory \cite{karoui2013asymptotic,dicker2016ridge,hastie2019surprises}, approximate message passing algorithm \cite{donoho2016high}, or the convex Gaussian min-max theorem\footnote{Note that convergence and uniqueness of AMP and CGMT can be difficult to establish when $\lambda<0$. Also, to our knowledge the current AMP framework cannot handle \textit{joint} relation between $\vSigma_x$ and $\vSigma_{\beta}$, which is crucial for our ``negative ridge'' analysis.}\cite{thrampoulidis2018precise}. Under general data covariance, closely related to our work is \cite{dobriban2018high}, which assumed an isotropic prior on target coefficients ($\vSigma_{\beta} = \vI_d$). Our risk calculation builds upon the general random matrix result of \cite{rubio2011spectral,ledoit2011eigenvectors}. Similar tools have been applied in the analysis of sketching \cite{liu2019ridge} and the connection between ridge regression and early stopping \cite{ali2019continuous,lolas2020regularization}. 
   
\vspace{1.5mm}
\noindent\textbf{Weighted Regularization.}
The formulation \eqref{eq:problem} was first introduced in \cite{hoerl1970ridge}, and many choices of $\vSigma_w$ have been proposed \cite{strawderman1978minimax,casella1980minimax,maruyama2005new,mori2018generalized}; but since these estimators are usually derived in the $n>p$ setup, their effectiveness in the high-dimensional and overparameterized regime is largely unknown.
In semi-supervised linear regression, it is known that weighted matrix estimated from unlabeled data can improve the model performance \cite{ryan2015semi,tony2020semisupervised}. 
In deep learning, anisotropic Gaussian prior on the parameters enjoyed empirical success \cite{louizos2017multiplicative,zhao2019learning}. Additionally, decoupled weight decay \cite{loshchilov2017decoupled} and elastic weight consolidation \cite{kirkpatrick2017overcoming} can both be interpreted as $\ell_2$ regularization weighted by an approximate Fisher information matrix \cite[Sec.~3]{zhang2018three}, which relates to the Fisher-Rao norm \cite{liang2017fisher}.  
Finally, beyond the $\ell_2$ penalty, weighted regularization is also effective in LASSO regression \cite{zou2006adaptive,candes2008enhancing,bogdan2015slope}. 

\vspace{1.5mm}
\noindent\textbf{Benefit of Overparamterization.}
Our overparameterized setting is partially motivated by the double descent phenomenon \cite{krogh1992simple,belkin2018reconciling}, which can be theoretically explained in linear regression \cite{advani2017high,hastie2019surprises,bartlett2019benign}, random features regression \cite{mei2019generalization,d2020double,hu2020universality}, and max-margin classification \cite{deng2019model,montanari2019generalization}, although translation to neural networks can be nuanced \cite{ba2020generalization}. 
For least squares regression, 
it has been shown in special cases that overparameterization induces an implicit $\ell_2$ regularization \cite{kobak2020optimal,derezinski2019exact}, which agrees with the absence of overfitting. This observation also leads to the speculation that the optimal ridge penalty in the overparameterized regime may be negative, to partially cancel out the implicit regularization. 
While the possibility of negative ridge parameter has been noted in \cite{hua1983generalized,bjorkstrom1999generalized}, theoretical understanding of its benefit is largely missing, expect for heuristic argument (and empirical evidence) in \cite{kobak2020optimal}. We provide a rigorous characterization of this
``negative ridge'' phenomenon.
 
\vspace{1.5mm}
\noindent\textbf{Concurrent Works.} Independent to our work, \cite{richards2020asymptotics} computed the asymptotic prediction risk under a similar extension of isotropic $\vSigma_\beta$, but did not consider the sign of $\lambda_{\text{opt}}$ nor the weighted objective. We remark that their result requires codiagonalizable covariances and certain functional relation between eigenvalues, which is much more restrictive than our setting.
\cite{tsigler2020benign} provided a non-asymptotic analysis of ridge regression and constructed a specific spike covariance model\footnote{In contrast, we show that the sign of $\lambda_{\text{opt}}$ depends on ``alignment'' between $\vSigma_x$ and $\vbeta_*$, which goes beyond the spike setup.} for which negative regularization may lead to better generalization bound than interpolation ($\lambda=0$).
In a companion work \cite{amari2020does}, we connect properties of the ridgeless limit of the generalized ridge regression estimator to the implicit bias of preconditioned updates (e.g., natural gradient), which allows us to decide the optimal preconditioner in the interpolation setting. 
 
} 
\section{Setup and Assumptions}
\vspace{-0.5mm} 
\label{sec:assumption}

In addition to the prediction risk of the weighted ridge estimator $\hvb_{\lambda}=\left(\vX^{\t}\vX+\lambda\vSigma_w\right)^{\dag}\vX^{\t}\vy$, the setup of which we outlined in Section~\ref{sec:intro}, we also analyze the principal component regression (PCR) estimator: 
for $\theta\in[0,1]$, the PCR estimator is given as $\hvb_{\theta}=(\vX_{\theta}^{\t}\vX_{\theta})^{\dag}\vX_{\theta}^{\t}\vy$, where $\vX_{\theta}=\vX\vU_\theta$ and the columns of $\vU_\theta\in \bbR^{p\times \theta p}$ are the leading $\theta p$ eigenvectors of $\vSigma_x$. 

Under the setting on $(\tilde{\vx},\vbeta_{\star},\tilde{\epsilon})$ described in Section \ref{sec:intro}, the prediction risk of \eqref{eq:problem} can be simplified as
\begin{eqnarray}
    \tilde{\bbE}\left(\tilde{y}-\tilde{x}^{\t}\hat{\vbeta}_{\lambda}\right)^2&=& \underbrace{\tilde{\sigma}^2\left(1+\frac{1}{n}\trace{\vSigma_{x/w}\left(\vX_{/w}^{\t}\vX_{/w}+\lambda\vI\right)^{-1}-\lambda \vSigma_{x/w}\left(\vX_{/w}^{\t}\vX_{/w}+\lambda\vI\right)^{-2}}\right)}_{\text{Part 1, Variance}}\n\\
&& 
    +\underbrace{\frac{\lambda^2}{n}\trace{\vSigma_{x/w}\left(\vX_{/w}^{\t}\vX_{/w}+\lambda\vI\right)^{-1}\vSigma_{w\beta}\left(\vX_{/w}^{\t}\vX_{/w}+\lambda\vI\right)^{-1}}}_{\text{Part 2, Bias}},
\label{eq:risk_eq1} 
\end{eqnarray}
where $\vX_{/w}\!=\!\vX\vSigma_w^{-1/2}, \vSigma_{x/w}\!=\!\vSigma_w^{-1/2}\vSigma_x\vSigma_w^{-1/2}, \vSigma_{w\beta}\!=\!\vSigma_w^{1/2}\vSigma_{\beta}\vSigma_w^{1/2}$. 
Note that the variance term does not depend on the true signal, and the bias is independent of the noise level. 
Let $\vd_{x/w}$ be the eigenvalues of $\vSigma_{x/w}$ and $\vSigma_{x/w}=\vU_{x/w}\vD_{x/w}\vU_{x/w}^{\t}$ be the eigendecomposition of $\vSigma_{x/w}$, where $\vU_{x/w}$ is the eigenvector matrix and $\vD_{x/w}=\diag{\vd_{x/w}}$. 
Let $\vd_{w\beta}\triangleq\diag{\vU_{x/w}^\top\vSigma_{w\beta}\vU_{x/w}}$. 
When $\vSigma_w=\vI$, $\vd_{w\beta}$ characterizes the strength of the signal $\vbeta_{\star}$ along the directions of the eigenvectors of feature covariance $\vSigma_x$. 
To simplify the RHS of \eqref{eq:risk_eq1}, we make the following assumption: 
 
\begin{assumption}\label{ass:eigenvalues}
Let $d_{x/w,i}$ and $d_{w\beta, i}$ be the $i$th element of $\vd_{x/w}$ and $\vd_{w\beta}$ respectively. Then the empirical distribution of $(d_{x/w,i}, d_{w\beta,i})$ jointly converges to $(h, g)$ where $h$ and $g$ are two non-negative random variables. Further, there exists constants $c_l,c_u>0$ independent of $n$ and $p$ such that $\min_i d_{x/w,i}\geq c_l$, $\max_i(d_{x/w,i},d_{w\beta,i}))\leq c_u$ and $\|\vSigma_{w\beta}\|\leq c_u$. 
\end{assumption}
One can check that $\vSigma_x$ and $\vSigma_{\beta}$ studied in \cite{dobriban2018high, hastie2019surprises,xu2019number} (with $\vSigma_w=\vI$) are special cases of Assumption \ref{ass:eigenvalues} with either $h$ or $g$ being a point mass. 
It is clear that our Assumption \ref{ass:eigenvalues} allows the eigenvalues of $\vSigma_x$ and $\vSigma_\beta$ (when $\vSigma_w=\vI$) to follow much more general distributions.


\section{Risk Characterization}\label{sec:riskcal}

With the aforementioned assumptions, we now present our characterization of the prediction risk.
\begin{theo}\label{thm:RiskCal}
Under Assumption \ref{ass:eigenvalues}, the asymptotic prediction risk is given as
\begin{eqnarray}
\tilde{\bbE}\left(\tilde{y}-\tilde{x}^{\t}\hat{\vbeta}_{\lambda}\right)^2
&\ipe&
\frac{m'(-\lambda)}{m^2(-\lambda)}\cdot\left(\gamma\bbE\frac{gh}{(h\cdot m(-\lambda)+1)^2}+\tilde{\sigma}^2\right):=R(\lambda), \ \forall \lambda>-c_0
\label{eq:riskformula}
\end{eqnarray}
where $c_0=(\sqrt{\gamma}-1)^2c_l$, and $m(z)$ is the Stieltjes transform of the limiting distribution of the eigenvalues of $\vX_{/w}\vX_{/w}^{\t}$. Additionally, $m(-\lambda), m'(-\lambda)>0$ satisfy the following:
\begin{eqnarray}
\lambda &=&\frac{1}{m(-\lambda)}-\gamma\bbE \frac{h}{1+h\cdot m(-\lambda)}\label{eq:relation_eq1}\\
1&=&\left(\frac{1}{m^2(-\lambda)}-\gamma\bbE\frac{h^2}{(h\cdot m(-\lambda)+1)^2}\right)m'(-\lambda).\label{eq:relation_eq2}
\end{eqnarray}
\end{theo}

\begin{figure}[!htb]
\vspace{-1.5mm}
\centering
\begin{minipage}{1.\linewidth}
\centering
\subcaptionbox{Aligned, noiseless }{\vspace{-0.15cm}\includegraphics[width=0.32\linewidth]{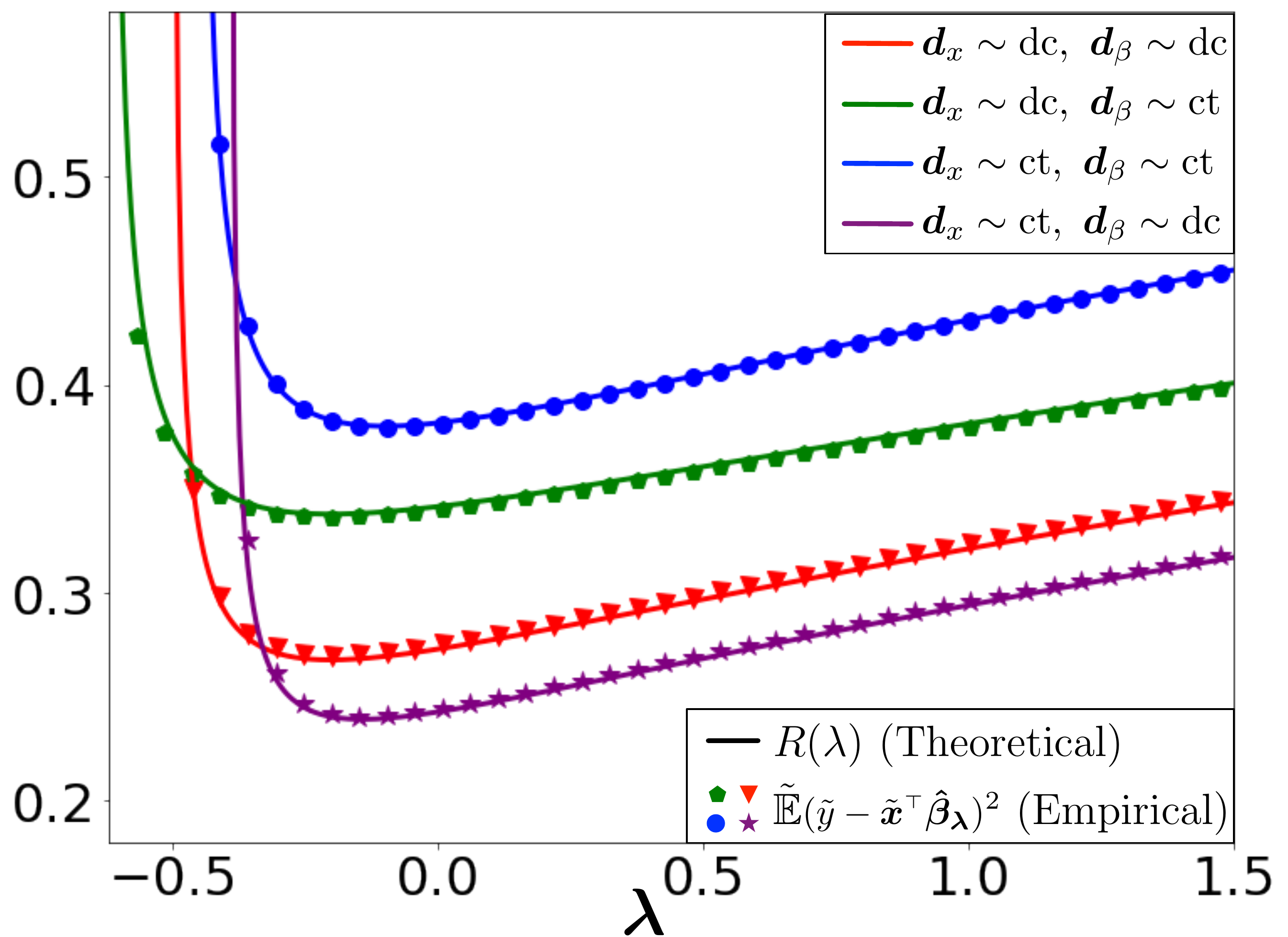}}
\subcaptionbox{Misaligned, noiseless
}{\vspace{-0.15cm}\includegraphics[width=0.328\linewidth]{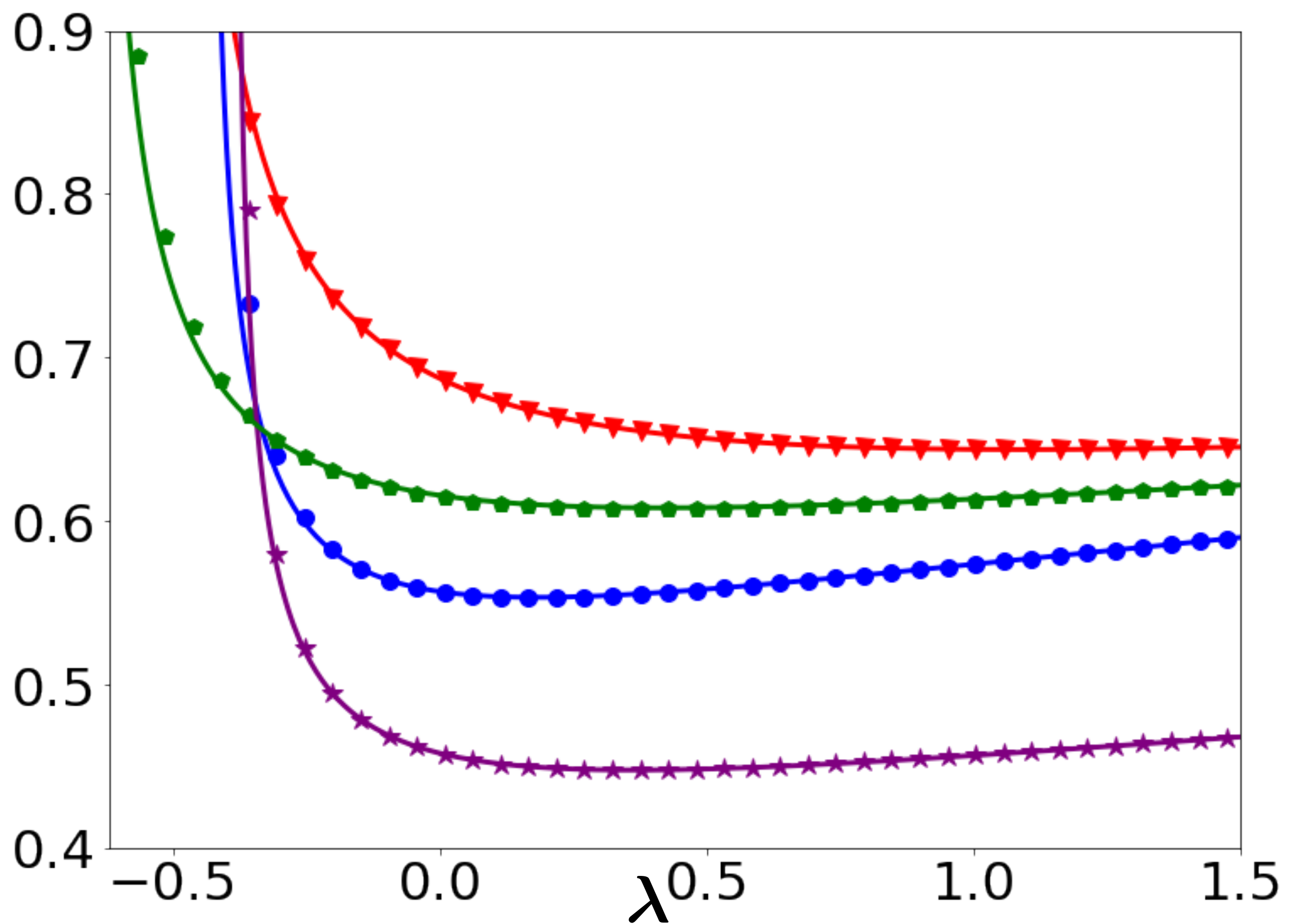}}
\subcaptionbox{Random, noiseless
}{\vspace{-0.15cm}\includegraphics[width=0.32\linewidth]{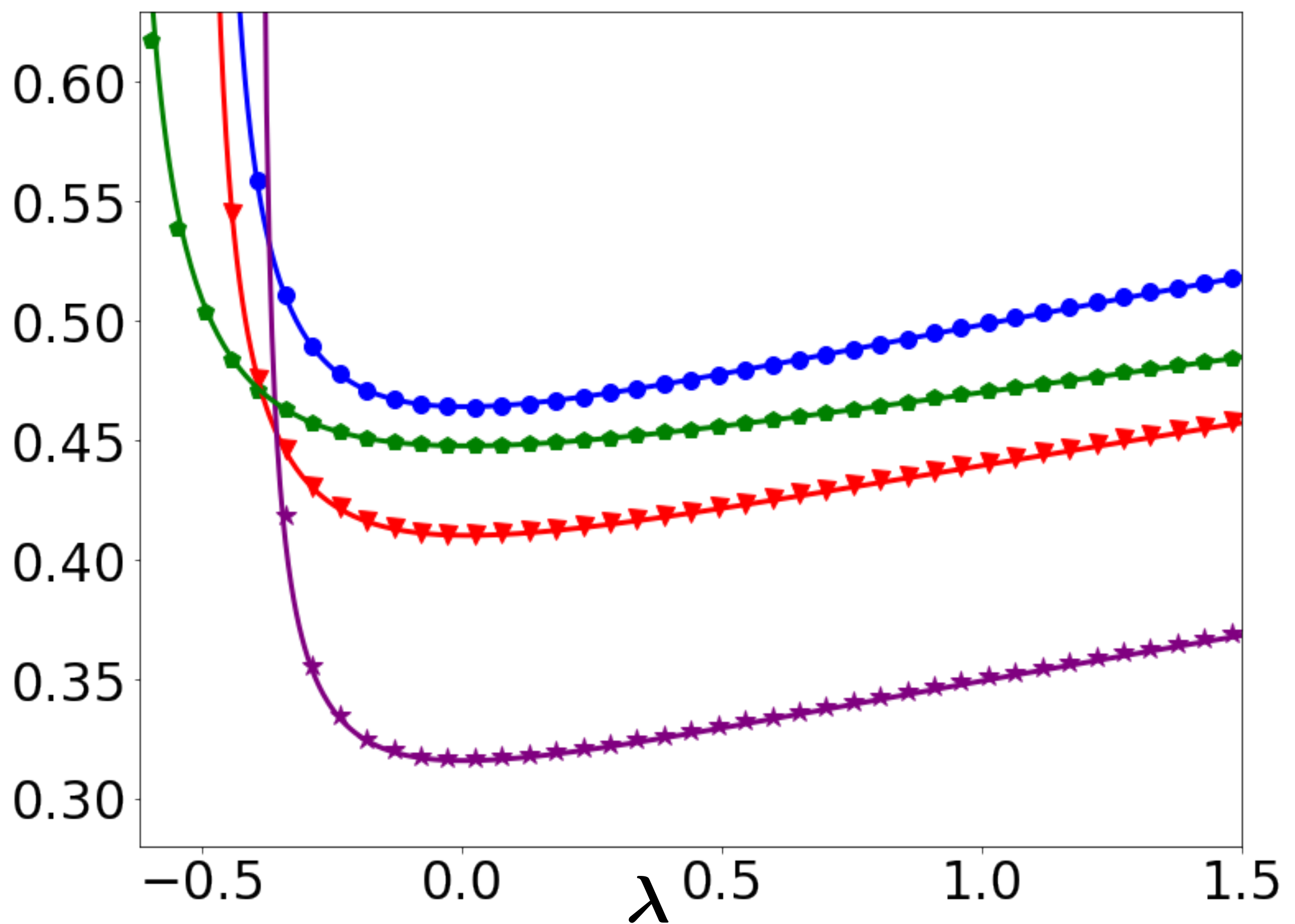}}
\end{minipage} 

\vspace{-0.15cm}
\caption{\small Finite sample prediction risk $\tilde{\bbE}(\tilde{y}-\tilde{\vx}^{\t}\vbeta_{\star})^2$ (experiment) and the asymptotic risk $R(\lambda)$ (theory) against $\lambda$ for standard ridge regression ($\vSigma_w = \vI_d$). We set $\gamma=2$ and $(n,p)=(300, 600)$. `dc' and `ct' stand for for discrete and continuous distribution, respectively. We write `aligned' if $\vd_x$ and $\vd_{\beta}$ have the same order, `misaligned' for the reverse and `random' for random order. Different colors indicate different combinations of $\vd_x$ and $\vd_{\beta}$. Note that our derived risk $R(\lambda)$ matches the experimental values, and in the aligned and noiseless case, the optimal risk is achieved when $\lambda<0$ (predicted by Theorem~\ref{thm:optimalridge}). Plots for the noisy case is presented in Appendix \ref{app:figure}.  
\vspace{-1mm}
}
\label{fig:risk} 
\end{figure}

Note that the condition $\lambda>-c_0$ ensures both $m(-\lambda)$ and $m'(-\lambda)$ exist and are positive. Furthermore, it can be shown from prior works \cite{dobriban2018high, xu2019number} that the variance term (part 1) in \eqref{eq:risk_eq1}, converges to $\tilde{\sigma}^2\frac{m'(-\lambda)}{m^2(-\lambda)}$. Our main contribution is to characterize the bias term, Part 2, under significantly less restrictive assumption on $(\vSigma_x,\vSigma_\beta, \vSigma_{w})$. In particular, building upon \cite{rubio2011spectral}, we show that
\[
    \text{Part 2} \ \ipe \ \frac{m'(-\lambda)}{m^2(-\lambda)}\cdot\gamma\bbE\frac{gh}{(h\cdot m(-\lambda)+1)^2}, \quad \forall \lambda>-c_0.
\]

We illustrate the results of Theorem \ref{thm:RiskCal} in Figure \ref{fig:risk} (noiseless case) and Figure \ref{fig:risk2} (noisy case) for both discrete and continuous design for $\vd_x$ and $\vd_{\beta}$ with $\vSigma_x=\diag{\vd_x}, \vSigma_{\beta}=\diag{\vd_{\beta}}$ and $\vSigma_w=\vI$ (see design details in Appendix \ref{app:figure}). Note that Assumption \ref{ass:eigenvalues} specifies a joint relation between $\vd_x(=\vd_{x/w})$ and $\vd_{\beta}(=\vd_{w\beta})$. In the following section, we mainly consider the three following relations, which allow us to precisely determine the sign of $\lambda_{\opt}$.
\begin{defi}
For two vectors $\va,\vb\in \bbR^p$, we say $\va$ is aligned (misaligned) with $\vb$ if the order of $\va$ is the same as (reverse of) the order of $\vb$, i.e., $a_i\ge a_j$ if and only if $b_i\ge (\le)~b_j$ for all $i,j$. Additionally, we say $\va$ and $\vb$ have random relation if given one order, the other is uniformly permuted at random.  
\end{defi}
 


\begin{wrapfigure}{R}{0.25\textwidth}  
\vspace{-4.6mm}
\centering 
\includegraphics[width=0.245\textwidth]{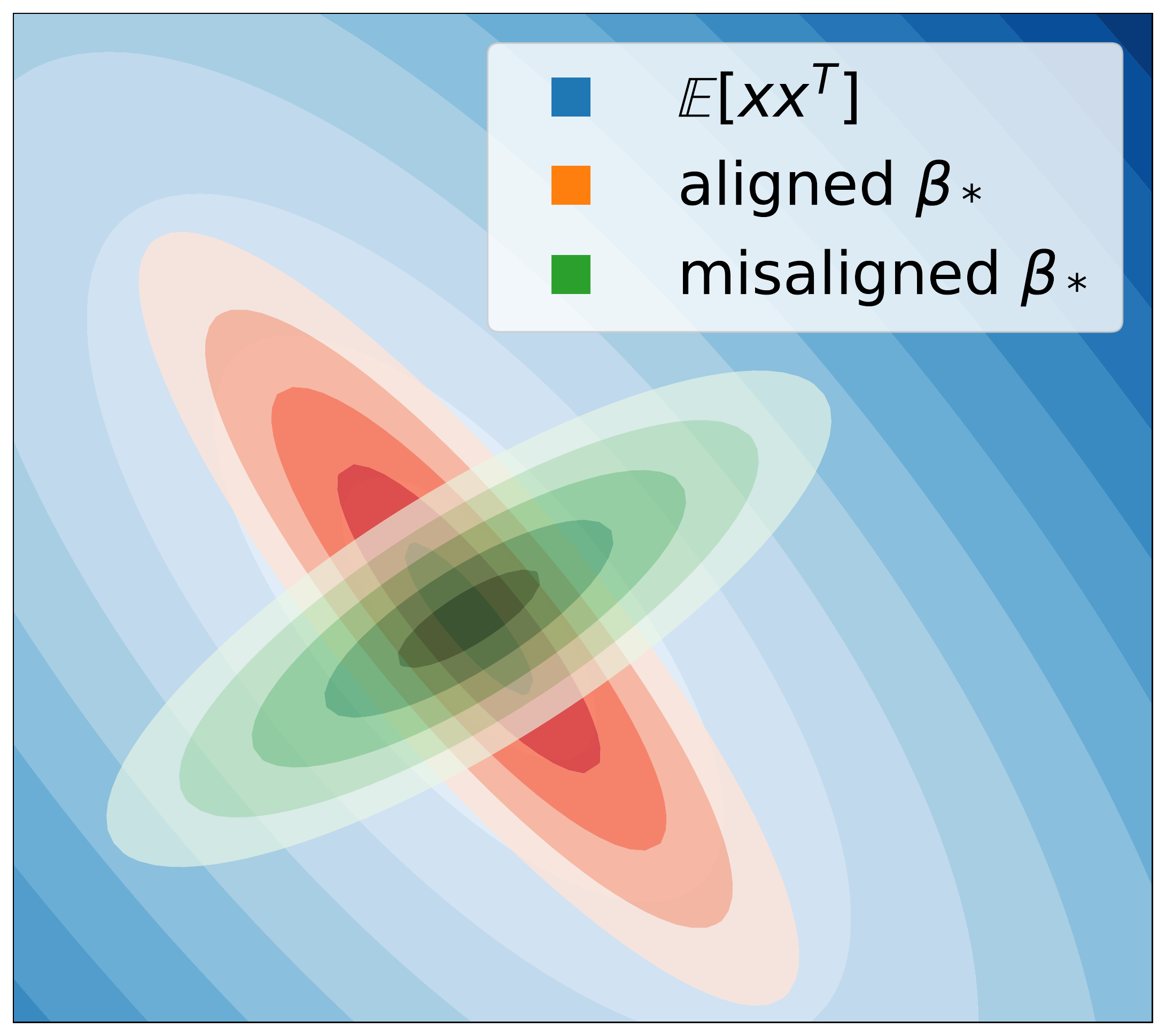}
\vspace{-7mm} 
\caption{\small 2D illustration of alignment between $\vx$ and $\vbeta_*$.}  
\label{fig:alignment-illustration}
\vspace{-7.5mm} 
\end{wrapfigure} 

Intuitively speaking (see Figure~\ref{fig:alignment-illustration}), aligned $\vd_x$ and $\vd_{\beta}$ implies that when one component in $\vd_x$ has large magnitude, then so does the corresponding component in $\vd_{\beta}$; in this case the features are informative and the learning problem is ``easy''. In contrast, misaligned $\vd_x$ and $\vd_{\beta}$ suggests that features with larger magnitude contribute less to the labels, and thus learning is ``difficult''.

In Figure \ref{fig:risk}, we plot the prediction risk of all three joint relations defined above (see Appendix \ref{app:figure} for details). In addition, note that Theorem~\ref{thm:RiskCal} allows us to compute the risk of the generalized ridge estimator $\hat{\vbeta}_{\lambda}$ as well as its ridgeless limit, which yields the minimum $\|\hat{\vbeta}\|_{\vSigma_w}$ norm solution (taking $\vSigma_w = \vI$ recovers the minimum $\ell_2$ norm solution studied in \cite{hastie2019surprises,belkin2019two}).

\vspace{-1.5mm}
\paragraph{Connection to PCR estimator.} Note that the principal component regression (PCR) estimator is closely related to the ridgeless estimator in the following sense: intuitively, picking the leading $\theta p$ eigenvectors of $\vSigma_x$ (for some $\theta\in [0,1]$) is equivalent to setting the remaining $(1-\theta)p$ eigenvalues of $\vSigma_w$ to be infinity \cite{hua1983generalized}. 
The following corollary characterizes the prediction risk of the PCR estimator $\hat{\beta}_{\theta}$:

\begin{coro}\label{cor:PCRrisk}
Given Assumption \ref{ass:eigenvalues} and $\vSigma_w=\vI$, and $h$ has continuous and strictly increasing quantile function $Q_h$. Then for all $\theta \in (0,1]$, as $n,p\rightarrow \infty$,
\begin{eqnarray}
\tilde{\bbE}\left(\tilde{y}-\tilde{x}\hat{\vbeta}_{\theta}\right)^2
&\ipe&
\left\{\begin{aligned}
&\frac{m'_{\theta}(0)}{m^2_{\theta}(0)}\cdot\left(\gamma\bbE\frac{gh}{(h_{\theta}\cdot m_{\theta}(0)+1)^2}+\tilde{\sigma}^2\right),&&\theta\gamma>1\\
&\left(\gamma\bbE [gh\cdot \bbI_{h<Q_h(1-\theta)}]+\tilde{\sigma}^2\right)\frac{1}{1-\theta\gamma},&&\theta\gamma<1
\end{aligned}\right.
\label{eq:riskPCRformula}
\end{eqnarray}
where $h_{\theta}=h\cdot \bbI_{h\geq Q_h(1-\theta)}$ and $m_{\theta}(z)$ satisfies 
$-z =m^{-1}_{\theta}(z)-\gamma\bbE h_{\theta}\cdot(1+h_{\theta}\cdot m_{\theta}(z))^{-1}$.

In addition, if $\bbE[g|h]$ is a decreasing function of $h$, and $h$ has continuous p.d.f., then the asymptotic prediction risk of $\hvb_{\theta}$ is a decreasing function of $\theta$ when $\theta\gamma>1$.   
\end{coro}

Corollary \ref{cor:PCRrisk} confirms the double descent phenomenon under more general settings of $(\vSigma_x,\vSigma_\beta)$ than \cite{xu2019number}, i.e. the prediction risk exhibits a spike as $\theta\gamma\rightarrow 1^{-}$, and then decreases as we further overparameterize by increasing $\theta$. 
In Section \ref{sec:optimal_w} we compare the PCR estimator $\hat{\vbeta}_{\theta}$ with the minimum $\|\hvb\|_{\vSigma_w}$ norm solution.
\begin{remark}
The PCR estimator \cite{xu2019number} and the ridgeless regression estimator (considered in \cite{hastie2019surprises}) are fundamentally different in the following way: in ridgeless regression, increasing the model size corresponds to changing $\gamma$, which also alters the dimensions of the true coefficients $\vbeta_\star$; in contrast, in PCR, increasing $\theta$ does not change the data generating process (which is a more natural setting). 

In terms of the risk curve, Figure~\ref{fig:multiple_descent}(a) shows that the ridgeless regression estimator can exhibit ``\textit{multiple descent}'' as $\gamma>1$ increases due to our general anisotropic setup, whereas Corollary~\ref{cor:PCRrisk} and Figure~\ref{fig:multiple_descent}(b) demonstrate that in the misaligned case, the PCR risk is monotonically decreasing in the overparameterized regime $\theta\gamma>1$, which illustrates the benefit of overparameterization.
\end{remark}


\section{Analysis of Optimal $\lambda_{\opt}$}\label{sec:sgnopt}

In this section, we focus on the optimal weighted ridge estimator and determine the sign of the optimal regularization parameter $\lambda_{\opt}$. Taking the derivatives of \eqref{eq:riskformula} yields
\begin{eqnarray}
R'(\lambda)=\frac{2\gamma(m'(-\lambda))^2}{m^3(-\lambda)}\left(  \underbrace{\left(-\tilde{\sigma}^2\frac{\bbE \frac{\zeta^2}{(1+\zeta)^3}}{1-\gamma\bbE\frac{\zeta^2}{(1+\zeta)^2}}\right)}_{\text{Part 3}}+\underbrace{\left(\bbE\frac{gh\zeta}{(1+\zeta)^3}-\frac{\gamma\bbE \frac{\zeta^2}{(1+\zeta)^3}\bbE\frac{gh}{(1+\zeta)^2}}{1-\gamma\bbE\frac{\zeta^2}{(1+\zeta)^2}}\right)}_{\text{Part 4}}\right),
\label{eq:derivative}
\end{eqnarray}
where $\zeta=h\cdot m(-\lambda)$. For certain special cases, we obtain a closed form solution for $\lambda_{\opt}$ (see details in Appendix \ref{sec:simplify}) and recover the result from \cite{hastie2019surprises,dobriban2018high}\footnote{In \cite{hastie2019surprises}, $h\aseq 1$. In \cite{dobriban2018high}, $\tilde{\sigma}^2=1$ and their signal strength $\alpha^2$ is equivalent to $c\gamma$ in our setting.} and beyond:
\begin{itemize}
    \item When $h\aseq c$ (i.e., isotropic features \cite{hastie2019surprises}), the optimal $\lambda_{\opt}$ is achieved at $c/\xi$. 
    \item When $g\aseq c$ (i.e., isotropic signals  \cite{dobriban2018high}), the optimal $\lambda_{\opt}$ is achieved at $\tilde{\sigma}^2/c$. 
    \item When $\bbE[g|h]\aseq \bbE[g]$ (e.g., random order), the optimal $\lambda_{\opt}$ is achieved at $\tilde{\sigma}^2/\bbE[g]$.
\end{itemize}

Although $\lambda_{\opt}$ may not have a tractable form in general, we may infer the sign of $\lambda_{\opt}$. Recall that in \eqref{eq:derivative}, Part 3 is due to the variance term (Part 1) and Part 4 from the bias term (Part 2) in \eqref{eq:risk_eq1}. We therefore consider the sign of Part 3 and Part 4 separately in the following theorem. 
\begin{theo}\label{thm:optimalridge}
Under Assumption \ref{ass:eigenvalues}, we have
\begin{itemize}
    \item Part 3 (derivative of variance) is negative for all $\lambda>-c_0$.
    \item If $\bbE[g|h]$ is an increasing function of $h$ on its support, then Part 4 (derivative of bias) is positive for all $\lambda>0$. At $\lambda=0$, Part 4 is non-negative and achieves $0$ only if $\bbE[g|h]\aseq \bbE[g]$. 
    \item If $\bbE[g|h]$ is a decreasing function of $h$ on its support, then Part 4 is negative for all $\lambda\in (-c_0,0)$. At $\lambda=0$, Part 4 is non-positive and achieves $0$ only if $\bbE[g|h]\aseq \bbE[g]$. 
\end{itemize}
\end{theo}

The first point in Theorem \ref{thm:optimalridge} is consistent with the well-understood variance reduction property of ridge regularization. On the other hand, when the prediction risk is dominated by the bias term (i.e., $\tilde{\sigma}^2=o(1)$) and both $\vd_{x/w}$ and $\vd_{w\beta}$ converge to non-trivial distributions, the second and third point of Theorem \ref{thm:optimalridge} reveal the following surprising phenomena (see Figure \ref{fig:risk} (a) and (b)): 
 
\begin{figure}[t]
\centering
\begin{minipage}[t]{0.4\linewidth}
\centering
{\includegraphics[width=0.96\textwidth]{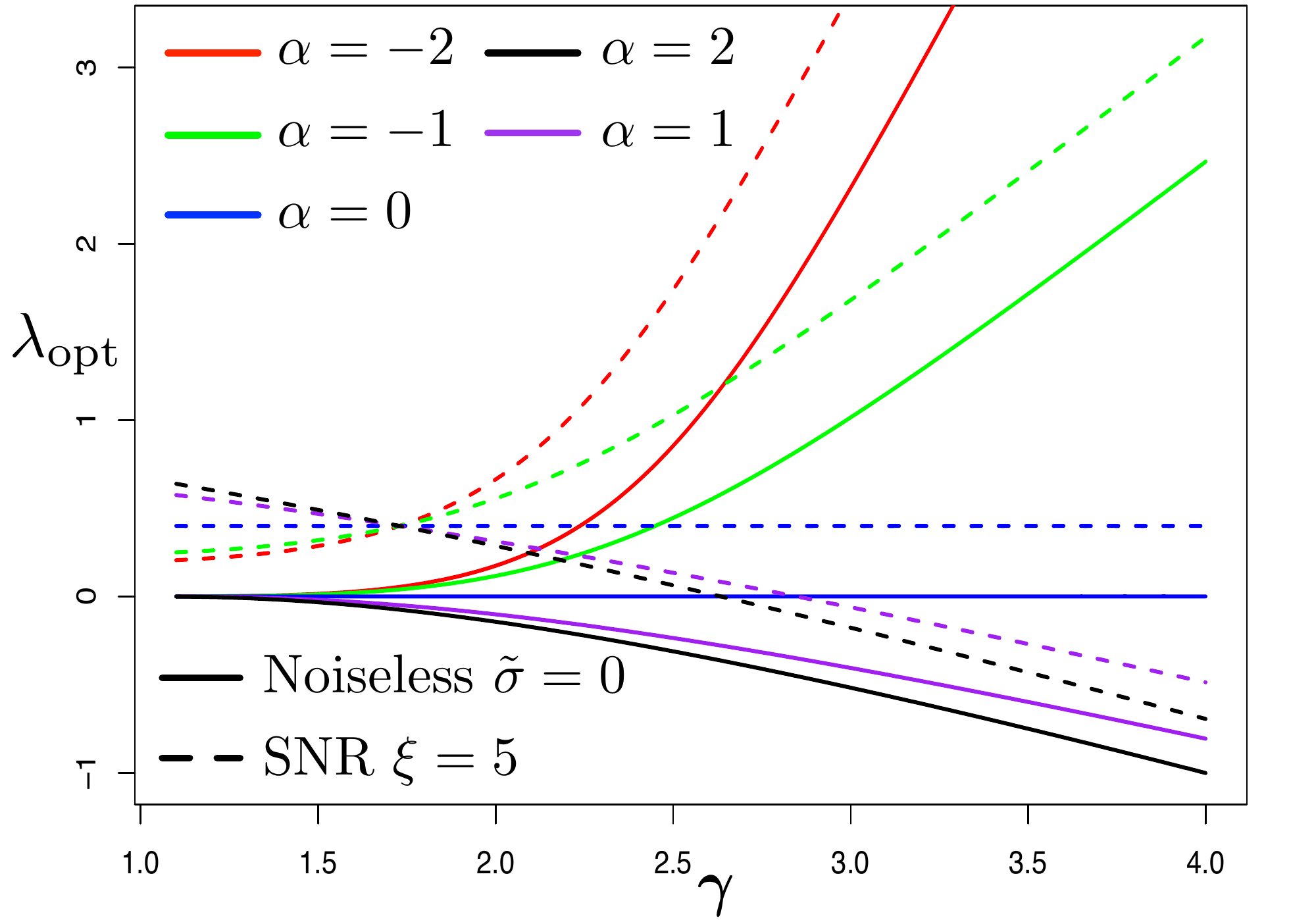}} \\ \vspace{-0.10cm}
Left: Optimal Ridge $\lambda_{\opt}$.
\end{minipage}
\begin{minipage}[t]{0.4\linewidth}
\centering
{\includegraphics[width=0.96\textwidth]{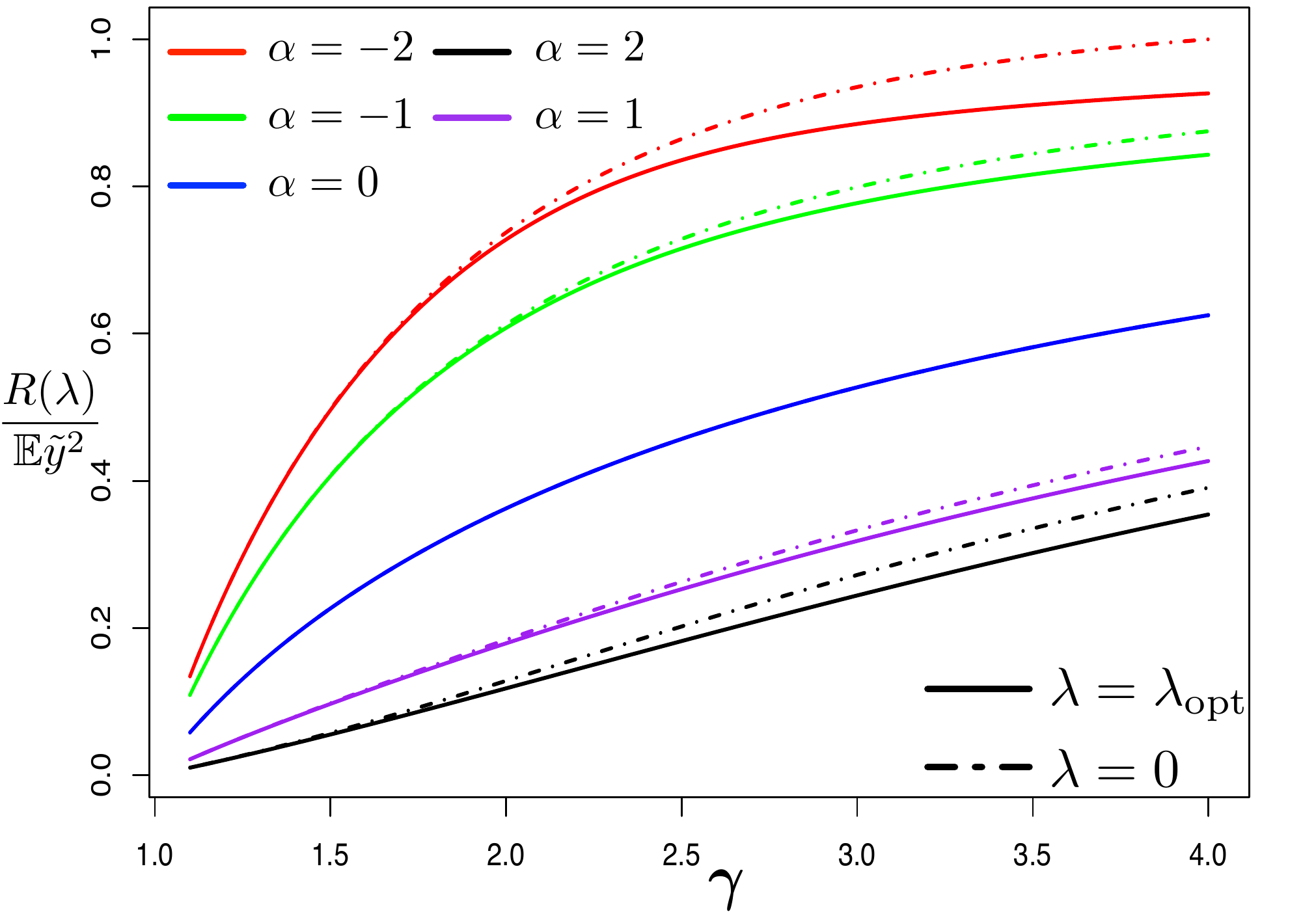}} \\ \vspace{-0.10cm}
Right: $R(\lambda)$: Optimal vs.~ridgeless.
\end{minipage}
\caption{\small We set $\vSigma_w=\vI$ and $\vSigma_\beta = \vSigma_x^{\alpha}$ where $\vd_x$ has two point masses on $1$ and $5$ with probability $3/4$ and $1/4$ respectively. \textbf{Left}: optimal $\lambda$; solid lines represents the noiseless case $\tilde{\sigma}=0$ and dashed lines represents the noisy case with a fixed SNR $\xi=5$. 
\textbf{Right}: comparison of prediction risk of the ridgeless ($R(\lambda_{\opt})$, dashed lines) and optimally regularized ($R(\lambda_{\opt})$, solid lines) estimator in the noiseless case. 
We normalize the prediction risk as $\bbE\tilde{y}^2= \bbE(\tilde{\vx}^{\t}\vbeta_{\star})^2$.  \vspace{-4mm}
}
\label{fig:sign_of_parameter}
\end{figure}
\begin{itemize}
    \item[\textbf{M1}] $\lambda_{\opt}<0$ when $\vd_{x/w}$ aligns with $\vd_{w\beta}$, or in general, $\bbE [g|h]$ is a strictly increasing function of $h$. In the context of standard ridge regression, it means that shrinkage regularization only \textit{increases the bias} in the overparameterized regime when features are informative, i.e., the projection of the signal is large in the directions where the feature variance is large.
    \item[\textbf{M2}] $\lambda_{\opt}>0$ when $\vd_{x/w}$ is misaligned with $\vd_{w\beta}$, or in general, $\bbE [g|h]$ is a strictly decreasing function of $h$. This is to say, in standard ridge regression, when features are not informative, i.e., the projection of the signal is small in the directions of large feature variance, shrinkage is beneficial even in the \textit{absence of label noise} (the variance term is zero). 
\end{itemize}

M1 and M2, together with aforementioned special case when $g$ and $h$ have random relation, provide a precise characterization of the sign of $\lambda_{\opt}$.
In particular, M1 confirms the ``negative ridge'' phenomenon empirically observed in \cite{kobak2020optimal} and outlines concise conditions under which it occurs. 
We emphasize that neither M1 nor M2 would be observed when one of $\vSigma_{x/w}$ and $\vSigma_{w\beta}$ is identity (as previously discussed). In other words, these observations arise from our more general assumption on $(\vSigma_{x/w}, \vSigma_{w\beta})$.

\vspace{-1.5mm}
\paragraph{Implicit regularization of overparameterization.}
Taking both the bias and variance into account, Theorem \ref{thm:optimalridge} demonstrates a bias-variance tradeoff between Part 3 and Part 4, and $\lambda_{\opt}$ will eventually become positive as $\tilde{\sigma}^2$ increases (i.e., the prediction risk is dominated by variance, for which a positive $\lambda$ is beneficial). For certain special cases, we can provide a lower bound for the transition from $\lambda_{\opt}\!<\!0$ to  $\lambda_{\opt}\!>\!0$.
\begin{prop}\label{thm:noiselevel}
Given Assumption \ref{ass:eigenvalues}, let $(h,g)=(1,1)$ with probability $1-q$ and $(h,g)=(h_1, g_1)$ with probability $q$, where $h_1>1$ and $g_1>1$. Denote $\bar{\gamma} = \gamma-1$. Then $\lambda_{\opt}<0$ if 
\[
\tilde{\sigma}^2<(h_1-1)(g_1-1)h_1\cdot \max\left(\dfrac{(\gamma q -1)^3\bar{\gamma}^3(1-q)}{(1-q)\gamma^2\left(\bar{\gamma}^3q^2+(\gamma q -1)^3h_1^2\right)},\dfrac{\gamma q(1-q)\bar{\gamma}^3}{(1-q)\left(h_1+\bar{\gamma}\right)^3+ q h_1^2\gamma^3} \right).
\]
\end{prop}
As $q$ approaches $0$ or $1$, the above upper bound goes to $0$ because $\vSigma_x$, $\vSigma_\beta$ becomes closer to $\vI$. Otherwise, when $\gamma q>1$, the upper bound suggests $\tilde{\sigma}^2=O(g_1 \gamma)$ which implies that SNR $\xi=\Omega(h_1/\gamma)$. Hence, as $\gamma$ increases, optimal $\lambda_{\opt}$ remains negative for a lower SNR (i.e., larger noise), which coincides with the intuition that overparameterization has an implicit effect of $\ell_2$ regularization (Figure~\ref{fig:sign_of_parameter} Left). 
Indeed, the following proposition suggests such implicit regularization is only significant in the overparameterized regime:
\begin{prop}\label{cor:under-parameterized}
When $\gamma<1$, $\lambda_{\opt}$ on $(-c_0,\infty)$ is always non-negative under Assumption \ref{ass:eigenvalues}. 
\end{prop}

Figure \ref{fig:sign_of_parameter} confirms our findings in Theorem \ref{thm:optimalridge} (for results on different distributions see Figure \ref{fig:sign_of_parameter_appendix}). 
Specifically, we set $\vSigma_w=\vI, \vSigma_x=\diag{\vd_x}$ and $\vSigma_\beta = \vSigma_x^{\alpha}$. As we increase $\alpha$ from negative to positive, the relation between $\vd_x$ and $\vd_\beta$ transitions from misaligned to aligned. The left panel shows that the sign of $\lambda_{\opt}$ is the exact opposite to the sign of $\alpha$ in the noiseless case (i.e. the variance is $0$), which is consistent with M1 and M2. Moreover, when $\vd_x$ aligns with $\vd_\beta$, $\lambda_{\opt}$ decreases as $\gamma$ becomes larger, which agrees with our observation on the implicit $\ell_2$ regularization of overparameterization.
Last but not least, in Figure \ref{fig:sign_of_parameter} (Right) we see that the optimal ridge regression estimator leads to considerable improvement over the ridgeless estimator. We comment that this improvement becomes more significant as $\gamma$ or condition number of $\vSigma_x$ and $\vSigma_{\beta}$ increases. 

\vspace{-1.5mm}
\paragraph{Risk monotonicity of optimal ridge regression.}  
\cite[Proposition 6]{dobriban2020wonder} showed that for isotropic data ($\vSigma_x=\vI$), the asymptotic prediction risk of optimally-tuned ridge regression monotonically increases with $\gamma$. This is to say, under proper regularization ($\lambda=\lambda_{\text{opt}}$), increasing the size of training data always helps the test performance.
Here we extend this result to data with general covariance and isotropic prior on $\vbeta_*$.
\begin{prop}\label{prop:risk-monotonicity}
Given $\E[\vx\vx^\top] = \vSigma_x$ and  $\E[\vbeta_*\vbeta_*^\top] = \frac{c}{p}\vI$\footnote{Note that the scaling of the parameters differs from the previous setting by $\gamma$ to be consistent with that of \cite{dobriban2020wonder}.} where $\vSigma_x$ satisfies Assumption~\ref{ass:eigenvalues}, the prediction risk $R(\lambda)$ of the optimally-tuned ridge regression estimator ($\vSigma_w=\vI$) with $\lambda_{\mathrm{opt}}=\gamma\tilde{\sigma}^2/c$ is an increasing function of $\gamma\in(0, \infty)$.  
\end{prop} 

We remark that establishing such characterization under general orientation of $\vbeta_*$ (anisotropic $\vSigma_\beta$) can be challenging, because the optimal regularization $\lambda_\opt$ may not have a convenient closed-form. We leave the analysis for the general case as future work.


\section{Optimal Weighting Matrix}\label{sec:optimal_w}
Having characterized the optimal regularization strength, we now turn to the optimal choice of weighting matrix $\vSigma_w$. Toward this goal, we additionally require the following assumptions on $(\vSigma_x,\vSigma_{\beta},\vSigma_w)$: \begin{assumption}\label{ass:eigenvector}
The covariance matrix $\vSigma_x$ and the weighting matrix $\vSigma_w$ share the same set of eigenvectors, i.e., we have the following eigendecompositions: $\vSigma_x=\vU\vD_x\vU^{\t}$ and $\vSigma_w=\vU\vD_w\vU^{\t}$, where $\vU\in \bbR^{p\times p}$ is orthogonal, and $\vD_x=\diag{\vd_x}, \vD_w=\diag{\vd_w}$.
\end{assumption}
We define $\bar{\vd}_{\beta}=\diag{\vU^\t\vSigma_{\beta}\vU}$. Note that when $\vSigma_\beta$ also shares the same eigenvector matrix $\vU$, then $\bar{\vd}_{\beta}=\vd_{\beta}$, which is simply the eigenvalues of $\vSigma_{\beta}$. 
\begin{assumption}\label{ass:optimalw}
Let $d_{x,i}, \bar{d}_{\beta,i}, d_{w,i}$ be the $i$th element of $\vd_x,\bar{\vd}_{\beta},\vd_w$ respectively. We assume that the empirical distribution of $(d_{x,i}, \bar{d}_{\beta,i}, d_{w,i})$ jointly converges to $(s, v, s/r)$, where $s,v,r$ are non-negative random variables. Further, there exists constants $c_l,c_u>0$ independent of $n$ and $p$ such that $\min_i(\min(d_{x,i}, \bar{d}_{\beta,i}, d_{w,i}))\geq c_l$, $\max_i(\max(d_{x,i},\bar{d}_{\beta,i},d_{w,i}))\leq c_u$ and $\|\vSigma_{\beta}\|\leq c_u$. 
\end{assumption}
For notational convenience, we define $\cH_w$ and $\cH_r$ to be the sets of all $\vSigma_w$ and $r$, respectively, that satisfy Assumption \ref{ass:eigenvector} and Assumption \ref{ass:optimalw}. 
Additionally, let $\cS_w$ and $\cS_r$ be the subset of $\cH_w$ and $\cH_r$ such that $r=f(s)$ for some function $f$ (this represents $\vSigma_w\in\cH_w$ that only depends on $\vSigma_x$ but not $\vSigma_{\beta}$). 
By Assumption \ref{ass:eigenvector} and \ref{ass:optimalw}, we know the empirical distribution of $(d_{x/w,i}, d_{w\beta,i})$ jointly converges to $(r, sv/r)$ and satisfies the boundedness requirement in Assumption \ref{ass:eigenvalues}. We therefore apply Theorem \ref{thm:RiskCal} to compute the prediction risk:
\begin{eqnarray}
&&R(r, \lambda)\ \triangleq\ \frac{m_r'(-\lambda)}{m_r^2(-\lambda)}\cdot\left(\gamma\bbE\frac{sv}{(r\cdot m_r(-\lambda)+1)^2}+\tilde{\sigma}^2\right),\label{eq:weightrisk}
\end{eqnarray}
where $m_r(-\lambda)$ satisfies the equation $\lambda =m^{-1}_r(-\lambda)-\gamma\bbE (1+r\cdot m_r(-\lambda))^{-1}r$.
It is clear that when $r\aseq s$, \eqref{eq:weightrisk} reduces to the standard ridge regression with $\vSigma_w=\vI$, and for $r\aseq 1$, the equation reduces to the cases of isotropic features ($\vSigma_w=\vSigma_x$). 
Note that \eqref{eq:weightrisk} indicates that the impact of $\vSigma_{\beta}$ on the risk is fully captured by $\bar{\vd}_{\beta}$. Hence we define $\bar{\vSigma}_{\beta}=\vU\diag{\bar{\vd}_{\beta}}\vU^{\t}$, which corresponds to $r\aseq sv$, and is equivalent to $\vSigma_{\beta}$ when $\vSigma_{\beta}$ also shares the same eigenvector matrix $\vU$. 
In the following subsections, we discuss the optimal $\vSigma_w$ for two types of estimator: the minimum $\|\hat{\beta}\|_{\vSigma_w}$ solution (taking $\lambda\to 0$), and the optimally-tuned generalized ridge estimator ($\lambda=\lambda_{\opt}$). Note that the risk for both estimators is scale-invariant over $\vSigma_w$ and $r$. Hence, when we define a specific choice of $(\vSigma_w, r)$, we simultaneously consider all pairs $(c\vSigma_w, \nicefrac{r}{c})$ for $c>0$. 
Finally, we note that the choice of $r\aseq s\cdot\bbE[v|s]\in \cS_r$ plays a key role in our analysis, and its corresponding choice of $\vSigma_w$ is given as $\vSigma_w=\left(f_v(\vSigma_x)\right)^{-1}$, where $f_v(s)\triangleq \bbE[v|s]$ and $f_v$ applies to the eigenvalues of $\vSigma_x$.

\subsection{Minimum $\|\hat{\beta}\|_{\vSigma_w}$ solution}
Taking the ridgeless limit leads to the following bias-variance decomposition of the prediction risk,
\[
\text{Bias:} \quad \rb(r)\triangleq \frac{m_r'(0)}{m_r^2(0)}\cdot\gamma\bbE\frac{sv}{(r\cdot m_r(0)+1)^2}\quad \text{Variance:} \quad \rv(r)\triangleq\frac{m_r'(0)}{m_r^2(0)}\cdot\tilde{\sigma}^2.
\]
In the previous sections we observe a bias-variance tradeoff in choosing the optimal $\lambda$. 
Interesting, the following theorem illustrates a similar bias-variance tradeoff in choosing the optimal $\vSigma_w$:
\begin{theo}\label{thm:optimalw_ridgeless}
Given Assumptions \ref{ass:eigenvector} and \ref{ass:optimalw},
\begin{itemize}
    \item $r \aseq sv$ (i.e., $\vSigma_w\!=\!\bar{\vSigma}_{\beta}^{-1}$) is the optimal choice in $\cH_r$ that minimizes the bias function $\rb(r)$. Additionally, $r \aseq \bbE[v|s]\cdot s$ (i.e., $\vSigma_w\!=\!(f_v(\vSigma_x))^{-1}$) is the optimal in $\cS_r$ that minimizes $\rb(r)$.
    \item $r \aseq 1$ (i.e., $\vSigma_w\!=\!\vSigma_{x}$) is the optimal choice in both $\cS_r$ and $\cH_r$ that minimizes the variance $\rv(r)$.
\end{itemize}
\end{theo}
Theorem \ref{thm:optimalw_ridgeless} implies that the variance is minimized when $\vSigma_w=\vSigma_x$. 
Since the variance term does not depend on $\vbeta_{\star}$, it is not surprising that the optimal $\vSigma_w$ is also independent of $\vSigma_{\beta}$. 
Furthermore, this result is consistent with the intuition that to minimize the variance, $\hat{\beta}_{\lambda}$ should be penalized more in the higher variance directions of $\vSigma_x$, and vice versa. 
On the other hand, Theorem \ref{thm:optimalw_ridgeless} also implies that the bias is minimized when $\vd_w=1/\bar{\vd}_{\beta}$ which does not depend on $\vd_x$. 
While this characterization may not be intuitive, when $\bar{\vd}_{\beta}=\vd_{\beta}$ (i.e., $\vSigma_{\beta}$ also shares the same eigenvector matrix $\vU$), one analogy is that since the quadratic regularization corresponds to the a Gaussian prior $\cN(\boldsymbol{0}, \vSigma_w^{-1})$, it is reasonable to match $\vSigma_w^{-1}$ with the covariance of $\vbeta_{\star}$, which gives the maximum \text{a posteriori} (MAP) estimate. 
In general, the optimal $\vSigma_w$ admits a bias-variance tradeoff (i.e., the bias and variance are optimal under different $\vSigma_w$) except for the special case of $\vSigma_x\bar{\vSigma}_{\beta}=\vI$.  

Additionally, the following proposition demonstrates the advantage of the minimum $\|\hvb\|_{\vSigma_{w}}$ solution over the PCR estimator in the noiseless case.
\begin{prop}\label{cor:optimal_wPCR}
Under Assumption \ref{ass:eigenvector} and  \ref{ass:optimalw} and in the noiseless setting $\tilde{\sigma}=0$, suppose $s$ and $\bbE[v|s]\cdot s$ both have continuous and strictly increasing quantile functions. Then the minimum $\|\hvb\|_{\vSigma_{w}}$ solution outperforms the PCR estimator for all $\theta \in [0,1)$ when $\vSigma_w=\bar{\vSigma}_{\beta}^{-1}\in \cH_w$, or when $\vSigma_w=(f_v(\vSigma_x))^{-1}\in\cS_w$.
\end{prop}

\subsection{Optimal weighted ridge estimator}
Finally, we consider the optimally-tuned weighted ridge estimator ($\lambda=\lambda_{\opt}$) and discuss the optimal choice of weighting matrix $\vSigma_w$. 

\begin{theo}\label{thm:optimalw_optimal}
Suppose Assumptions \ref{ass:eigenvector} and \ref{ass:optimalw} hold. Then $r\aseq sv$ (i.e., $\vSigma_w\!=\!\bar{\vSigma}_{\beta}^{-1}$) is the optimal solution in $\cH_r$ that minimizes $\min_{\lambda} R(r, \lambda)$. Additionally, $r\aseq \bbE[v|s]\cdot s$ (i.e., $\vSigma_w\!=\!(f_v(\vSigma_x))^{-1}$) is the optimal solution in $\cS_r$ that minimizes $\min_{\lambda} R(r, \lambda)$.
\end{theo}
\vspace{-5pt}
\begin{figure}[!htb]
\centering
\begin{minipage}[t]{0.4\linewidth}
\centering
{\includegraphics[width=0.99\textwidth]{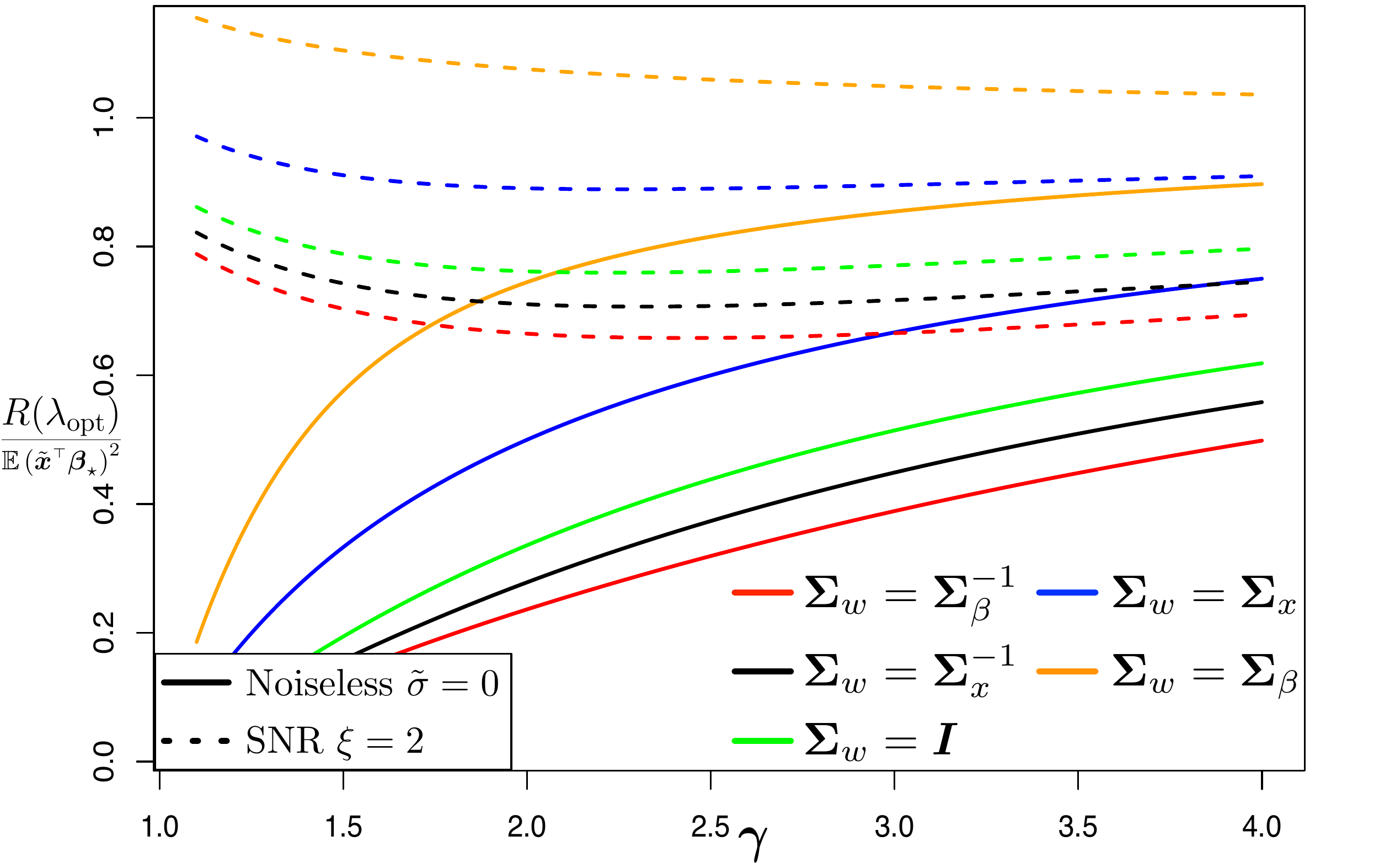}} \\ \vspace{-0.10cm}
\end{minipage}
\begin{minipage}[t]{0.395\linewidth}
\centering
{\includegraphics[width=0.99\textwidth]{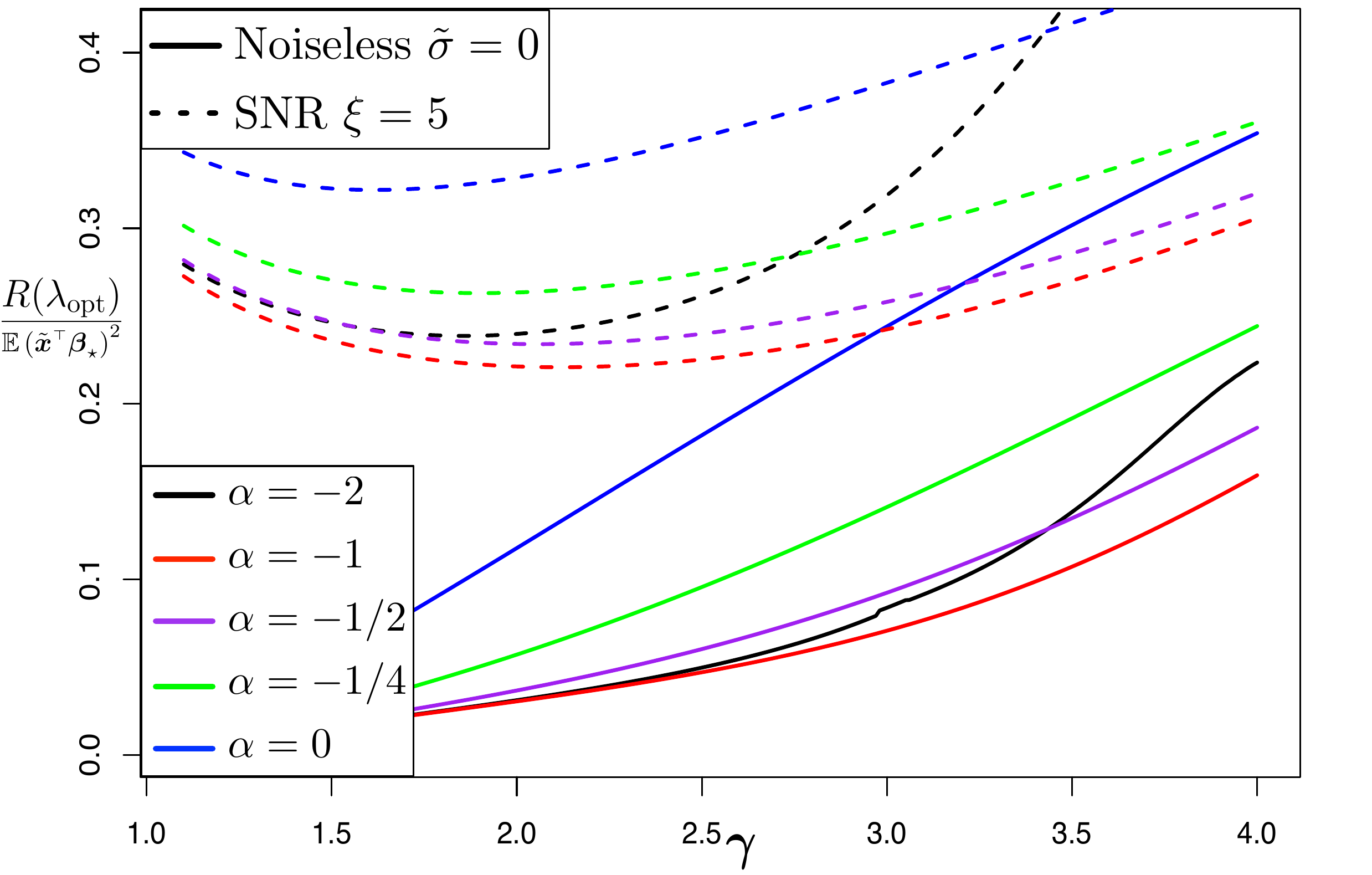}} \\ \vspace{-0.10cm}
\end{minipage}
\caption{\small $R(\lambda_{\opt})/\bbE(\tilde{\vx}^{\t}\vbeta_{\star})^2$ against $\gamma$ for various weighting matrix $\vSigma_w$. Solid lines represent the noiseless case $\tilde{\sigma}=0$ and the dashed lines represent the noisy case with fixed SNR $\xi$. We set $\vd_x$ to be aligned with $\vd_{\beta}$ and \textbf{Left}: $\vd_x$ to have 4 point masses $(1,2,3,4)$ with equal probabilities and $\vd_{\beta}$ with 2 point masses on $1$ and $5$ with probabilities $3/4$ and $1/4$, respectively; \textbf{Right}: $\vd_x$ has 2 point masses on $1$ and $5$ with probabilities $3/4$ and $1/4$, respectively, and $\vSigma_{\beta}=\vSigma_x^2$; we set $\vSigma_w=\vSigma_{\beta}^\alpha$.
}
\label{fig:optimal_w}
\end{figure} 

In contrast to the ridgeless setting in Theorem \ref{thm:optimalw_ridgeless}, the optimal $\vd_w$ for the optimally-tuned $\lambda_{\opt}$ does not depend on the noise level but only on $\bar{\vd}_{\beta}$, the strength of the signal in the directions of the eigenvectors of $\vSigma_x$. 
One interpretation is that in the optimally weighted estimator, $\lambda_{\opt}$ is capable of balancing the bias-variance tradeoff in the prediction risk; therefore the weighting matrix may not need to adjust to the label noise and can be chosen solely based on the signal $\vbeta_{\star}$. Indeed, as discussed in the previous section, $\vSigma_w=\vSigma_{\beta}^{-1}$ is a preferable choice of prior under the Bayesian perspective when $\vd_{\beta}=\bar{\vd}_{\beta}$.

Theorem \ref{thm:optimalw_optimal} is supported by Figure \ref{fig:optimal_w}, in which we plot the prediction risk of the generalized ridge regression estimator under different $\vSigma_w$ and optimally tuned $\lambda_{\opt}$.
We consider a simple discrete construction for aligned $\vd_x$ and $\vd_{\beta}(=\bar{\vd}_{\beta})$. On the left panel, we enumerate a few standard choices of $\vSigma_w$: $\vSigma_x,\vSigma_\beta,\vI, \vSigma_x^{-1}$ and the optimal choice $\vSigma_{\beta}^{-1}$. On the right, we take $\vSigma_w$ to be powers of $\vSigma_{\beta}$ around the optimal $\vSigma_{\beta}^{-1}$. In both setups, we confirm that $\vSigma_{\beta}^{-1}$ achieves the lowest risk uniformly over $\gamma$, as predicted by Theorem \ref{thm:optimalw_optimal}.  
 
\begin{figure}[!htb]
\centering
{\includegraphics[width=0.4\linewidth]{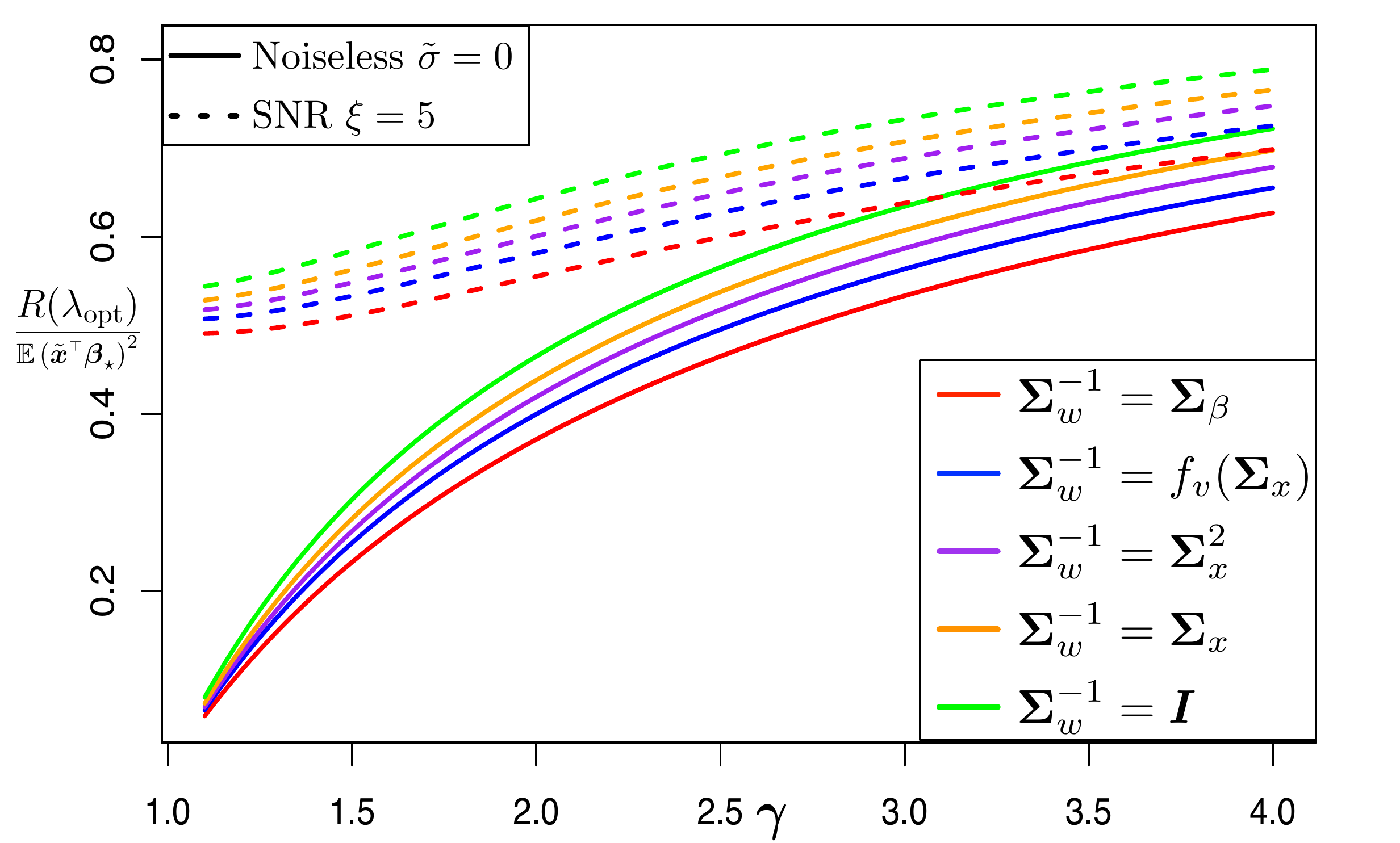}}
{\includegraphics[width=0.41\linewidth]{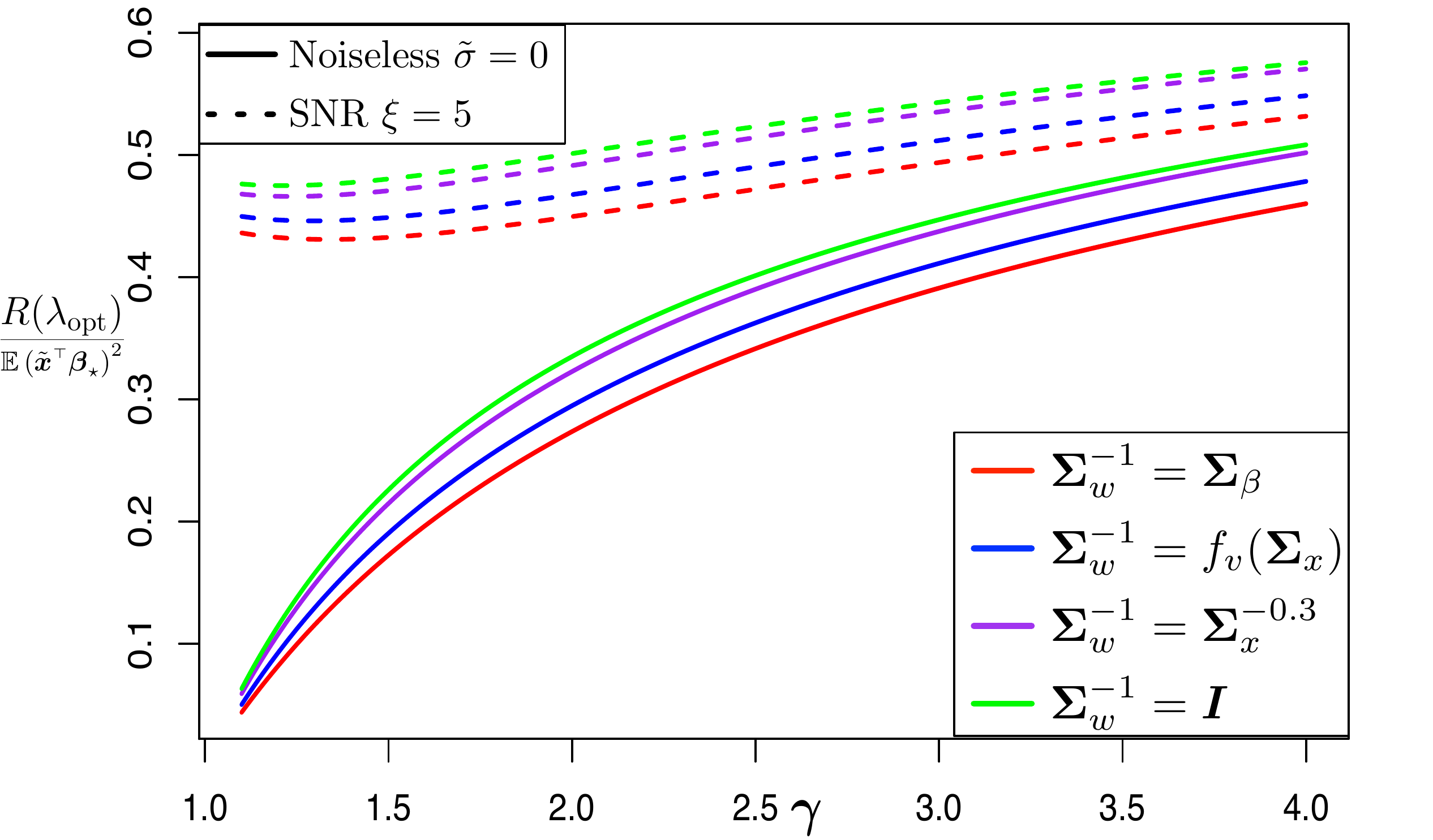}} \\ \vspace{-0.10cm}
\caption{\small $R(\lambda_{\opt})/\bbE(\tilde{\vx}^{\t}\vbeta_{\star})^2$ against $\gamma$ for various weighting matrix $\vSigma_w$ under noiseless $\tilde{\sigma}=0$ (solid lines) and noisy setting with fixed SNR $\xi$ (dashed lines). \textbf{Left}: We set $f_v(s)$ as an increasing function of $s$ on its support; \textbf{Right}: We set $f_v(s)$ as a decreasing function of $s$ on its support. 
Note that the heuristically chosen weighting matrices often outperform the standard ridge regression estimator (green).}
\vspace{-2mm}
\label{fig:optimal_egh}
\end{figure}

Note that our main results generally require knowledge of $\vSigma_x$ and $\bar{\vSigma}_{\beta}$. While $\vSigma_x$ can be estimated in a semi-supervised setting using unlabeled data (e.g., \cite{ryan2015semi,tony2020semisupervised}), it is typically difficult to estimate $\bar{\vSigma}_{\beta}$ directly from data. 
Without prior knowledge on $\bar{\vSigma}_{\beta}$, Theorem \ref{thm:optimalw_optimal} suggests that $r\aseq \bbE[v|s]\cdot s$ is the optimal $r$ that only depends on $s$. That is, $\vSigma_w = (f_v(\vSigma_x))^{-1}$ is the optimal $\vSigma_w$ that only depends on $\vSigma_x$. In the special case of $\bbE[v|s]=\bbE[v]$, the optimal $\vSigma_w$ in $\cS_w$ is equivalent to $\vSigma_w=\vI$ (standard ridge regression) due to the scale invariance. When the exact form of $f_v(s)$ is also not known, we may use a polynomial or power function of $s$ to approximate either $f_v(s)$ or $1/f_v(s)$, whose coefficients can be considered as hyper-parameters and cross-validated. 
We demonstrate the effectiveness of this heuristic in Figure \ref{fig:optimal_egh}: although our proposed $\vSigma_w=f_v(\vSigma_x)^{-1}$ (blue) is worse than the actual optimal (red) $\vSigma_w=\vSigma_{\beta}^{-1}$ (same as $\bar{\vSigma}_{\beta}^{-1}$ due to diagonal design), it is the best choice among weighting matrices that only depend on $\vSigma_x$.  
In addition, we seek the best approximation of $f_v(s)$ by applying a power transformation on $\vSigma_{x}$, and we observe that certain powers of $\vSigma_x$ also outperform the standard isotropic regularization.

\section{Conclusion}

\begin{wrapfigure}{R}{0.28\textwidth}  
\vspace{-6mm}
\centering 
\includegraphics[width=0.28\textwidth]{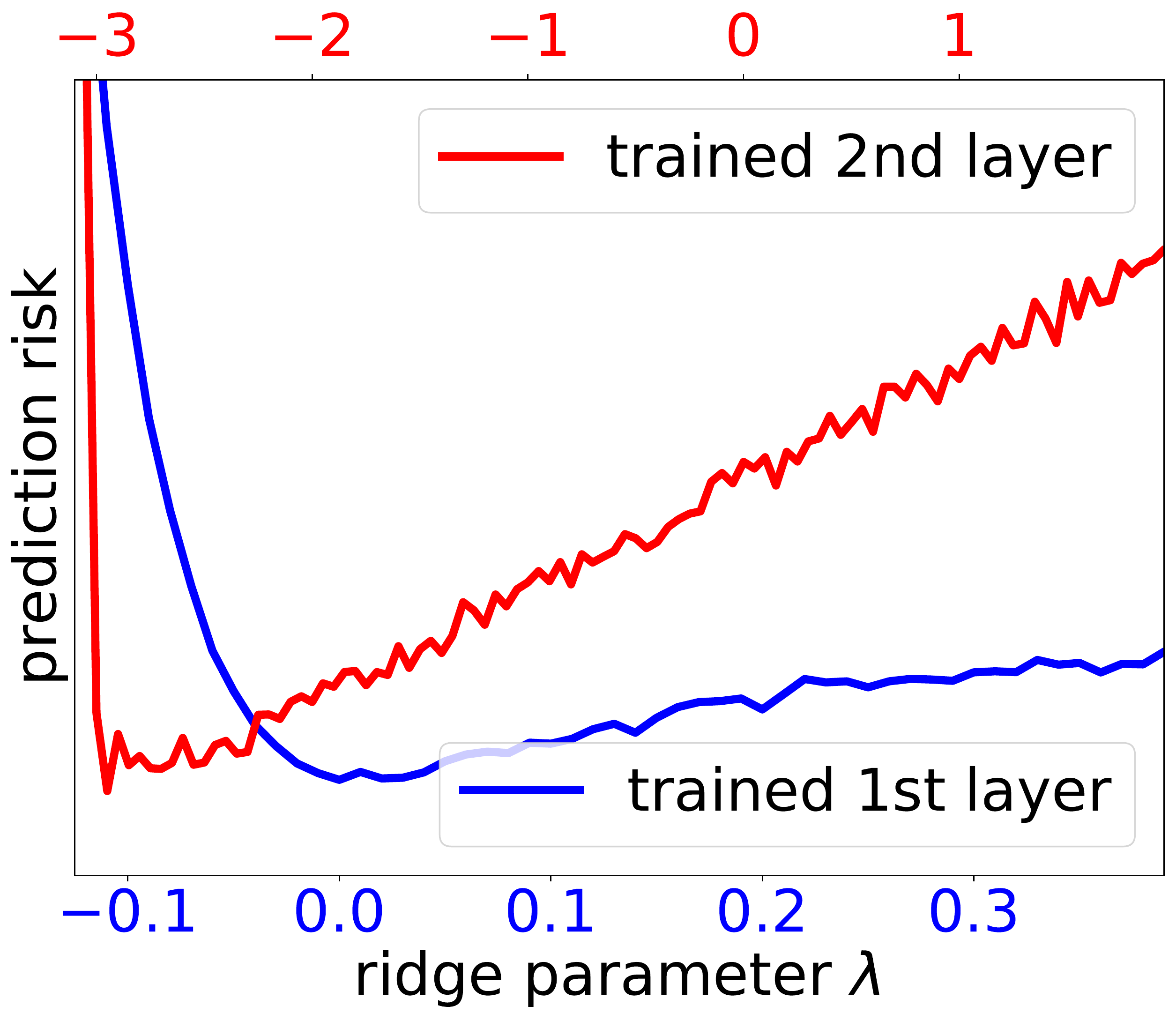}
\vspace{-6mm}
\caption{\small vanishing or negative regularization can be beneficial in certain two-layer neural nets.}  
\label{fig:negative-ridge-nn}
\vspace{-0.2cm} 
\end{wrapfigure} 

We provide a precise asymptotic characterization of the prediction risk of the generalized ridge regression estimator in the overparameterized regime. Our result greatly extends previous high-dimensional analysis of ridge regression, and enables us to discover and theoretically justify various interesting findings, including the negative ridge phenomenon, the implicit regularization of overparameterization, and a concise description of the optimal weighted ridge penalty.
We remark that some assumptions in our derivation may be further relaxed; for instance, the bounded eigenvalue assumption can be relaxed to certain polynomial decay (e.g., see \cite{xu2019number}). We also believe that similar findings can be observed in more complicated models, such as random features regression (see red line in Figure~\ref{fig:negative-ridge-nn}). 
Another fruitful direction is to construct weighting matrix $\vSigma_w$ solely from training data that outperforms isotropic shrinkage in the overparameterized regime. 

\bigskip
\subsection*{Acknowledgement}
{
The authors would like to thank Murat A.~Erdogdu, Daniel Hsu and Taiji Suzuki for comments and suggestions, and also anonymous NeurIPS reviewers 1 and 3 for helpful feedback. DW was partially funded by CIFAR, NSERC and LG Electronics. JX was supported by a Cheung-Kong Graduate School of Business Fellowship.

}

\bigskip
{
\bibliography{others/citation}
\bibliographystyle{amsalpha}

}
\newpage
\appendix
\allowdisplaybreaks


\section{Proofs omitted in Section \ref{sec:riskcal}}

\subsection{Proof of Theorem \ref{thm:RiskCal}}\label{sec:proofRiskCal}
We first claim that function $m(-\lambda)$ that satisfies \eqref{eq:relation_eq1} is indeed the Stieltjes transform of the limiting distribution of the eigenvalues of $\vX_{/w}\vX_{/w}^{\t}$. This is because the empirical distribution of the eigenvalues of $\vSigma_{x/w}$ converges to the distribution of $h$ due to Assumption \ref{ass:eigenvalues}. By the Marchenko-Pastur law, it is straightforward to show that the minimal eigenvalue of $\vX_{/w}\vX_{/w}^{\t}$ is lower bounded by $c_0$  as $n\to\infty$. Hence, we have $m(-\lambda)>0$ for all $\lambda>-c_0$. Then by taking derivatives of \eqref{eq:relation_eq1}\footnote{We can exchange expectation and derivatives because $\left|\frac{\partial \frac{h}{1+hm}}{\partial m}\right|=\frac{h}{(1+hm)^2} < \sup h$ when $m(-\lambda)>0$}, we know that \eqref{eq:relation_eq2} holds. The rest of the proof is to characterize Part 1 and Part 2 in \eqref{eq:risk_eq1} and show \eqref{eq:riskformula}.

For Part 1 in \eqref{eq:risk_eq1}, based on prior works \cite{dobriban2018high, xu2019number}, we have 
\begin{eqnarray}
\text{Part 1 in \eqref{eq:risk_eq1}}\ \ipe \ \tilde{\sigma}^2\frac{m'(-\lambda)}{m^2(-\lambda)}.
\end{eqnarray}
Hence, we only need to show that
\begin{eqnarray}
\text{Part 2 in \eqref{eq:risk_eq1}}\ \ipe \ \frac{m'(-\lambda)}{m^2(-\lambda)}\cdot\gamma\bbE\frac{gh}{(h\cdot m(-\lambda)+1)^2}.\label{eq:goal_calrisk}
\end{eqnarray}
Towards this goal, we first assume that $\vSigma_{w\beta}$ is invertible and
define $\vS=\vX_{/w^2\beta}^{\t}\vX_{/w^2\beta}+\lambda \vSigma_{w\beta}^{-1}$, where $\vX_{/w^2\beta}=\vX_{/w}\vSigma_{w\beta}^{-1/2}\sim \cN\left(0, \frac{1}{n}\vSigma_{x/w^2\beta}\right)$ and $\vSigma_{x/w^2\beta}=\vSigma_{w\beta}^{-1/2}\vSigma_{x/w}\vSigma_{w\beta}^{-1/2}$. Simplification of Part 2 yields
\[
\frac{\lambda^2}{n}\trace{\vSigma_{x/w^2\beta}\vS^{-2}}.
\]
To analyze the above quantity, we adopt the similar strategy used in \cite{ledoit2011eigenvectors} and first characterize a related quantity $\frac{1}{n}\trace{\vX_{/w^2\beta}^{\t}\vX_{/w\beta^2}}$. Note that, on one hand, we know that
\begin{eqnarray}
\frac{1}{n}\trace{\vS^{-2}\vX_{/w^2\beta}^{\t}\vX_{/w^2\beta}}&=&\frac{1}{n}\trace{\vS^{-1}-\lambda \vS^{-2}\vSigma_{w\beta}^{-1}}\n\\
&=&\frac{1}{n}\trace{\vSigma_{w\beta}(\vX_{/w}^{\t}\vX_{/w}+\lambda\vI)^{-1}-\lambda\vSigma_{w\beta}(\vX_{/w}^{\t}\vX_{/w}+\lambda\vI)^{-2}}.
\label{eq:trace_eq_1}
\end{eqnarray}
On the other hand, let $\vx_{/w^2\beta,i}$ be the $i$th row of $\vX_{/w^2\beta}$ and $\vS_{\backslash i}=\vS-\vx_{/w^2\beta,i}\vx_{/w^2\beta,i}^{\t}$, then we have
\begin{eqnarray}
\frac{1}{n}\trace{\vS^{-2}\vX_{/w^2\beta}^{\t}\vX_{/w^2\beta}}&=&\frac{1}{n}\sum_{i=1}^n\vx_{/w^2\beta,i}^{\t}\vS^{-2}\vx_{/w^2\beta,i}\ = \ \frac{1}{n}\sum_{i=1}^n\frac{\lambda^2\vx_{/w^2\beta,i}^{\t}\vS^{-2}_{\backslash i}\vx_{/w^2\beta,i}}{\left(\lambda+\lambda\vx_{/w^2\beta,i}^{\t}\vS^{-1}_{\backslash i}\vx_{/w^2\beta,i}\right)^2},
\label{eq:proof_eq1}
\end{eqnarray}
where the last equality holds due to the Matrix Inversion Lemma. Note that from Assumption \ref{ass:eigenvalues}, the eigenvalues of $\vSigma_{x/w}$ is lower bounded and upper bounded away from $0$ and $\infty$. Also, $\|\vSigma_{w\beta}\|$ is bounded away from $\infty$. Hence, by the Marchenko–Pastur law, we know that
\[
\left\|\lambda^2\vSigma_{x/w}^{1/2}\vSigma_{w\beta}^{-1/2}S_{\backslash i}^{-2}\vSigma_{w\beta}^{-1/2}\vSigma_{x/w}^{1/2}\right\|=\left\|\lambda^2\vSigma_{x/w}^{1/2}(\vX_{/w}^{\t}\vX_{/w}+\lambda\vI)^{-1}\vSigma_{w\beta}(\vX_{/w}^{\t}\vX_{/w}+\lambda\vI)^{-1}\vSigma_{x/w}^{1/2}\right\|
\]
and 
\[
\left\|\lambda\vSigma_{x/w}^{1/2}\vSigma_{w\beta}^{-1/2}S_{\backslash i}^{-1}\vSigma_{w\beta}^{-1/2}\vSigma_{x/w}^{1/2}\right\|=\left\|\lambda\vSigma_{x/w}^{1/2}(\vX_{/w}^{\t}\vX_{/w}+\lambda\vI)^{-1}\vSigma_{x/w}^{1/2}\right\|
\]
are upper bounded away from $\infty$ for any $\lambda>-c_0$\footnote{We take pseudo-inverse when $\lambda=0$}. Furthermore, observe that $\vx_{/w^2\beta,i}$ is independent of $\vS_{\backslash i}$ and\eqref{eq:proof_eq1}. Hence by Lemma 2.1 in \cite{ledoit2011eigenvectors}, we can show that
\begin{eqnarray}
\frac{1}{n}\trace{\vS^{-2}\vX_{/w^2\beta}^{\t}\vX_{/w^2\beta}}&\ipe&\frac{1}{n}\sum_{i=1}^n
\frac{\frac{\lambda^2}{n}\trace{\vSigma_{x/w^2\beta}\vS_{\backslash i}^{-2}}}{\left(\lambda+\frac{\lambda}{n}\trace{\vSigma_{x/w^2\beta}\vS_{\backslash i}^{-1}}\right)^2}, \  \forall \lambda>-c_0.\n
\end{eqnarray}
Next, we replace $\vS_{\backslash i}^{-1}$ by $\vS^{-1}$ and show the difference made by this rank-1 perturbation is negligible. From the Matrix Inversion Lemma, we have
\begin{eqnarray}
\lefteqn{\sup_i\left|\frac{\lambda}{n}\trace{\vSigma_{x/w^2\beta}\vS_{\backslash i}^{-1}}-\frac{\lambda}{n}\trace{\vSigma_{x/w^2\beta}\vS^{-1}}\right|}\n\\
&=&
\sup_i\frac{\lambda}{n}\frac{\vx_{/w^2\beta,i}^{\t}\vS_{\backslash i}^{-1}\vSigma_{x/w^2\beta}\vS_{\backslash i}^{-1}\vx_{/w^2\beta,i}}{1+\vx_{/w^2\beta,i}^{\t}\vS_{\backslash i}^{-1}\vx_{/w^2\beta,i}}\n\\
&\leq&\sup_i\frac{\lambda}{n}\frac{\|\vx_{/w^2\beta,i}^{\t}\vS_{\backslash i}^{-1}\vSigma_{x/w^2\beta}\vS_{\backslash i}^{-1/2}\|\cdot\| \vS_{\backslash i}^{-1/2}\vx_{/w^2\beta,i}\|}{1+\vx_{/w^2\beta,i}^{\t}\vS_{\backslash i}^{-1}\vx_{/w^2\beta,i}}\n\\
&\leq&\sup_i\frac{\lambda}{n}\frac{\|\vx_{/w^2\beta,i}^{\t}\vS_{\backslash i}^{-1/2}\|\cdot\|\vS_{\backslash i}^{-1/2}\vSigma_{x/w^2\beta}\vS_{\backslash i}^{-1/2}\|\cdot\| \vS_{\backslash i}^{-1/2}\vx_{/w^2\beta,i}\|}{1+\vx_{/w^2\beta,i}^{\t}\vS_{\backslash i}^{-1}\vx_{/w^2\beta,i}}\n\\
&\leq&\|\vSigma_{x/w}\|\cdot \sup_i\lambda\left\|\left(\vX_{/w}^{\t}\vX_{/w}+\lambda\vI-\vx_{/w,i}\vx_{/w,i}^{\t}\right)^{-1}\right\|\cdot \sup_i\frac{1}{n}\frac{\vx_{/w^2\beta,i}^{\t}\vS_{\backslash i}^{-1}\vx_{/w^2\beta,i}}{1+\vx_{/w^2\beta,i}^{\t}\vS_{\backslash i}^{-1}\vx_{/w^2\beta,i}} \n\\
&\leq& O_{\text{p}}\left(\frac{1}{n}\right).\n
\end{eqnarray}
Similarly,
\begin{eqnarray}
\lefteqn{\sup_i\left|\frac{\lambda^2}{n}\trace{\vSigma_{x/w^2\beta}\vS_{\backslash i}^{-2}}-\frac{\lambda^2}{n}\trace{\vSigma_{x/w^2\beta}\vS^{-2}}\right|}\n\\
&=&\sup_i\left|\frac{\lambda^2}{n}\trace{\vSigma_{x/w^2\beta}\vS_{\backslash i}^{-1}\left(\vS_{\backslash i}^{-1}-\vS^{-1}
\right)}\right|+\sup_i\left|\frac{\lambda^2}{n}\trace{\vSigma_{x/w^2\beta}\left(\vS_{\backslash i}^{-1}-\vS^{-1}
\right)\vS^{-1}}\right|\n\\
&\leq&\|\vSigma_{x/w}\|\cdot\|\vSigma_{w\beta}\|\left(\sup_i\lambda\left\|\left(\vX_{/w}^{\t}\vX_{/w}+\lambda\vI-\vx_{/w,i}\vx_{/w,i}^{\t}\right)^{-1}\right\|+\lambda\left\|\left(\vX_{/w}^{\t}\vX_{/w}+\lambda\vI\right)^{-1}\right\|\right)\n\\
&&\times \sup_i \lambda\left\|\left(\vX_{/w}^{\t}\vX_{/w}+\lambda\vI-\vx_{/w,i}\vx_{/w,i}^{\t}\right)^{-1}\right\|\cdot\sup_i\frac{1}{n}\frac{\vx_{/w^2\beta,i}^{\t}\vS_{\backslash i}^{-1}\vx_{/w^2\beta,i}}{1+\vx_{/w^2\beta,i}^{\t}\vS_{\backslash i}^{-1}\vx_{/w^2\beta,i}}\n\\
&=& O_{\text{p}}\left(\frac{1}{n}\right).\n
\end{eqnarray}
Hence, we have
\begin{eqnarray}
\frac{1}{n}\trace{\vS^{-2}\vX_{/w^2\beta}^{\t}\vX_{/w^2\beta}}&\ipe&
\frac{\frac{\lambda^2}{n}\trace{\vSigma_{x/w^2\beta}\vS^{-2}}}{\left(\lambda+\frac{\lambda}{n}\trace{\vSigma_{x/w^2\beta}\vS^{-1}}\right)^2}\n\\
&=&\frac{\frac{\lambda^2}{n}\trace{\vSigma_{x/w^2\beta}\vS^{-2}}}{\left(\lambda+\frac{\lambda}{n}\trace{\vD_{x/w}\left(\vX_{/w}^{\t}\vX_{/w}+\lambda\vI\right)^{-1}}\right)^2}\n\\
&\ipe&
\frac{\frac{\lambda^2}{n}\trace{\vSigma_{x/w^2\beta}\vS^{-2}}}{\left(\frac{1}{m(-\lambda)}\right)^2}, \quad \forall \lambda>-c_0,
\label{eq:trace_eq_2}
\end{eqnarray}
where the last equality used the following known results in \cite{ledoit2011eigenvectors, dobriban2018high, xu2019number}:
\[
    \frac{\lambda}{n}\trace{\vSigma_{x/w}\left(\vX_{/w}^{\t}\vX_{/w}+\lambda\vI\right)^{-1}}\ \ipe \ \frac{1}{m(-\lambda)} -\lambda.
\]
Combine \eqref{eq:trace_eq_1} and \eqref{eq:trace_eq_2}, we have
\begin{eqnarray}
    \text{Part 2}&\ipe& \frac{1}{m^2(-\lambda)}\cdot \frac{1}{n}\trace{\vSigma_{w\beta}(\vX_{/w}^{\t}\vX_{/w}+\lambda\vI)^{-1}-\lambda\vSigma_{w\beta}(\vX_{/w}^{\t}\vX_{/w}+\lambda\vI)^{-2}}\quad \forall \lambda>-c_0.\n
\end{eqnarray}
Our next step is to characterize $\frac{1}{n}\trace{\vSigma_{w\beta}(\vX_{/w}^{\t}\vX_{/w}+\lambda\vI)^{-1}}$. From Theorem 1 in \cite{rubio2011spectral}, for any deterministic sequence of matrices $\vTheta_n$ such that $\frac{1}{n}\trace{\left(\vTheta_n^{\t}\vTheta_n\right)^{1/2}}$ is finite, we know that as $n,p\rightarrow \infty$,
\[
\frac{1}{n}\trace{\vTheta_n\left(\vX_{/w}^{\t}\vX_{/w}-z\vI\right)^{-1}}\ase \frac{1}{n}\trace{\vTheta_n\left(c_n(z)\vSigma_{x/w}-z\vI\right)^{-1}}, \quad \forall z\in \bbC^{+}-\bbR^{+},
\]
where $c_n(z)$ satisfies
\[
c_n(z) = 1 - \gamma \bbE \frac{h_n c_n(z)}{h_n c_n(z)-z}, 
\]
and $h_n$ follows the empirical distribution of $\vD_{x/w}$. Hence, it is clear that $c_{n}(z)\rightarrow -zm(z)$ for all $z\in \bbC^{+}-\bbR^{+}$ due to \eqref{eq:relation_eq1} and the dominated convergence theorem. Now let $\vTheta_n=\vSigma_{w\beta}$. Since $\vSigma_{w\beta}$ is a positive semi-definite matrix, we have
\[
\frac{1}{n}\trace{\left(\vSigma_{w\beta}^{\t}\vSigma_{w\beta}\right)^{1/2}}\ =\ \frac{1}{n}\trace{\vSigma_{w\beta}} \ \leq \ \frac{d}{n}c_{u}<\infty.
\]
Therefore, applying Theorem 1 in \cite{rubio2011spectral} yields
\begin{eqnarray}
\frac{1}{n}\trace{\vSigma_{w\beta}\left(\vX_{/w}^{\t}\vX_{/w}-z\vI\right)^{-1}}&\ase& \frac{1}{n}\trace{\vSigma_{w\beta}\left(-zm(z)\cdot \vSigma_{x/w}-z\vI\right)^{-1}}\n\\
&=& \frac{1}{n}\trace{\vU_{x/w}\vSigma_{w\beta}\vU_{x/w}^{\t}\left(-zm(z)\cdot \vD_{x/w}-z\vI\right)^{-1}}\n\\
&=&\frac{1}{-zn}\sum_{i=1}^d\frac{d_{w\beta, i}}{d_{x/w,i}m(z)+1}, \quad \forall z\in \bbC^{+}-\bbR^{+}.\n
\end{eqnarray}
From Assumption \ref{ass:eigenvalues} and dominated convergence theorem, we have
\begin{eqnarray}
\frac{\lambda}{n}\trace{\vSigma_{w\beta}\left(\vX_{/w}^{\t}\vX_{/w}+\lambda\vI\right)^{-1}}\ase \gamma\bbE\frac{g}{h\cdot m(-\lambda)+1}, \quad \forall -\lambda\in \bbC^{+}-\bbR^{+}.\label{eq:riskproof_eqC}
\end{eqnarray}
Note that both $\vSigma_{w\beta}$ and $\left(\vX_{/w}^{\t}\vX_{/w}+\lambda\vI\right)^{-1}$ are positive semi-definite matrices, and thus
\begin{eqnarray}
\frac{\lambda}{n}\trace{\vSigma_{w\beta}\left(\vX_{/w}^{\t}\vX_{/w}+\lambda\vI\right)^{-1}} &\leq&\lambda\left\|\left(\vX_{/w}^{\t}\vX_{/w}+\lambda\vI\right)^{-1}\right\|\cdot \frac{1}{n}\trace{\vSigma_{w\beta}}\n\\
&\leq&\lambda\left\|\left(\vX_{/w}^{\t}\vX_{/w}+\lambda\vI\right)^{-1}\right\|\cdot \frac{d}{n}c_u.\n
\end{eqnarray}
Hence $\frac{\lambda}{n}\trace{\vSigma_{w\beta}\left(\vX_{/w}^{\t}\vX_{/w}+\lambda\vI\right)^{-1}}$ is bounded on $\lambda>-c_0$; by the dominated convergence theorem, we can extend \eqref{eq:riskproof_eqC} to $\lambda>-c_0$ and conclude that 
\[
\frac{\lambda}{n}\trace{\vSigma_{w\beta}(\vX_{/w}^{\t}\vX_{/w}+\lambda\vI)^{-1}}\ \ase \ \gamma\bbE \frac{g}{h\cdot m(-\lambda) + 1}, \quad \forall \lambda>-c_0. 
\]
It is straightforward to check $\frac{1}{n}\trace{\vSigma_{w\beta}(\vX_{/w}^{\t}\vX_{/w}+\lambda\vI)^{-2}}$ is bounded as well. With arguments similar to \cite{dobriban2018high} and \cite{hastie2019surprises}, we have
\[
\frac{1}{n}\trace{\vSigma_{w\beta}(\vX_{/w}^{\t}\vX_{/w}+\lambda\vI)^{-2}}\ = \ -\frac{\partial \frac{1}{n}\trace{\vSigma_{w\beta}(\vX_{/w}^{\t}\vX_{/w}+\lambda\vI)^{-1}}}{\partial \lambda}.
\]
We therefore arrive at the desired result
\[
\frac{\lambda}{n}\trace{\vSigma_{w\beta}(\vX_{/w}^{\t}\vX_{/w}+\lambda\vI)^{-2}}\ \ase \ \frac{\gamma}{\lambda}\bbE \frac{g}{h\cdot m(-\lambda) + 1} -\gamma \bbE \frac{g\cdot m'(-\lambda)}{(h\cdot m(-\lambda) + 1)^2}.
\]

Combining the above calculations, we know \eqref{eq:goal_calrisk} holds when $\vSigma_{w\beta}$ is invertible. 
Finally, we extend \eqref{eq:goal_calrisk} to the case when $\vSigma_{w\beta}$ is not invertible. For any $\epsilon>0$, we let $\vSigma_{w\beta}^{\epsilon}=\vSigma_{w\beta}+\epsilon \vI$. Then, from the above analysis, we have 
\[
\frac{\lambda^2}{n}\trace{\vSigma_{x/w}\left(\vX_{/w}^{\t}\vX_{/w}+\lambda\vI\right)^{-1}\vSigma_{w\beta}^{\epsilon}\left(\vX_{/w}^{\t}\vX_{/w}+\lambda\vI\right)^{-1}}\ase \frac{m'(-\lambda)}{m^2(-\lambda)}\cdot\gamma\bbE\frac{(g+\epsilon)h}{(h\cdot m(-\lambda)+1)^2}.
\]
Note that the LHS of above equation is decreasing as $\epsilon$ decreases to $0$ and the RHS of above equation is always bounded for any $\epsilon<1$. Hence by the dominated convergence theorem, we know that \eqref{eq:goal_calrisk} holds for non-invertible $\vSigma_{w\beta}$ as well.


\subsection{Proof of Corollary \ref{cor:PCRrisk}} \label{app:cor:PCRrisk}

We only provide the proof for the overparameterized regime when $\theta\gamma>1$, because the calculation is straightforward when $\theta\gamma<1$ (see \cite{xu2019number}). Since $h$ has continuous strictly increasing quantile function $Q_h$, we know that the $1-\theta$ quantile of $\vd_{x/w}$ (which is the threshold of top $\theta p$ elements of $\vd_{x/w}$) converges to $Q_h(1-\theta)$. Therefore, the empirical distribution of the top $\theta p$ elements of $\vd_{x/w}$ and the corresponding $\vd_{w\beta}$ jointly converges to the conditional distribution of $(h, g)$ given $h\geq Q_h(1-\theta)$. Hence, we can apply Theorem \ref{thm:RiskCal} and obtain that
\[
\tilde{\bbE}\left(\tilde{y}-\tilde{x}\hat{\vbeta}_{\theta}\right)^2
\ \ipe\
\frac{m'_{\theta}(0)}{m^2_{\theta}(0)}\cdot\left(\gamma\theta\bbE\left[\frac{gh}{(h\cdot m_{\theta}(0)+1)^2}\big|h\geq Q_h(1-\theta)\right]+\gamma\bbE gh\bbI_{h<Q_h(1-\theta)}+\tilde{\sigma}^2\right).
\]
Here the extra term $\gamma\bbE gh\bbI_{h<Q_h(1-\theta)}$ comes from the ``misspecification'' by dropping the small $(1-\theta)p$ number of eigenvalues, and $m_{\theta}(z)$ should satisfy that
\[
-z \ = \ \frac{1}{m_{\theta}(z)} - \gamma \theta \bbE\left[\frac{h}{1+h\cdot m_{\theta}(z)}\big|h\geq Q_h(1-\theta)\right].
\]
By replacing the conditional expectation with the normal expectation, we complete the calculation of the asymptotic prediction risk in Corollary \ref{cor:PCRrisk}.

Next, when $\bbE[g|h]$ is a decreasing function of $h$ and $h$ has continuous p.d.f.~denoted by $f(h)$ (in this proof), we show that the asymptotic prediction risk $\frac{m'_{\theta}(0)}{m^2_{\theta}(0)}\cdot\left(\gamma\bbE\left[\frac{gh}{(h_{\theta}\cdot m_{\theta}(0)+1)^2}\right]+\tilde{\sigma}^2\right)\triangleq R_{\theta}$ is a decreasing function of $\theta$. 
Let $q_{\theta}$ and $m_{\theta}$ be the shorthand for $Q_h(1-\theta)$ and $m_{\theta}(0)$ respectively. Because $Q_h$ is a strictly increasing continuous function and $h$ has continuous p.d.f., we know that $\frac{\partial q_{\theta}}{\partial \theta}$ exists and is negative. Hence, by the chain rule, we only need to show that
$\frac{\partial R_{\theta}}{\partial q_{\theta}}>0$, which is equivalent to 
\begin{eqnarray}
0&<&\left(-\frac{\bbE[g|h]hf(h)}{(hm_{\theta}+1)^2}\Big|_{h=q_{\theta}}+\bbE[g|h]hf(h)\Big|_{h=q_{\theta}}-2\bbE\frac{gh_{\theta}^2}{(h_{\theta}m_{\theta}+1)^3}\cdot \frac{\partial m_{\theta}}{\partial q_{\theta}}\right)\bbE\frac{h_{\theta}m_{\theta}}{(h_{\theta}m_{\theta}+1)^2}\n\\
&&-\left(\bbE\frac{gh}{(h_{\theta}m_{\theta}+1)^2}+\tilde{\sigma}^2\right)\left(\frac{h^2m_{\theta}^2f(h)}{(hm_{\theta}+1)^2}\Big|_{h=q_{\theta}}-2\bbE\frac{h_{\theta}^2m_{\theta}}{(h_{\theta}m_{\theta}+1)^3}\cdot \frac{\partial m_{\theta}}{\partial q_{\theta}}\right),\label{eq:corPCRrisk_eq1}
\end{eqnarray}
where we use the fact that  
\[
\frac{m'_{\theta}(0)}{m^2_{\theta}(0)}=\left(1-\gamma \bbE \left(\frac{h_{\theta}m_{\theta}}{h_{\theta}m_{\theta}+1}\right)^2\right)^{-1}\quad \text{and} \quad 1=\gamma \bbE\frac{h_{\theta}m_{\theta}}{1+h_{\theta}m_{\theta}}.
\]
We simplify the RHS of \eqref{eq:corPCRrisk_eq1} by breaking it into three parts:
\begin{eqnarray}
\text{RHS of \eqref{eq:corPCRrisk_eq1}}&=&\underbrace{\left(\bbE[g|h]f(h)\frac{h^2m_{\theta}^2(2+hm_{\theta})}{(hm_{\theta}+1)^2}\Big|_{h=q_{\theta}}\right)\bbE\frac{h_{\theta}}{(h_{\theta}m_{\theta}+1)^2}-\left(\frac{h^2m_{\theta}^2f(h)}{(hm_{\theta}+1)^2}\Big|_{h=q_{\theta}}\right)\bbE\frac{\bbE[g|h]h_{\theta}}{(h_{\theta}m_{\theta}+1)^2} }_{\text{part (i)}}\n\\
&&+\underbrace{\frac{2}{m_{\theta}^2}\frac{\partial m_{\theta}}{\partial q_{\theta}} \cdot \left(\bbE\frac{h_{\theta}^2m_{\theta}^2}{(h_{\theta}m_{\theta}+1)^3}\bbE\frac{gh_{\theta}m_{\theta}}{(h_{\theta}m_{\theta}+1)^2}- \bbE\frac{gh_{\theta}^2m_{\theta}^2}{(h_{\theta}m_{\theta}+1)^3}\bbE\frac{h_{\theta}m_{\theta}}{(h_{\theta}m_{\theta}+1)^2} \right)}_{\text{part (ii)}}\n\\
&&+\underbrace{\left(\bbE gh\bbI_{h<q_{\theta}}+\tilde{\sigma}^2\right)\left(2\bbE\frac{h_{\theta}^2m_{\theta}}{(h_{\theta}m_{\theta}+1)^3}\cdot \frac{\partial m_{\theta}}{\partial q_{\theta}}-\frac{h^2m_{\theta}^2f(h)}{(hm_{\theta}+1)^2}\Big|_{h=q_{\theta}}\right)}_{\text{part (iii)}}\label{eq:corPCRrisk_eq2}
\end{eqnarray}
To show part (i) is positive, note that since $\bbE[g|h]$ is a decreasing function of $h$, we have 
\[
\bbE\frac{\bbE[g|h]h_{\theta}}{(h_{\theta}m_{\theta}+1)^2}\ \leq\ \bbE[g|h]\Big|_{h=q_{\theta}}\cdot \bbE \frac{h_{\theta}}{(h_{\theta}m_{\theta}+1)^2},\]
Therefore,
\[
\text{part (i)} \ \geq\ \bbE[g|h]\Big|_{h=q_{\theta}}\cdot \bbE \frac{h_{\theta}}{(h_{\theta}m_{\theta}+1)^2}\cdot\left( \frac{h^2m_{\theta}^2f(h)}{(hm_{\theta}+1)}\Big|_{h=q_{\theta}}\right)
\]
Hence part (i) is positive because $q_{\theta}<\sup h$.

To show that part (ii) is non-negative, observe that by taking derivatives with respect to $q_{\theta}$ on both sides of $1=\gamma \bbE \frac{h_{\theta}m_{\theta}}{h_{\theta}m_{\theta}+1}$, we have
\begin{eqnarray}
\bbE \frac{h_{\theta}}{(h_{\theta}m_{\theta}+1)^2}\cdot\frac{\partial m_{\theta}}{\partial q_{\theta}} &=&\frac{hm_{\theta}f(h)}{hm_{\theta}+1}\Big|_{h=q_{\theta}}. \label{eq:PCRmprime}
\end{eqnarray}
Hence, we know that $\frac{\partial m_{\theta}}{\partial q_{\theta}}>0$. What remains is to show that
\begin{eqnarray}
\bbE\frac{h_{\theta}^2m_{\theta}^3}{(h_{\theta}m_{\theta}+1)^2}\bbE\frac{\bbE[g|h]h_{\theta}m_{\theta}}{(h_{\theta}m_{\theta}+1)^2}\ \geq\ \bbE\frac{\bbE[g|h]h_{\theta}^2m_{\theta}^2}{(h_{\theta}m_{\theta}+1)^3}\bbE\frac{h_{\theta}m_{\theta}}{(h_{\theta}m_{\theta}+1)^2}.\label{eq:corPCRrisk_eq3}
\end{eqnarray}

Denote the probability measure of $h$ as $\mu$ and let $\tilde{\mu}$ be the new measure of $h_{\theta}m_{\theta}\mu\cdot \Large[(h_{\theta}m_{\theta}+1)^2\bbE\frac{h_{\theta}m_{\theta}}{(h_{\theta}m_{\theta}+1)^2}\Large]^{-1}$. Let $\tilde{h}$ be a random variable following the new measure $\tilde{\mu}$ and $\tilde{h}_{\theta}=\tilde{h}\bbI_{\tilde{h}\geq q_{\theta}}$. 
Then since $\frac{\tilde{h}_{\theta}m_{\theta}}{\tilde{h}_{\theta}m_{\theta}+1}$ is an increasing function of $\tilde{h}$ and $\bbE[g|h=\tilde{h}]$ is a decreasing function of $\tilde{h}$, we have
\[
    \bbE\frac{\tilde{h}_{\theta}m_{\theta}}{\tilde{h}_{\theta}m_{\theta}+1}\cdot \bbE\left(\bbE[g|h=\tilde{h}]\right) \geq  \bbE\frac{\tilde{h}_{\theta}m_{\theta}\bbE[g|h=\tilde{h}]}{\tilde{h}_{\theta}m_{\theta}+1}.
\]
We then change $\tilde{h}$ back to $h$ and obtain that \eqref{eq:corPCRrisk_eq3} holds. We therefore conclude that part (ii) is non-negative.

To show that part (iii) is non-negative, we only need to confirm that
\[
2\bbE\frac{h_{\theta}^2m_{\theta}}{(h_{\theta}m_{\theta}+1)^3}\cdot \frac{\partial m_{\theta}}{\partial q_{\theta}}\ \geq\ \frac{h^2m_{\theta}^2f(h)}{(hm_{\theta}+1)^2}\Big|_{h=q_{\theta}}
\]
From \eqref{eq:PCRmprime}, this is equivalent to
\[
2\bbE\frac{h_{\theta}^2m_{\theta}}{(h_{\theta}m_{\theta}+1)^3}\cdot \frac{hm_{\theta}f(h)}{hm_{\theta}+1}\Big|_{h=q_{\theta}} \ \geq\ \frac{h^2m_{\theta}^2f(h)}{(hm_{\theta}+1)^2}\Big|_{h=q_{\theta}} \cdot \bbE\frac{h_{\theta}}{(h_{\theta}m_{\theta}+1)^2},
\]
which is then equivalent to
\[
2\bbE\frac{h_{\theta}^2}{(h_{\theta}m_{\theta}+1)^3} \ \geq \ \frac{h}{(hm_{\theta}+1)}\Big|_{h=q_{\theta}} \cdot \bbE\frac{h_{\theta}}{(h_{\theta}m_{\theta}+1)^2}.
\]
The above equation clearly holds because $\frac{h}{hm_{\theta}+1}$ is an increasing function of $h$.

The proof of Corollary \ref{cor:PCRrisk} is completed by combining the above calculations.


\section{Proofs omitted in Section \ref{sec:sgnopt}}

\subsection{Optimal $\lambda_{\opt}$ for simple cases}\label{sec:simplify}
When $h \aseq c$, then $\zeta=h\cdot m(-\lambda)$ is a single point mass at $c\cdot m(-\lambda)$. Thus \eqref{eq:derivative} achieves $0$ is equivalent to
\[
    \bbE [g]\cdot \left(1-\gamma\frac{\zeta^2}{(1+\zeta)^2}\right)-\gamma \frac{\zeta}{(1+\zeta)^2}\bbE [g] -\tilde{\sigma}^2m(-\lambda)\ =\ 0.
\]
which is also equivalent to
\begin{eqnarray}
\frac{\bbE [g]}{\tilde{\sigma}^2}\left(1-\gamma\frac{\zeta}{(1+\zeta)}\right) \ = \ m(-\lambda).\label{eq:simplify_eq1}
\end{eqnarray}
Note that \eqref{eq:relation_eq1} is now simplified to
\[
1 = \lambda m(-\lambda) + \gamma \frac{\zeta}{1+\zeta}.
\]
Furthermore, under Assumption \ref{ass:eigenvalues}, the SNR can be simplified to
\[
\xi= \frac{c\bbE [g]}{\tilde{\sigma}^2}.
\]
Plug the above calculations into \eqref{eq:simplify_eq1}, we have
\[
\lambda_{\opt} = \frac{\tilde{\sigma}^2}{\bbE [g]} = \frac{c}{\xi}.
\]

On the other hand, when $g\aseq c$, then \eqref{eq:derivative} achieves $0$ is equivalent to
\[
    c\bbE\frac{\zeta^2}{(1+\zeta)^3}\cdot \left(1-\gamma\bbE\frac{\zeta^2}{(1+\zeta)^2}\right)-c\gamma \bbE\frac{\zeta^2}{(1+\zeta)^3}\bbE \frac{\zeta}{(1+\zeta)^2} -\tilde{\sigma}^2m(-\lambda)\bbE\frac{\zeta^2}{(1+\zeta)^3}\ =\ 0,
\]
which is equivalent to
\[
c\left(1-\gamma\bbE\frac{\zeta}{1+\zeta}\right) -\tilde{\sigma}^2m(-\lambda)\ =\ 0.
\]
Plug \eqref{eq:relation_eq1} in above equation, we recover
\[
\lambda_{\opt} = \frac{\tilde{\sigma}^2}{c}.
\]

Finally when $\bbE[g|h]\aseq \bbE g$, then \eqref{eq:derivative} achieving $0$ is equivalent to
\[
    \bbE g\cdot \bbE\frac{\zeta^2}{(1+\zeta)^3}\cdot \left(1-\gamma\bbE\frac{\zeta^2}{(1+\zeta)^2}\right)-\bbE g\cdot \gamma \bbE\frac{\zeta^2}{(1+\zeta)^3}\bbE \frac{\zeta}{(1+\zeta)^2} -\tilde{\sigma}^2m(-\lambda)\bbE\frac{\zeta^2}{(1+\zeta)^3}\ =\ 0,
\]
which is equivalent to
\[
\bbE g\cdot \left(1-\gamma\bbE\frac{\zeta}{1+\zeta}\right) -\tilde{\sigma}^2m(-\lambda)\ =\ 0.
\]
Plug \eqref{eq:relation_eq1} in above equation yields the desired result
\[
\lambda_{\opt} = \frac{\tilde{\sigma}^2}{\bbE[g]}.
\]

\subsection{Proof of Theorem \ref{thm:optimalridge}}\label{sec:proofoptimalridge}

Let $\zeta=h\cdot m(-\lambda)$. Taking derivatives of \eqref{eq:relation_eq2} with respect to $\lambda$ on both sides, we have
\begin{eqnarray}
m''(-\lambda)&=&2\frac{1-\gamma\bbE \frac{\zeta^3}{(1+\zeta)^3}}{1-\gamma\bbE\frac{\zeta^2}{(1+\zeta)^2}}\cdot \frac{(m'(-\lambda))^2}{m(-\lambda)}. \label{eq:relation_m''}
\end{eqnarray}
Also, rearranging \eqref{eq:relation_eq1} and \eqref{eq:relation_eq2} yields
\begin{eqnarray}
\lambda m(-\lambda)&=&1-\gamma\bbE \frac{\zeta}{1+\zeta},\label{eq:relation_m}\\
m'(-\lambda)&=&\left(1-\gamma\bbE\frac{\zeta^2}{(\zeta+1)^2}\right)^{-1}m^2(-\lambda).\label{eq:relation_m'}
\end{eqnarray}
By \eqref{eq:relation_m''}-\eqref{eq:relation_m'}, we have
\begin{eqnarray}
\frac{\dif \frac{m'(-\lambda)}{m^2(-\lambda)}}{\dif \lambda}&=&\frac{-m''(-\lambda)m(-\lambda)+2(m'(-\lambda))^2}{m^3(-\lambda)}\nonumber\\
&=&-\frac{2\gamma(m'(-\lambda))^2}{m^3(-\lambda)} \cdot \frac{1}{1-\gamma\bbE\frac{\zeta^2}{(1+\zeta)^2}}\cdot \bbE \frac{\zeta^2}{(1+\zeta)^3}.\label{eq:derivative_eq1}
\end{eqnarray}
Hence with \eqref{eq:derivative_eq1}, it is straightforward to obtain \eqref{eq:derivative}. In addition, note that $m(-\lambda), m'(-\lambda)>0$ for all $\lambda>-c_0$. Therefore, from \eqref{eq:relation_m'}, we know that $1-\gamma\bbE\frac{\zeta^2}{(1+\zeta)^2}>0$ and thus Part 3 is always negative for all $\lambda>-c_0$.

Next, we analyze the sign of Part 4. Note that 
\begin{eqnarray}
\text{Part 4 }\gtreqless\ 0 &\Leftrightarrow& \left(\frac{1}{\gamma}-\bbE\frac{\zeta^2}{(1+\zeta)^2}\right)\bbE\frac{gh\zeta}{(\zeta+1)^3}\ \gtreqless \ \bbE \frac{\zeta^2}{(1+\zeta)^3}\bbE\frac{gh}{(1+\zeta)^2}\n\\
&\Leftrightarrow& \left(\frac{\lambda m(-\lambda)}{\gamma}+\bbE\frac{\zeta}{(1+\zeta)^2}\right)\bbE\frac{g\zeta^2}{(\zeta+1)^3}\ \gtreqless \ \bbE \frac{\zeta^2}{(1+\zeta)^3}\bbE\frac{g\zeta}{(1+\zeta)^2},\n\\
\label{eq:derivative_eq2}
\end{eqnarray}
where the last equivalence holds due to \eqref{eq:relation_m} and $m(-\lambda)>0$ for all $\lambda>-c_0$. Denote the probability measure of $h$ as $\mu(h)$. We introduce a new probability measure $\tilde{\mu}(h) = \frac{\zeta \mu(h)}{(1+\zeta)^2\bbE\frac{\zeta}{(1+\zeta)^2}}$. Let $\tilde{h}$ follow this new measure $\tilde{\mu}$ and $\tilde{\zeta}=\tilde{h}\cdot m(-\lambda)$. In addition define $f(h)=\bbE[g|h]$.
\begin{itemize}
\item When $\bbE[g|h]=f(h)$ is an increasing function of $h$, then for any fixed $m(-\lambda)>0$, we have
\[
    \bbE \frac{f(\tilde{h})\tilde{\zeta}}{1+\tilde{\zeta}}\geq \bbE \frac{\tilde{\zeta}}{1+\tilde{\zeta}}\bbE f(\tilde{h}),
\]
because both $\frac{\tilde{\zeta}}{1+\tilde{\zeta}}$ and $f(\tilde{h})$ are increasing function of $\tilde{h}$. 
Then we change $\tilde{h}$ back to $h$ and obtain
\[
    \bbE\frac{\bbE[g|h] \zeta^2}{(\zeta+1)^3} \bbE\frac{\zeta}{(\zeta+1)^2}\geq \bbE \frac{\zeta^2}{(1+\zeta)^3}\bbE \frac{\bbE[g|h] \zeta}{(\zeta+1)^2}.
\]
Hence, for all $\lambda>0$, we know that Part 4 is positive; at $\lambda=0$, Part 4 is non-negative. Moreover, the equality in above equation is only achieved when $\bbE[g|h]$ is constant almost surely or $h$ is constant almost surely, which is equivalent to $\bbE[g|h]\aseq \bbE[g]$. Hence, Part 4 is $0$ at $\lambda=0$ only when $\bbE[g|h]\aseq\bbE[g]$. 

\item When $\bbE[g|h]=f(h)$ is a decreasing function of $h$, then for any fixed $m(-\lambda)>0$, we have
\[
    \bbE \frac{f(\tilde{h})\tilde{\zeta}}{1+\tilde{\zeta}}\leq \bbE \frac{\tilde{\zeta}}{1+\tilde{\zeta}}\bbE f(\tilde{h}),
\]
due to the fact that $\frac{\tilde{\zeta}}{1+\tilde{\zeta}}$ and $f(\tilde{h})$ have different monotonicity w.r.t. $\tilde{h}$. Replacing $\tilde{h}$ with $h$, we arrive at
\[
    \bbE\frac{\bbE[g|h] \zeta^2}{(\zeta+1)^3} \bbE\frac{\zeta}{(\zeta+1)^2}\leq \bbE \frac{\zeta^2}{(1+\zeta)^3}\bbE \frac{\bbE[g|h] \zeta}{(\zeta+1)^2}.
\]
Hence, for all $\lambda <0$, we know that Part 4 is negative; at $\lambda=0$, Part 4 is non-positive. Similarly, Part 4 is $0$ at $\lambda=0$ only when $\bbE[g|h]\aseq\bbE[g]$. 
\end{itemize}
This completes the proof of Theorem \ref{thm:optimalridge}.


\subsection{Proof of Proposition \ref{thm:noiselevel}}\label{sec:proofnoiselevel}
From the proof of Theorem \ref{thm:optimalridge} in Appendix \ref{sec:proofoptimalridge}, we know that to obtain $\lambda_{\opt}<0$, it is sufficient to show
\[
\bbE \frac{g h \zeta}{(1+\zeta)^3}\bbE \frac{\zeta}{(1+\zeta)^2}-\bbE\frac{\zeta^2}{(1+\zeta)^3}\bbE\frac{g h }{(1+\zeta)^2}-\frac{\tilde{\sigma}^2}{\gamma}\bbE \frac{\zeta^2}{(1+\zeta)^3}>0.
\]
With the distribution assumption on $(h,g)$, this is equivalent to the following
\[
\gamma q(1-q)\frac{(h_1-1)(g_1-1)h_1m^2}{(1+m)^3(1+h_1m)^3}>\tilde{\sigma}^2\left((1-q)\frac{m^2}{(1+m)^3}+q\frac{h_1^2m^2}{(1+h_1m)^3}\right),
\]
where $m=m(0)$ satisfies that
\begin{eqnarray}
1=\gamma\left((1-q)\frac{m}{1+m} + q\frac{h_1m}{h_1m+1}\right).
\label{eq:noiselevel_eq1}
\end{eqnarray}
This gives the following upper bound for $\tilde{\sigma}^2$:
\begin{eqnarray}
\tilde{\sigma}^2<\gamma q(1-q) \frac{(h_1-1)(g_1-1)h_1}{(1-q)(1+h_1m)^3+q h_1^2(1+m)^3}.\label{eq:noiselevel_eq2}
\end{eqnarray}
To provide a more intuitive result, we remove $m$ from \eqref{eq:noiselevel_eq2}. Note that from \eqref{eq:noiselevel_eq1}, we can derive the following straightforward upper bound for $m$:
\[
m\leq \max\left(\frac{1}{h_1(\gamma q-1)}, \frac{1}{\gamma-1}\right),
\]
which we plug in \eqref{eq:noiselevel_eq2} and obtain
\[
\tilde{\sigma}^2<\gamma q(1-q)\max\left(\frac{(h_1-1)(g_1-1)h_1}{(1-q)\left(\frac{\gamma q}{\gamma q -1}\right)^3+q h_1^2\left(\frac{\gamma}{\gamma-1}\right)^3},\frac{(h_1-1)(g_1-1)h_1(\gamma-1)^3}{(1-q)\left(h_1+\gamma-1\right)^3+q h_1^2\gamma^3} \right).
\]


\subsection{Proof of Proposition \ref{cor:under-parameterized}}\label{sec:proofunder-parameterized}
Note that \eqref{eq:risk_eq1} holds for $\gamma<1$ as well. It is clear that the bias term is non-negative and is strictly positive when $\lambda \neq 0$. Hence, we know the bias achieve its minimum only at $\lambda=0$. We therefore only need to demonstrate that the variance term converges to a decreasing function of $\lambda$ for $\lambda>-c_0$. 

Let $s(z)$ be the Stieltjes transform of the limiting distribution of the eigenvalues of $\vX_{/w}^{\t}\vX_{/w}$, then we have $s(-\lambda)$ satisfying
\begin{eqnarray}
    s(-\lambda)\ =\ \bbE\frac{1}{h (1-\gamma+\gamma\lambda s(-\lambda))+\lambda}. \label{eq:sequation}
\end{eqnarray}
In addition, from the Marchenko-Pastur law, the minimal eigenvalue of $\vX_{/w}^{\t}\vX_{/w}$ is bounded by $\inf_{v\in[c_l,c_u]}h \cdot (1-\sqrt{\gamma})^2$. Hence, when $\gamma<1$, for all $\lambda>-c_0$, we know that $s(-\lambda)$ is well defined and positive. 
Observe that for all $\lambda\neq 0$, $m(-\lambda)$ and $s(-\lambda)$ satisfies the following relation
\begin{eqnarray}\label{eq:smrelation}
m(-\lambda) \ = \ \frac{1-\gamma}{\lambda}+\gamma s(-\lambda). 
\end{eqnarray}
Therefore from the proof of Theorem \ref{thm:RiskCal}, we have the exact same expression of Part 1 for $\gamma<1$ and $\lambda\neq 0$:
\[
\text{Part 1~}
\ \ipe\
\tilde{\sigma}^2 \frac{m'(-\lambda)}{m^2(-\lambda)}\quad \forall \lambda>-c_0, \lambda\neq 0.
\]
When $\lambda=0$, we should replace $m(-\lambda)$ by $s(-\lambda)$ using \eqref{eq:smrelation}. Since Part 1 is a continuous function of $\lambda$, we only need to focus on $\lambda\neq 0$ and show the following equation for all $\lambda>-c_0$ and $\lambda\neq 0$:
\begin{eqnarray}
-\frac{2\gamma(m'(-\lambda))^2}{m^3(-\lambda)}\tilde{\sigma}^2\frac{\bbE\frac{\zeta^2}{(1+\zeta)^3}}{1-\gamma\bbE\frac{\zeta^2}{(1+\zeta)^2}}\ <\ 0.\label{eq:under_extra_1}
\end{eqnarray}
Although we have proved \eqref{eq:under_extra_1} for the case $\gamma>1$ in Appendix \ref{sec:proofoptimalridge}, we used the fact that $m(-\lambda)>0$ which is not guaranteed when $\gamma<1$. In fact, only $s(z)$ and its any order derivatives are guaranteed to be positive on $z<c_0$, and $m(-\lambda)$ can be negative. Hence, we need to rederive \eqref{eq:under_extra_1}  for $\gamma<1$. From \eqref{eq:derivative_eq1} and \eqref{eq:relation_m'}, what is left to be shown is that
\begin{eqnarray}
\frac{m'(-\lambda)}{m(-\lambda)} \cdot \bbE\frac{\zeta^2}{(1+\zeta)^3} \ =\ m'(-\lambda)\cdot \bbE\frac{h^2\cdot m(-\lambda)}{(1+h\cdot m(-\lambda))^3}\ > \ 0 , \quad \forall \lambda>-c_0, \lambda\neq 0.\label{eq:under_goal_1}
\end{eqnarray}
By taking derivatives on both sides of \eqref{eq:smrelation} and from $s'(-\lambda)>0$, we have
\[
m'(-\lambda)-\frac{1-\gamma}{\lambda^2} \ =\ \gamma s'(-\lambda) > 0.
\]
We therefore have $m'(-\lambda)>0$ and \eqref{eq:under_goal_1} clearly holds when $m(-\lambda)>0$. Since $\lambda>0$ implies $m(-\lambda)>0$ due to \eqref{eq:smrelation}, we only need to show \eqref{eq:under_goal_1} when $\lambda<0$ and $m(-\lambda)<0$. We claim that when $\lambda<0$ and $m(-\lambda)<0$, $1+h\cdot m(-\lambda)<0$ holds almost surely, and thus \eqref{eq:under_goal_1} is true due to \[
    m(-\lambda)<0 \quad \text{and} \quad \bbE\frac{h^2}{(1+h\cdot  m(-\lambda))^3} < 0.
\]
We use contradiction to prove the claim. Suppose there exists $c_v>\inf h \triangleq c_h$ such that $1+h\cdot m(-\lambda)>0$ for all $h < c_v$ and the probability of $h< c_v$ is positive. Then let $c_m=-m(-\lambda)^{-1}$, we have $c_m>c_v>c_h>0$. Furthermore, from \eqref{eq:relation_m} and definition of $c_0$, we have
\[
-c_h(1-\sqrt{\gamma})^2=-c_0<\lambda = -c_m -\gamma \bbE \frac{h }{h\cdot  m(-\lambda)+1} \leq -c_m + \gamma \bbE \frac{h\cdot c_m}{h-c_m }\bbI_{h > c_m}. 
\]
Therefore, we have
\begin{eqnarray}
\bbE \frac{h }{h-c_m }\bbI_{h > c_m} > \frac{1}{\gamma}\left(1-\frac{c_h}{c_m}(1-\sqrt{\gamma})^2\right) > \frac{2-\sqrt{\gamma}}{\sqrt{\gamma}} > 0. \label{eq:under_eq_ineq}
\end{eqnarray}
On the other hand, since $m'(-\lambda)>0$, from \eqref{eq:relation_m'}, we have 
\[
0<\frac{1}{m^2(-\lambda)}-\gamma \bbE \frac{h^2}{(1+h\cdot m(-\lambda))^2}
\leq c_m^2-\gamma \bbE \frac{h^2c_m^2}{(h-c_m )^2}\bbI_{h > c_m},
\]
which is equivalent to 
\[
\frac{1}{\gamma}> \bbE \left(\frac{h }{h-c_m }\bbI_{h > c_m}\right)^2.
\]
However, from \eqref{eq:under_eq_ineq} and Jensen's inequality, we have
\[
\bbE \left(\frac{h }{h-c_m }\bbI_{h > c_m}\right)^2 > \frac{(2-\sqrt{\gamma})^2}{\gamma}> \frac{1}{\gamma}.
\]
We have arrived at a contradiction and thus $1+h m(-\lambda)<0$ should hold almost surely when $\lambda<0$ and $m(-\lambda)<0$.

\subsection{Proof of Proposition \ref{prop:risk-monotonicity}}\label{sec:proofrisk-monotonicity} 

First note that in the setup of general data covariance and isotropic prior on $\vbeta_*$, the prediction risk under optimal ridge regularization is given in \cite[Theorem 2.1]{dobriban2018high} as
\begin{eqnarray}
R(\lambda_{\mathrm{opt}}) = \frac{1}{\lambda_\opt m(-\lambda_\opt)}, \label{eq:risk-optimal-ridge-isotropic}
\end{eqnarray}
where $\lambda_\opt = \tilde{\sigma}^2\gamma/c$.
Note that \eqref{eq:relation_eq1} implies that
$m(-\lambda_\opt)$ satisfies the following equation when $\gamma>1$ or when $\gamma>0$ and $\tilde{\sigma}^2>0$\footnote{Note that when $\vSigma_\beta=\vI$ and $\tilde{\sigma}^2>0$ we have $\lambda_\opt>0$. Hence $m(-\lambda_\opt)$ exists for all $\gamma>0$.}: 
\begin{eqnarray}
\lambda_\opt m(-\lambda_\opt)\ = \ 1-\gamma \E \frac{h\cdot m(-\lambda_\opt)}{1+h\cdot m(-\lambda_\opt)}.\label{eq:extra_mark_eq1}
\end{eqnarray}
Therefore, taking the derivative of \eqref{eq:risk-optimal-ridge-isotropic} with respect to $\gamma$ yields
\begin{eqnarray}
\frac{\mathrm{d}R(\lambda_\opt)}{\mathrm{d}\gamma} &\propto& -\frac{1}{\gamma^2}\frac{1}{m(-\lambda_\opt)} + \frac{1}{\gamma}\frac{\partial m^{-1}(-\lambda_\opt)}{\partial\gamma} \n \\
&=&
-\frac{1}{\gamma^2}\frac{1}{m(-\lambda_\opt)}+ \frac{1}{\gamma}\left(\frac{\tilde{\sigma}^2}{c} + \E\frac{h}{1 + h\cdot m(-\lambda_\opt)} + \gamma \frac{\mathrm{d}}{\mathrm{d}\gamma}\E\frac{h}{1 + h\cdot m(-\lambda_\opt)}\right) \n \\
&\propto&
\frac{\mathrm{d}}{\mathrm{d}\gamma}\E\frac{h}{1 + h\cdot m(-\lambda_\opt)} \n \\
&=& 
-\left(\E\frac{h^2}{(1 + h\cdot m(-\lambda_\opt))^2}\right)\frac{\partial m(-\lambda_\opt)}{\partial \gamma}. 
\label{eq:optimal-risk-derivative-isotropic}
\end{eqnarray} 
For notational convenience we define $\alpha=\tilde{\sigma}^2 / c$, i.e.~$\lambda_\opt = \alpha\gamma$.
Note that for a fixed $\alpha$, $m(-\lambda_\opt)$ is a function of $\gamma$. Thus we let $u(\gamma)\triangleq m(-\alpha\gamma)$. Then \eqref{eq:optimal-risk-derivative-isotropic} implies that we only need to show $\frac{\text{d} u(\gamma)}{\text{d}\gamma}<0$. Taking the derivative with respect to $\gamma$ on both sides of \eqref{eq:extra_mark_eq1}, we have
\begin{eqnarray}
\alpha u(\gamma) + \alpha\gamma \frac{\text{d} u(\gamma)}{\text{d} \gamma}&=&
 -\E\frac{h\cdot u(\gamma)}{1+h\cdot u(\gamma)} - \gamma\E\frac{h^2}{(1+h\cdot u(\gamma))^2} \cdot \frac{\text{d} u(\gamma)}{\text{d} \gamma}, \n
 \end{eqnarray}
 which is equivalent to
\begin{eqnarray}
\frac{\text{d} u(\gamma)}{\text{d} \gamma} &=&
\frac{-\alpha u(\gamma)-\E\frac{h\cdot u(\gamma)}{1+h\cdot u(\gamma)}}{\alpha\gamma+\gamma\E\frac{h^2}{(1+h\cdot u(\gamma))^2}}\n\\
&=&
\frac{-\alpha m(-\lambda_\opt)-\E\frac{h\cdot m(-\lambda_\opt)}{1+h\cdot m(-\lambda_\opt)}}{\alpha\gamma+\gamma\E\frac{h^2}{(1+h\cdot m(-\lambda_\opt))^2}}\n\\
&=&- \frac{1}{\gamma\left(\alpha\gamma+\gamma\E\frac{h^2}{(1+h\cdot m(-\lambda_\opt))^2}\right)},\n
\end{eqnarray}
where the last inequality holds due to \eqref{eq:extra_mark_eq1}. 
Finally, when $\tilde{\sigma}^2=0$ and $\gamma<1$, we know that $\lambda_\opt=0$ and $R(\lambda_\opt)=0$.
We thus know that the prediction risk $R$ is increasing as a function of $\gamma\in(0,\infty)$.


\section{Proofs omitted in Section \ref{sec:optimal_w}}
\subsection{Proof of Theorem \ref{thm:optimalw_ridgeless}}\label{sec:proofoptimalw_ridgeless}
We first show that $\vSigma_w=\vSigma_x^{-1}$, i.e., $r$ being a point mass, is the optimal $\vSigma_w$ for the variance term. From \eqref{eq:relation_m} and \eqref{eq:relation_m'}, we know the the variance function $\rv(r)$ can be written as
\begin{eqnarray}
\rv(r)&=&\tilde{\sigma}^2\frac{1}{1-\gamma \bbE\frac{\zeta_r^2}{(1+\zeta_r)^2}}\ = \ \tilde{\sigma}^2\frac{1}{\gamma \bbE\frac{\zeta_r}{(1+\zeta_r)^2}},\n
\end{eqnarray}
where we define $\zeta_r=r\cdot m(0)$ in this proof. Note that $\frac{\zeta_r}{1+\zeta_r}$ and $\frac{1}{1+\zeta_r}$ are both monotonic function of $\zeta_r$ with different monotonicity, we thus have
\[
    \bbE\frac{\zeta_r}{(1+\zeta_r)^2} \leq \bbE\frac{\zeta_r}{1+\zeta_r}\bbE\frac{1}{1+\zeta_r}=\frac{1}{\gamma}\left(1-\frac{1}{\gamma}\right),
\]
where the last equality holds due to \eqref{eq:relation_m}. The equality is achieved only when $r$ is a single point mass. Hence, we have
\[
\rv(r)\geq \frac{\tilde{\sigma}^2}{(1-\frac{1}{\gamma})} = \frac{\tilde{\sigma}^2\gamma}{\gamma-1}.
\]
The minimum variance is achieved when $r$ is a single point mass, i.e., $\vD_{x/w}=\vI$ and therefore, $\vSigma_w=\vSigma_x^{-1}$. 

For the bias term, we first show that $r \aseq sv $, i.e., $\vSigma_w=\bar{\vSigma}_{\beta}^{-1}$ is the optimal choice of $r$ for all non-negative random variable\footnote{We do not require $r$ being bounded away from $0$ and $\infty$ because we focus on the function $\rb$ directly.}. The result for $r\in \cS_r$ immediately follows because as long as $r\in\cS_r$, $\rb$ remains the same when we replace $v$ by $\bbE[v|s]$. Suppose $r\not = sv $ almost surely. Let us define $r_{\alpha}=\alpha\cdot sv +(1-\alpha)r$ and consider the following bias function $\rb(\alpha)$:
\[
\rb(\alpha)\triangleq \frac{m'_{\alpha}(0)}{m^2_{\alpha}(0)}\gamma \bbE\frac{sv }{(r_{\alpha}\cdot m_{\alpha}(0)+1)^2},
\]
where $m_{\alpha}(-\lambda), m'_{\alpha}(-\lambda)>0$ satisfy that
\begin{eqnarray}
\lambda &=&\frac{1}{m_{\alpha}(-\lambda)}-\gamma\bbE \frac{r_{\alpha}}{1+r_{\alpha}\cdot m_{\alpha}(-\lambda)}\n\\
1&=&\left(\frac{1}{m_{\alpha}^2(-\lambda)}-\gamma\bbE\frac{r_{\alpha}^2}{(r_{\alpha}\cdot m_{\alpha}(-\lambda)+1)^2}\right)m'_{\alpha}(-\lambda).\label{eq:relation_alpham}
\end{eqnarray}
Note that $m_{\alpha}(z)$ is the Stieltjes transform of the limiting distribution of the eigenvalues of $\frac{1}{n}\vX_{\alpha}\vX_{\alpha}^{\t}$ where the covariance matrix of the rows of $\vX_{\alpha}$ has its eigenvalues weakly converging to the random variable $r_{\alpha}$. Hence $m_{\alpha}(0)>0$ and $m'_{\alpha}(0)>0$ are well defined.

Our goal is to show that $1\in\argmin_{\alpha} \rb(\alpha)$. We define $\zeta_{\alpha}=r_{\alpha}\cdot m_{\alpha}(0)$. Then from \eqref{eq:relation_eq1} and \eqref{eq:relation_alpham}, we know that \eqref{eq:relation_m} and \eqref{eq:relation_m'} hold with $\zeta$ replaced by $\zeta_{\alpha}$. Hence, we have
\[
\rb(\alpha)\ =\ \frac{\gamma \bbE\frac{sv }{(\zeta_{\alpha}+1)^2}}{1-\gamma \bbE \frac{\zeta_{\alpha}^2}{(\zeta_{\alpha}+1)^2}} \ = \ \frac{ \bbE\frac{sv }{(\zeta_{\alpha}+1)^2}}{\bbE \frac{\zeta_{\alpha}}{(\zeta_{\alpha}+1)^2}},
\]
where the last equality holds due to \eqref{eq:relation_m}. By taking derivatives with respect to $\alpha$ in \eqref{eq:relation_m}, we know that
\begin{eqnarray}
\frac{\partial m_{\alpha}(-\lambda)}{\partial \alpha}\Big|_{\lambda=0} &=&-\frac{\bbE\frac{(sv -r)m^2_{\alpha}(-\lambda)}{(1+r_{\alpha}\cdot m_{\alpha}(-\lambda))^2}}{\bbE \frac{r_{\alpha}\cdot m_{\alpha}(-\lambda)}{(1+r_{\alpha}\cdot m_{\alpha}(-\lambda))^2}}\Big|_{\lambda=0}\ = \ -\frac{m_{\alpha}(0)\cdot \bbE\frac{\psi_{\alpha}-\zeta_{\alpha}}{(1+\zeta_{\alpha})^2}}{(1-\alpha)\bbE \frac{\zeta_{\alpha}}{(1+\zeta_{\alpha})^2}} ,\label{eq:m'alpha}
\end{eqnarray}
where $\psi_{\alpha}=sv\cdot m_{\alpha}(0)$. With \eqref{eq:m'alpha}, we have
\begin{eqnarray}
\frac{\dif \rb(\alpha)}{\dif \alpha}&\stackrel{(i)}{\propto}&-2\bbE\frac{\psi_{\alpha}(\psi_{\alpha}-\zeta_{\alpha})}{(1+\zeta_{\alpha})^3}\left(\bbE\frac{\zeta_{\alpha}}{(1+\zeta_{\alpha})^2}\right)^2+2\bbE\frac{\psi_{\alpha}\zeta_{\alpha}}{(1+\zeta_{\alpha})^3}\bbE\frac{(\psi_{\alpha}-\zeta_{\alpha})}{(1+\zeta_{\alpha})^2}\bbE\frac{\zeta_{\alpha}}{(1+\zeta_{\alpha})^2}\n\\
&&-\bbE\frac{\zeta_{\alpha}}{(1+\zeta_{\alpha})^2}\bbE\frac{(1-\zeta_{\alpha})(\psi_{\alpha}-\zeta_{\alpha})}{(1+\zeta_{\alpha})^3}\bbE\frac{\psi_{\alpha}}{(1+\zeta_{\alpha})^2}+\bbE\frac{(1-\zeta_{\alpha})\zeta_{\alpha}}{(1+\zeta_{\alpha})^3}\bbE\frac{\psi_{\alpha}-\zeta_{\alpha}}{(1+\zeta_{\alpha})^2}\bbE\frac{\psi_{\alpha}}{(1+\zeta_{\alpha})^2},\n\\
\label{eq:rv'_alpha}
\end{eqnarray}
where in equation (i) we omitted the following positive multiplicative scalar:
\[
\left((1-\alpha)m_{\alpha}(0)\left(\bbE \frac{\zeta_{\alpha}}{(1+\zeta_{\alpha})^2}\right)^3\right)^{-1}
\]
We claim that the RHS of \eqref{eq:rv'_alpha} is equivalent to the following 
\begin{eqnarray}
-2\underbrace{\bbE\frac{(\psi_{\alpha}-\zeta_{\alpha})^2}{(1+\zeta_{\alpha})^3}\left(\bbE\frac{\zeta_{\alpha}}{(1+\zeta_{\alpha})^2}\right)^2}_{A}-2\underbrace{\left(\bbE\frac{\psi_{\alpha}-\zeta_{\alpha}}{(1+\zeta_{\alpha})^2}\right)^2\bbE\frac{\zeta_{\alpha}^2}{(1+\zeta_{\alpha})^3}}_{B}+4\underbrace{\bbE\frac{\zeta_{\alpha}(\psi_{\alpha}-\zeta_{\alpha})}{(1+\zeta_{\alpha})^3}\bbE\frac{\psi_{\alpha}-\zeta_{\alpha}}{(1+\zeta_{\alpha})^2}\bbE\frac{\zeta_{\alpha}}{(1+\zeta_{\alpha})^2}}_{C}.\n\\
\label{eq:quantity}
\end{eqnarray}
We apply the AM-GM inequality on the first two terms and obtain that $A+B\geq 2\sqrt{AB}$, and then apply Cauchy-Schwartz inequality on $\sqrt{AB}$ and obtain that $\sqrt{AB}\geq C$. Hence we know $-2(A+B-2C)\leq 0$ and the equality is achieved only when $\psi_{\alpha}\aseq \zeta_{\alpha}$ which implies $\alpha=1$ or both $\psi_{\alpha}$ and $\zeta_{\alpha}$ are single point mass. For the later case, we have $\frac{\dif \rb(\alpha)}{\dif \alpha}\equiv 0$ for all $\alpha$ and therefore $\alpha=1$, i.e., $r\aseq sv$, is one of the minimum solutions. For the first case, we know $\rb(\alpha)$ is a strictly decreasing function of $\alpha$ and achieves its minimum at $\alpha=1$ which is $r\aseq sv$. 
To show \eqref{eq:quantity}, we first simplify the first two terms in the RHS of \eqref{eq:rv'_alpha}.
\begin{eqnarray}
\lefteqn{-2\bbE\frac{\psi_{\alpha}(\psi_{\alpha}-\zeta_{\alpha})}{(1+\zeta_{\alpha})^3}\left(\bbE\frac{\zeta_{\alpha}}{(1+\zeta_{\alpha})^2}\right)^2+2\bbE\frac{\psi_{\alpha}\zeta_{\alpha}}{(1+\zeta_{\alpha})^3}\bbE\frac{(\psi_{\alpha}-\zeta_{\alpha})}{(1+\zeta_{\alpha})^2}\bbE\frac{\zeta_{\alpha}}{(1+\zeta_{\alpha})^2}}\n\\
&=&-2A-2\bbE\frac{\zeta_{\alpha}(\psi_{\alpha}-\zeta_{\alpha})}{(1+\zeta_{\alpha})^3}\left(\bbE\frac{\zeta_{\alpha}}{(1+\zeta_{\alpha})^2}\right)^2+4C+2\bbE\frac{2\zeta_{\alpha}^2-\psi_{\alpha}\zeta_{\alpha}}{(1+\zeta_{\alpha})^3}\bbE\frac{(\psi_{\alpha}-\zeta_{\alpha})}{(1+\zeta_{\alpha})^2}\bbE\frac{\zeta_{\alpha}}{(1+\zeta_{\alpha})^2}\n\\
&=&-2A+4C-2\bbE\frac{\zeta_{\alpha}(\psi_{\alpha}-\zeta_{\alpha})}{(1+\zeta_{\alpha})^3}\bbE\frac{\zeta_{\alpha}}{(1+\zeta_{\alpha})^2}\bbE\frac{\psi_{\alpha}}{(1+\zeta_{\alpha})^2}+2\bbE\frac{\zeta_{\alpha}^2}{(1+\zeta_{\alpha})^3}\bbE\frac{(\psi_{\alpha}-\zeta_{\alpha})}{(1+\zeta_{\alpha})^2}\bbE\frac{\zeta_{\alpha}}{(1+\zeta_{\alpha})^2}\n\\
&=&-2A-2B+4C\n\\
&&-2\bbE\frac{\zeta_{\alpha}(\psi_{\alpha}-\zeta_{\alpha})}{(1+\zeta_{\alpha})^3}\bbE\frac{\zeta_{\alpha}}{(1+\zeta_{\alpha})^2}\bbE\frac{\psi_{\alpha}}{(1+\zeta_{\alpha})^2}+2\bbE\frac{\zeta_{\alpha}^2}{(1+\zeta_{\alpha})^3}\bbE\frac{\psi_{\alpha}-\zeta_{\alpha}}{(1+\zeta_{\alpha})^2}\bbE\frac{\psi_{\alpha}}{(1+\zeta_{\alpha})^2}.\n\\\label{eq:ridgelessoptimalproof_eq1}
\end{eqnarray}
Similarly for the last two terms of \eqref{eq:rv'_alpha},
\begin{eqnarray}
\lefteqn{\bbE\frac{(1-\zeta_{\alpha})\zeta_{\alpha}}{(1+\zeta_{\alpha})^3}\bbE\frac{\psi_{\alpha}-\zeta_{\alpha}}{(1+\zeta_{\alpha})^2}\bbE\frac{\psi_{\alpha}}{(1+\zeta_{\alpha})^2}-\bbE\frac{\zeta_{\alpha}}{(1+\zeta_{\alpha})^2}\bbE\frac{(1-\zeta_{\alpha})(\psi_{\alpha}-\zeta_{\alpha})}{(1+\zeta_{\alpha})^3}\bbE\frac{\psi_{\alpha}}{(1+\zeta_{\alpha})^2}}\n\\
&=&\bbE\frac{(1-\zeta_{\alpha})\zeta_{\alpha}}{(1+\zeta_{\alpha})^3}\bbE\frac{\psi_{\alpha}-\zeta_{\alpha}}{(1+\zeta_{\alpha})^2}\bbE\frac{\psi_{\alpha}}{(1+\zeta_{\alpha})^2}\n\\
&&-\bbE\frac{\zeta_{\alpha}}{(1+\zeta_{\alpha})^2}\bbE\frac{(\psi_{\alpha}-\zeta_{\alpha})}{(1+\zeta_{\alpha})^2}\bbE\frac{\psi_{\alpha}}{(1+\zeta_{\alpha})^2}+2\bbE\frac{\zeta_{\alpha}}{(1+\zeta_{\alpha})^2}\bbE\frac{\zeta_{\alpha}(\psi_{\alpha}-\zeta_{\alpha})}{(1+\zeta_{\alpha})^3}\bbE\frac{\psi_{\alpha}}{(1+\zeta_{\alpha})^2}\n\\
&=&-2\bbE\frac{\zeta_{\alpha}^2}{(1+\zeta_{\alpha})^3}\bbE\frac{\psi_{\alpha}-\zeta_{\alpha}}{(1+\zeta_{\alpha})^2}\bbE\frac{\psi_{\alpha}}{(1+\zeta_{\alpha})^2}+2\bbE\frac{\zeta_{\alpha}}{(1+\zeta_{\alpha})^2}\bbE\frac{\zeta_{\alpha}(\psi_{\alpha}-\zeta_{\alpha})}{(1+\zeta_{\alpha})^3}\bbE\frac{\psi_{\alpha}}{(1+\zeta_{\alpha})^2}.\n\\\label{eq:ridgelessoptimalproof_eq2}
\end{eqnarray}
Combine \eqref{eq:ridgelessoptimalproof_eq1} and \eqref{eq:ridgelessoptimalproof_eq2}, we have \eqref{eq:quantity} holds.

\subsection{Proof of Proposition \ref{cor:optimal_wPCR}}\label{sec:proofoptimal_wPCR}
Since $\vSigma_{w}=\bar{\vSigma}_{\beta}^{-1}$ is the optimal choice for $\vSigma_w\in\cH_w$, we only need to prove this proposition in the case when $\vSigma_w=\left(f_v(\vSigma_x)\right)^{-1}$.

Note that the proposition holds in the regime $\theta\gamma>1$ due to the proof of Theorem \ref{thm:optimalw_ridgeless} and Corollary \ref{cor:PCRrisk}. When $\theta\gamma<1$, denote the quantile functions of $s$ and $\tilde{s}\triangleq \bbE[v|s]\cdot s$ as $Q_1$ and $Q_2$ respectively. We have
\begin{eqnarray}
\bbE\left[sv\cdot \bbI_{s<Q_1(1-\theta)}\right]
&=&\bbE\left[\bbE[v|s]\cdot s\bbI_{s<Q_1(1-\theta)}\right]\n\\
&=&\bbE \left[\tilde{s}\bbI_{s<Q_1(1-\theta),\tilde{s}<Q_2(1-\theta)}\right]+\bbE\left[\tilde{s}\bbI_{s<Q_1(1-\theta),\tilde{s}\geq Q_2(1-\theta)}\right]\n\\
&\geq&\bbE \left[\tilde{s}\bbI_{s<Q_1(1-\theta),\tilde{s}<Q_2(1-\theta)}\right]+Q_2(1-\theta)\Pr{s<Q_1(1-\theta),\tilde{s}\geq Q_2(1-\theta)}\n\\
&=&\bbE \left[\tilde{s}\bbI_{s<Q_1(1-\theta),\tilde{s}<Q_2(1-\theta)}\right]+Q_2(1-\theta)\Pr{s\geq Q_1(1-\theta),\tilde{s}< Q_2(1-\theta)}\n\\
&\geq&\bbE \left[\tilde{s}\bbI_{s<Q_1(1-\theta),\tilde{s}<Q_2(1-\theta)}\right]+\bbE \left[\tilde{s} \bbI_{s\geq Q_1(1-\theta),\tilde{s}< Q_2(1-\theta)}\right]\n\\
&=&\bbE \left[\tilde{s}\bbI_{\tilde{s}<Q_2(1-\theta)}\right].\n
\end{eqnarray}
From Corollary \ref{cor:PCRrisk}, we know that the risk achieved by the PCR estimator is at least the same as that of a second PCR estimator where we replace $(\vSigma_x,\vSigma_{\beta})$ by $(\vSigma_x\vSigma_{w}^{-1}, \vI)$. 
From \cite{xu2019number}, the optimal risk achieved by the PCR estimate for $\theta\gamma<1$ in the second PCR problem is worse than the full model risk $R(\bbE[v|s]\cdot s, 0)$ which is the same risks achieved by the minimum $\|\hvb\|_{\vSigma_{w}}$ solution. 
Hence we know that the minimum $\|\hvb\|_{\vSigma_{w}}$ solution outperforms the PCR estimate for $\theta\gamma<1$ as well.


\subsection{Proof of Theorem \ref{thm:optimalw_optimal}}\label{sec:proofoptimalw_optimal}
From \eqref{eq:relation_m} and \eqref{eq:relation_m'}, we have the following equivalent formula for the risk function $R(r, \lambda)$:
\[
R(r,\lambda)\ = \ \frac{\tilde{\sigma}^2+\gamma \bbE\frac{sv }{(1+\zeta_r)^2}}{1-\gamma\bbE\frac{\zeta_r^2}{(1+\zeta_r)^2}},
\]
where we define $\zeta_r=r\cdot m_r(-\lambda)$ in this proof. We also know that $m_r(-\lambda)$ satisfies
\[
1=\lambda m_{r}(-\lambda)+\gamma\bbE\frac{\zeta_r}{1+\zeta_r}. 
\]
Let us first consider $r\in \cH_r$. The result for $r\in \cS_r$ immediately follows because as long as $r\in\cS_r$, $R(r,\lambda)$ remains the same when we replace $v$ by $\bbE[v|s]$. We now apply similar proof strategy of Theorem \ref{thm:optimalw_ridgeless} in Section \ref{sec:proofoptimalw_ridgeless}. Consider any $r\in\cH_r$ with $r\not= sv $ almost surely. We define $r_{\alpha}=\alpha\cdot sv +(1-\alpha)r$ and consider the following risk function $R_{\alpha}(\lambda)$:
\[
R_{\alpha}(\lambda)\ = \ \frac{\tilde{\sigma}^2+\gamma \bbE\frac{sv }{(1+\zeta_{\alpha}(\lambda))^2}}{1-\gamma\bbE\frac{\zeta_{\alpha}(\lambda)^2}{(1+\zeta_{\alpha}(\lambda))^2}},
\]
where $\zeta_{\alpha}(\lambda)=r_{\alpha}\cdot m_{\alpha}(-\lambda)$, and $m_{\alpha}(-\lambda)$ satisfies that
\begin{eqnarray}
\lambda &=&\frac{1}{m_{\alpha}(-\lambda)}-\gamma\bbE \frac{r_{\alpha}}{1+r_{\alpha}\cdot m_{\alpha}(-\lambda)}\n
\end{eqnarray}
Note that $m_{\alpha}(z)$ is the Stieltjes transform of the limiting distribution of the eigenvalues of $\frac{1}{n}\vX_{\alpha}\vX_{\alpha}^{\t}$ where the covariance matrix of the rows of $\vX_{\alpha}$ has its eigenvalues weakly converges to the random variable $r_{\alpha}$. We define $c_{\alpha}=-\inf_{x\in \cK} x>0$, in which
\[\cK=\text{support of the limiting distribution of the eigenvalues of $\frac{1}{n}\vX_{\alpha}\vX_{\alpha}^{\t}$}.
\]
We know that $m_{\alpha}(-\lambda)\geq 0$ and $m'_{\alpha}(-\lambda)>0$ for all $\lambda >c_{\alpha}$ and from Section 4 of \cite{silverstein1995analysis}, we know that
\[
\lim_{\lambda\rightarrow c_{\alpha}^{+}}\gamma \bbE \frac{\zeta_{\alpha}^2(\lambda)}{(1+\zeta_{\alpha}(\lambda))^2}=1.
\]
Furthermore, $m_{\alpha}(-\lambda) \rightarrow 0$ as $\lambda\rightarrow \infty$. Hence we know that $\lambda_{\opt}(\alpha)=\argmin_{\lambda} R_{\alpha}(\lambda)$ exists\footnote{As $\lambda\to\infty$, the LHS of \eqref{eq:7condition} remains finite and the RHS of \eqref{eq:7condition} goes to infinity. On the other hand, as $\lambda\rightarrow c_{\alpha}^{+}$, the LHS of \eqref{eq:7condition} goes to infinity and the RHS of \eqref{eq:7condition} remains finite.}, and by taking derivatives with respect to $\lambda$ for $R_{\alpha}(\lambda)$, it is clear that $\lambda_{\opt}(\alpha)$ should satisfy that
\begin{eqnarray}
    \frac{\tilde{\sigma}^2+\gamma \bbE\frac{sv }{(\zeta_{\alpha}+1)^2}}{1-\gamma \bbE\frac{\zeta_\alpha^2}{(1+\zeta_\alpha)^2}} \ = \ \frac{\bbE \frac{sv\cdot \zeta_\alpha}{(1+\zeta_\alpha)^3}}{\bbE \frac{\zeta_\alpha^2}{(1+\zeta_\alpha)^3}},\label{eq:7condition}
\end{eqnarray}
where we slightly abuse the notation and use $\zeta_{\alpha}$ as a shorthand for $\zeta_{\alpha}(\lambda_{\opt}(\alpha))$. We now consider the following optimization problem:
\[
    \min_{\alpha} R_{\alpha}(\lambda_{\opt}(\alpha)).
\]
Our goal is to show that $1\in\argmin_{\alpha}R_{\alpha}(\lambda_{\opt}(\alpha))$, from which we have
\[
\min_{\lambda}R_{sv}(\lambda)=R_{\alpha}(\lambda_{\opt}(\alpha))|_{\alpha=1}\leq R_{\alpha}(\lambda_{\opt}(\alpha))|_{\alpha=0}=\min_{\lambda}R_{r}(\lambda).
\]
Which informs us that the optimal $r$ for optimal weighted ridge regression is $r\aseq sv $.

Taking the derivatives of $R_{\alpha}(\lambda_{\opt}(\alpha))$ with respect to $\alpha$ yields
\begin{eqnarray}
\frac{\dif R_{\alpha}(\lambda_{\opt}(\alpha))}{\dif \alpha}&=&\frac{\partial R_{\alpha}(\lambda)}{\partial m_{\alpha}(-\lambda)}\cdot \left(\frac{\partial m_{\alpha}(-\lambda)}{\partial \lambda}\cdot\frac{\dif \lambda_{\opt}(\alpha)}{\dif \alpha}+\frac{\partial m_{\alpha}(-\lambda)}{\partial \alpha} \right)\Big|_{\lambda=\lambda_{\opt}(\alpha)} + \frac{\partial R_{\alpha}(\lambda)}{\partial \alpha}\Big|_{\lambda=\lambda_{\opt}(\alpha)}\n\\
&\stackrel{(i)}{=}&\frac{\partial R_{\alpha}(\lambda)}{\partial \alpha}\Big|_{\lambda=\lambda_{\opt}(\alpha)}\n\\
&\propto&-\bbE\frac{sv (\psi_{\alpha}-\zeta_{\alpha})}{(1+\zeta_{\alpha})^3}\left(1-\gamma \bbE\frac{\zeta_\alpha^2}{(1+\zeta_\alpha)^2}\right)+\bbE\frac{\zeta_{\alpha}(\psi_{\alpha}-\zeta_{\alpha})}{(1+\zeta_{\alpha})^3}\left(\tilde{\sigma}^2+\gamma \bbE\frac{sv }{(\zeta_{\alpha}+1)^2}\right),\n\\
\label{eq:extra_eq1}
\end{eqnarray}
where equality (i) holds due to $\frac{\partial R_{\alpha}(\lambda)}{\partial m_{\alpha}(-\lambda)}\Big|_{\lambda=\lambda_{\opt}(\alpha)}=0$, and we defined $\psi_{\alpha}=sv\cdot m_{\alpha}(-\lambda_{\opt}(\alpha))$ in this proof. In addition, the multiplicative scalar omitted in the last equation is the following positive constant
\[
\left(\frac{1-\alpha}{2\gamma}\left(1-\gamma \bbE\frac{\zeta_\alpha^2}{(1+\zeta_\alpha)^2}\right)^2\right)^{-1}.
\]
Combining \eqref{eq:7condition} and \eqref{eq:extra_eq1} yields
\begin{eqnarray}
\frac{\dif R_{\alpha}(\lambda_{\opt}(\alpha))}{\dif \alpha}\leq 0 &\Leftrightarrow& \bbE \frac{\psi_{\alpha}\zeta_\alpha}{(1+\zeta_\alpha)^3}\bbE\frac{\zeta_{\alpha}(\psi_{\alpha}-\zeta_{\alpha})}{(1+\zeta_{\alpha})^3}\leq \bbE\frac{\psi_{\alpha}(\psi_{\alpha}-\zeta_{\alpha})}{(1+\zeta_{\alpha})^3}\bbE \frac{\zeta_\alpha^2}{(1+\zeta_\alpha)^3}\n\\
&\Leftrightarrow&\left(\bbE\frac{\zeta_{\alpha}(\psi_{\alpha}-\zeta_{\alpha})}{(1+\zeta_{\alpha})^3}\right)^2\leq \bbE\frac{(\psi_{\alpha}-\zeta_{\alpha})^2}{(1+\zeta_{\alpha})^3}\bbE \frac{\zeta_\alpha^2}{(1+\zeta_\alpha)^3}.\n
\end{eqnarray}
Where the last inequality holds for all $\alpha\leq 1$ by  Cauchy-Schwartz. Therefore,
\[
1\in \argmin_{\alpha}R_{\alpha}(\lambda_{\opt}(\alpha)).
\]
This completes the proof of the theorem.

\bigskip

\section{Auxiliaries}\label{app:figure}

\subsection{Experiment Setup}

We include the detailed constructions of $\vd_x$ and $\vd_{\beta}$ and figures mentioned in the main text.
The values of $\vd_x$ and $\vd_{\beta}$ used in Figure \ref{fig:risk} are constructed in the following way:

\begin{itemize}
    \item \textbf{Discrete to discrete}: For $\vd_x$, we set each quarter of elements to be $1,3,5$ and $7$ respectively; For $\vd_{\beta}$, we set one forth elements to be $8$ and rest of the elements to be $1$.
    \item \textbf{Discrete to continuous}: For $\vd_x$, we set half of the elements to be $1$ and rest of the elements to be $8$; For $\vd_{\beta}$, we i.i.d.~sample from $\text{unif}([1,8])$.
    \item \textbf{Continuous to continuous}: For $\vd_x$, we i.i.d.~sample from $\text{unif}([1,5])$; For $\vd_{\beta}$, we i.i.d.~sample from random variable $a=\min(u^2+1,5)$ where $u\sim \cN(0,1)$.  
    \item \textbf{Continuous to discrete}: For $\vd_x$, we i.i.d.~sample from $\text{unif}([1,8])$; For $\vd_{\beta}$, we set half of the elements to be $1$ and rest of the elements to be $7$.
\end{itemize}

The values of $\vd_x$ and $\vd_{\beta}$ used in Figure \ref{fig:optimal_egh} are constructed as:

\begin{itemize}
    \item We construct $a$ and $b$ to be two independent Gaussian $\cN(0,1)$ random variables.
    \item \textbf{Left}: Let $(d_{x,i},d_{\beta,i})\stackrel{\text{i.i.d.}}{\sim} (s, v)$ where $s=|a|+5$ and $v = (a+b/2)^2+1$. It is then straightforward to show that $f_v(s)=\bbE[v|s] = (s-5)^2+5/4$. Hence, the optimal $\vSigma_w\in\cS_w$ is $\vSigma_w=\left((\vSigma_x-5\vI)^2+1.25\vI\right)^{-1}$.
    \item \textbf{Right}: Let $(d_{x,i},d_{\beta,i})\stackrel{\text{i.i.d.}}{\sim} (s, v)$ where $s=|a|^{-1}+2$ and $v = (a+b/2)^2+1$. It is then straightforward to show that $f_v(s)=\bbE[v|s] = \frac{1}{(s-2)^{2}}+5/4$. Hence, the optimal $\vSigma_w\in\cS_w$ is $\vSigma_w=\left((\vSigma_x-2\vI)^{-2}+1.25\vI\right)^{-1}$.
\end{itemize}

\begin{figure}[!htb]
\centering
\begin{minipage}[t]{0.4\linewidth}
\centering
{\includegraphics[width=0.9\textwidth]{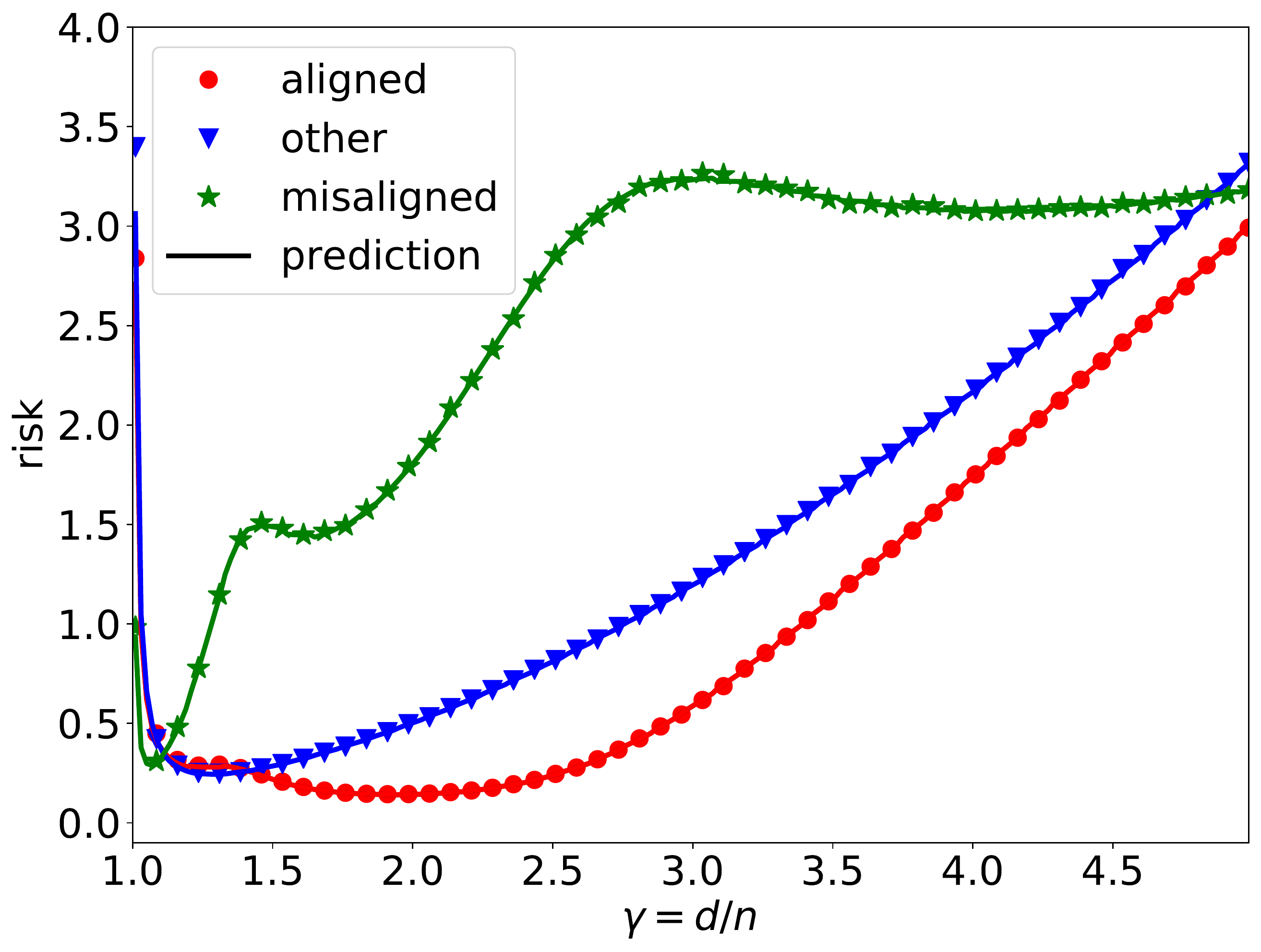}} \\ \vspace{-0.10cm}
(a) ridgeless regression risk.
\end{minipage}
\begin{minipage}[t]{0.4\linewidth} 
\centering
{\includegraphics[width=0.945\textwidth]{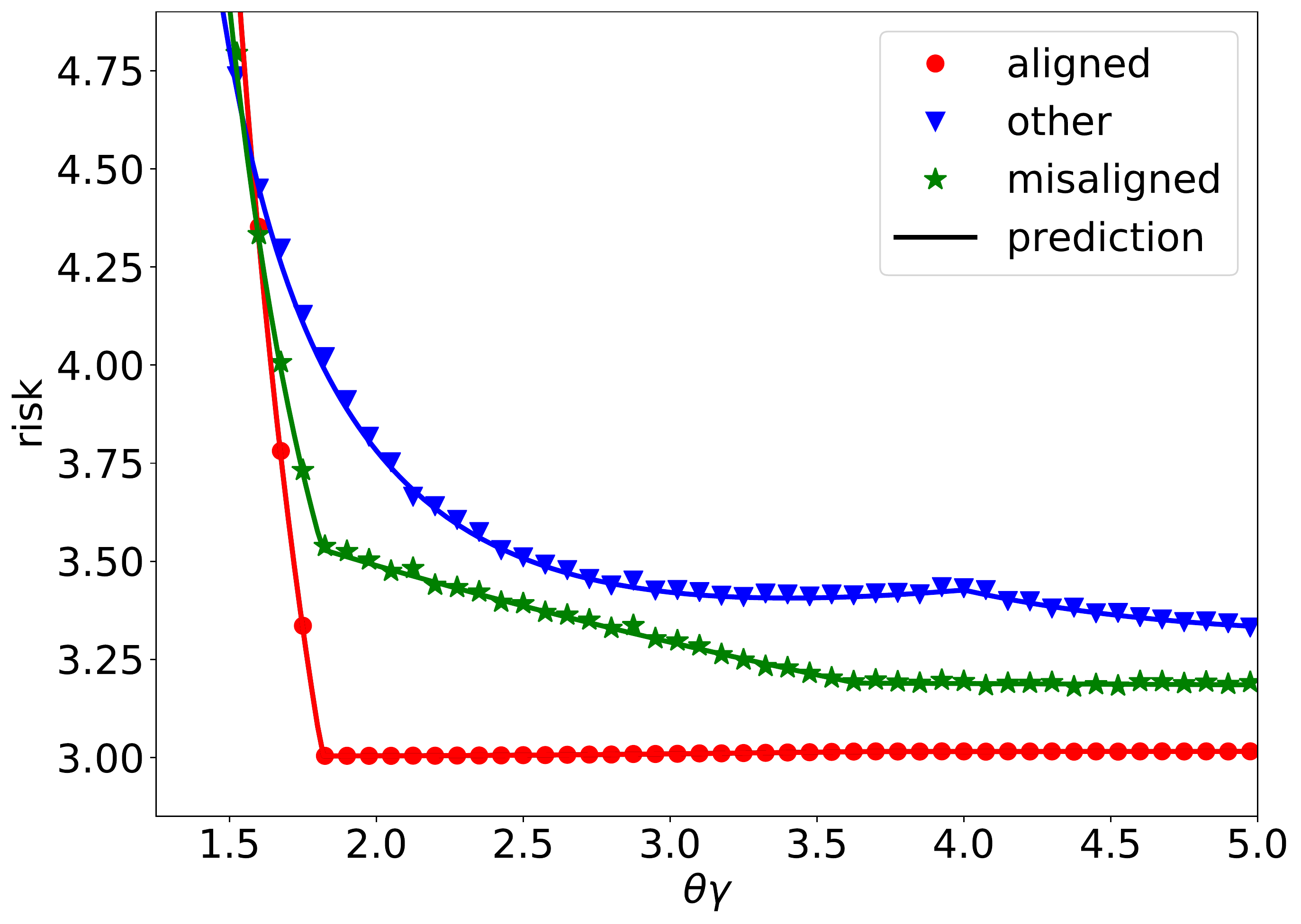}} \\ \vspace{-0.10cm}
(b) PCR risk.
\end{minipage}
\vspace{-1mm}
\caption{\small Comparison of the ridgeless regression estimator \cite{hastie2019surprises} and the PCR estimator. We set $n=300$, $\gamma=5$ and SNR=50.
Observe that the ridgeless regression risk exhibits multiple peaks in the overparameterized regime due to the anisotropic covariances (especially when $\vd_x$ and $\vd_\beta$ are misaligned). In contrast, the PCR risk is largely decreasing with $\theta$, especially for the misaligned case, which agrees with Proposition~\ref{cor:PCRrisk}. We remark that the PCR risk is not always monotone for $\theta\gamma>1$, as illustrated by the blue curve.
}
\label{fig:multiple_descent}
\end{figure}

The covariances in Figure~\ref{fig:multiple_descent} are constructed as follow (we remark that the slightly different scaling is to ensure that the resulting risk for each choice is roughly of the same magnitude to be presented in one figure):
\begin{itemize}
    \item \textbf{Aligned:} We construct $\vd_x$ to be three point masses $(a,b,c)$ with weights $(4/11,4/11, 3/11)$, respectively. We choose $\kappa=50$ and locate $a=1$, $b=a/\kappa$ and $c=b/\kappa$. We set $\vSigma_\beta = 18/5\cdot\vSigma_x$.
    \item \textbf{Misaligned:} We construct $\vd_x$ to be the same as the aligned case, and set $\vSigma_\beta = 4/9\cdot\vSigma_x^{-1}$.
    \item \textbf{Other:} We construct $\vd_x$ and $\vd_\beta$ to be the sum of two vectors $\vd_1$ and $\vd_2$, both of which consists of two point masses. $\vd_1$ has its first $1/5$ entries to be 1 and the rest $1/10$, and $\vd_2$ has its first $4/5$ entries to be 1 and the rest $1/10$. We set $\vd_x = 2\cdot\vd_1$ and $\vd_\beta = (\vd_1+\vd_2)/3$, 
\end{itemize}

\bigskip
\subsection{Additional Figures}
\begin{figure}[!htb]
\centering

\begin{minipage}{.99\linewidth} 
\centering
\subcaptionbox{Aligned, SNR $\xi=5$}{\vspace{-0.2cm}\includegraphics[width=0.32\linewidth]{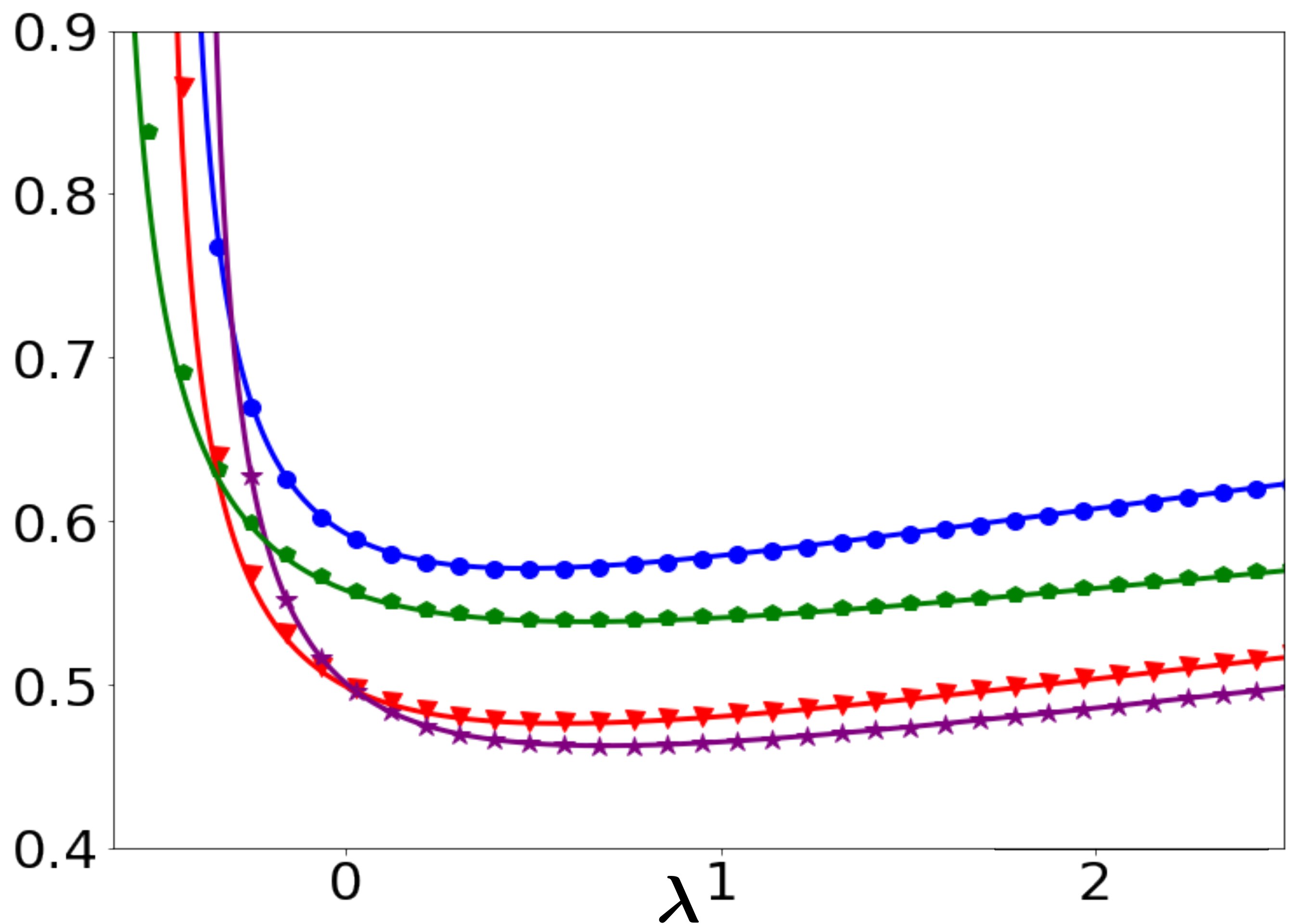}}
\subcaptionbox{Misaligned, SNR $\xi=5$}{\vspace{-0.2cm} \includegraphics[width=0.32\linewidth]{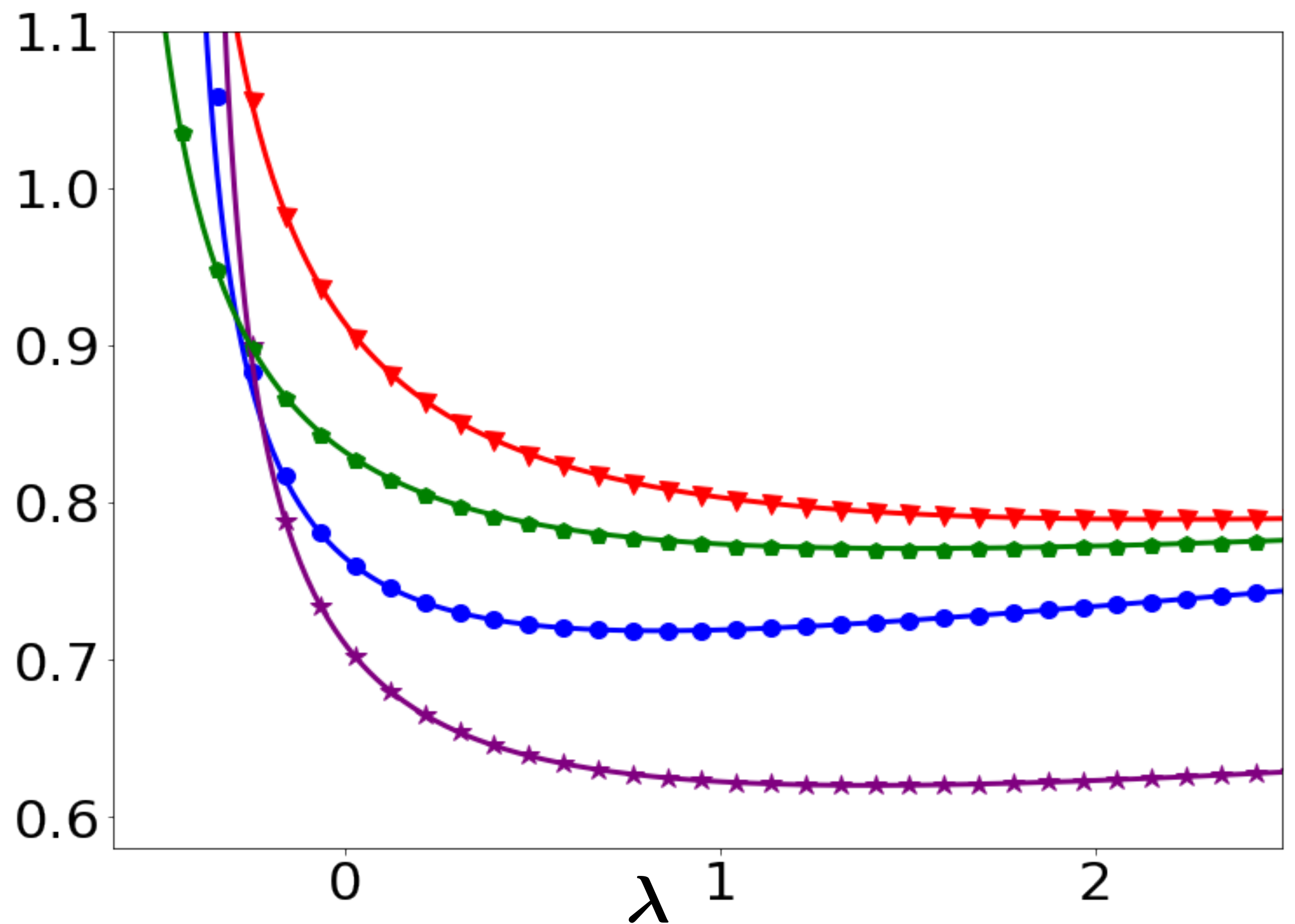}}
\subcaptionbox{Random, SNR $\xi=5$}{\vspace{-0.2cm} \includegraphics[width=0.32\linewidth]{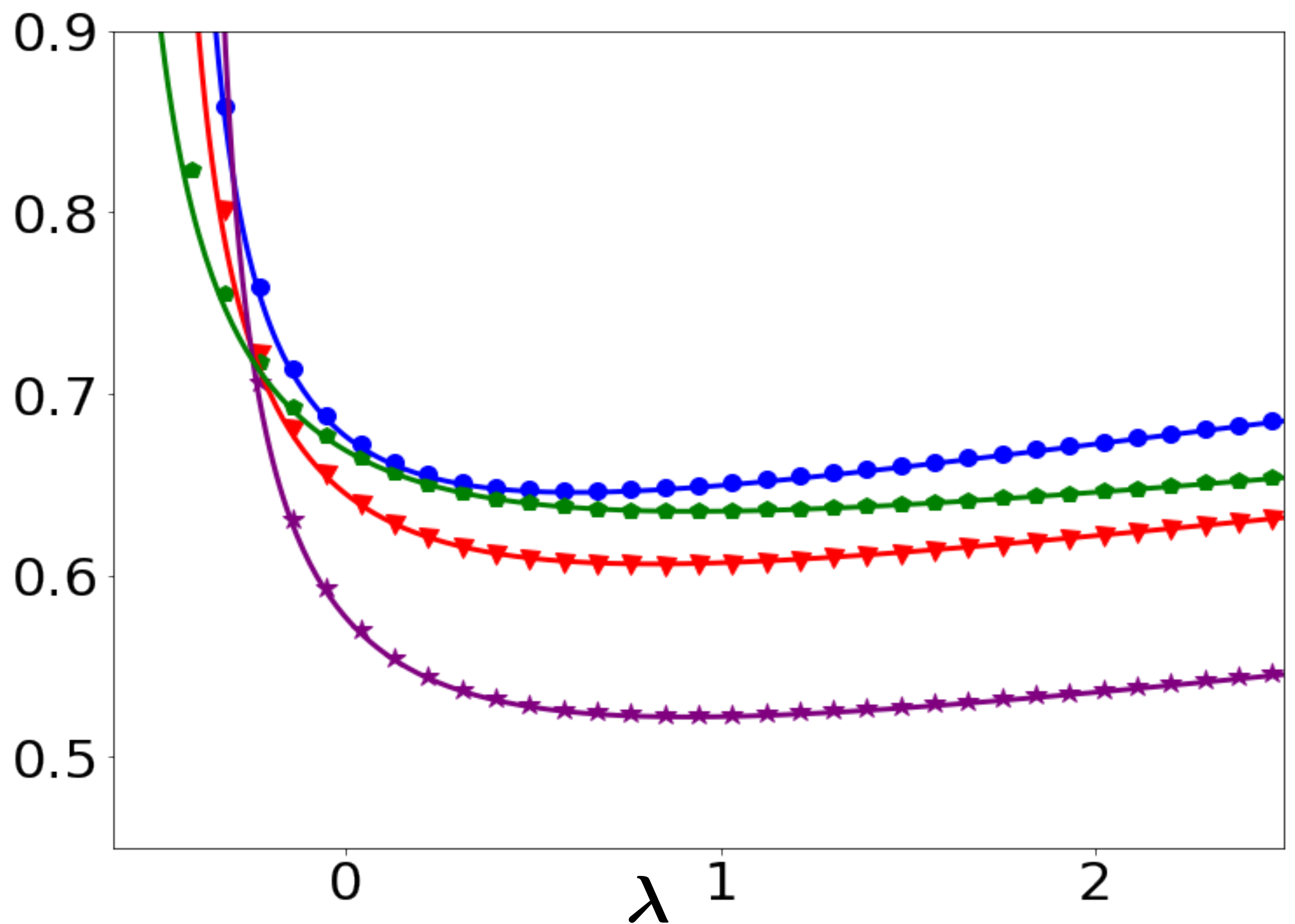}}
\end{minipage}
\vspace{-0.5mm}
\caption{\small Finite sample prediction risk $\tilde{\bbE}(\tilde{y}-\tilde{\vx}^{\t}\vbeta_{\star})^2$ (experiment) and the asymptotic risk $R(\lambda)$ (theory) against $\lambda$ for standard ridge regression ($\vSigma_w = \vI_d$) under label noise with SNR $\xi=5$. We set $\gamma=2$ and $(n,p)=(300, 600)$. `dc' and `ct' stand for for discrete and continuous distribution, respectively. We write `aligned' if $\vd_x$ and $\vd_{\beta}$ have the same order, `misaligned' for the reverse and `random' for random order. Colors indicate different combinations of $\vd_x$ and $\vd_{\beta}$. Note that our derived risk $R(\lambda)$ matches the experimental values for all cases.     
} 
\vspace{1mm}
\label{fig:risk2}
\end{figure}

\begin{figure}[!htb]
\centering
\begin{minipage}[t]{0.4\linewidth}
\centering
{\includegraphics[width=0.96\textwidth]{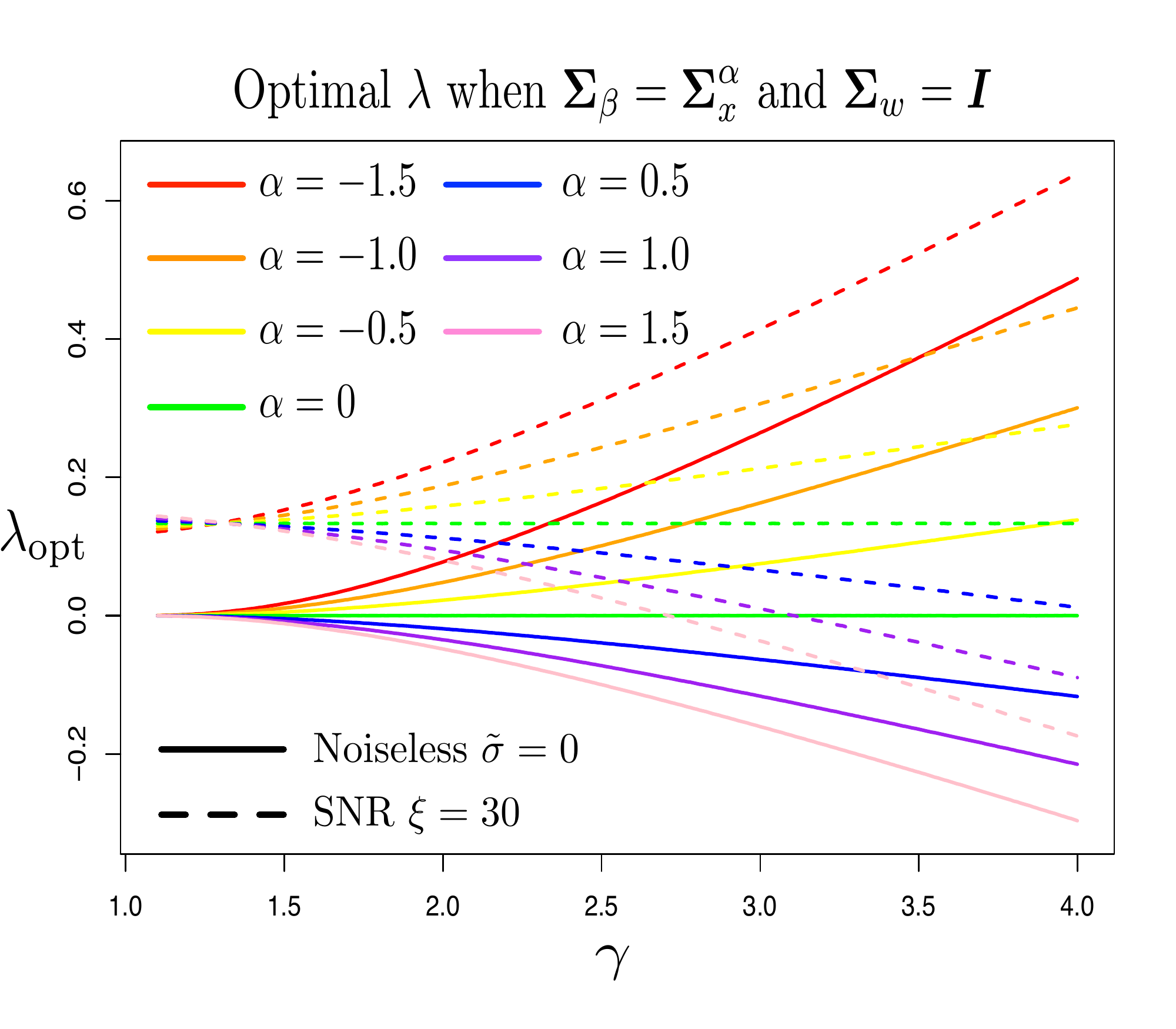}} \\ \vspace{-0.10cm}
(a) $\vd_x\sim \text{unif}([1,3])$.
\end{minipage}
\begin{minipage}[t]{0.4\linewidth}
\centering
{\includegraphics[width=0.96\textwidth]{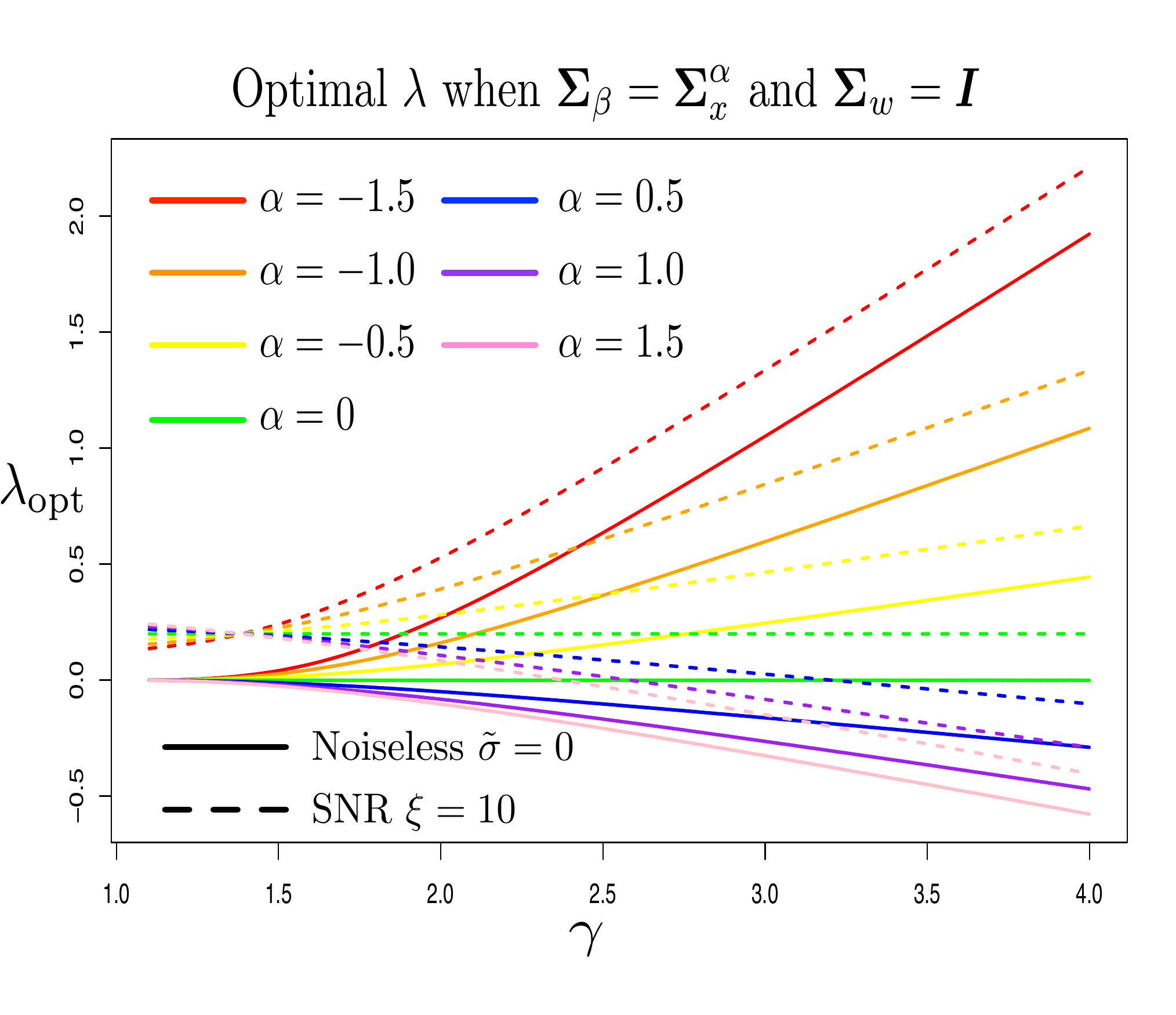}} \\ \vspace{-0.10cm}
(b) $\vd_x\sim \frac{1}{2}\delta_1 +\frac{1}{2}\delta_3$.
\end{minipage}
\vspace{-1mm}
\caption{\small We set $\vSigma_w=\vI$ and $\vSigma_\beta = \vSigma_x^{\alpha}$. As $\gamma$ increases from $1.1$ to $4$, we show the optimal value of $\lambda$ and the solid lines represents the noiseless case $\tilde{\sigma}=0$ and the dashed lines represents the noisy case with a fixed SNR $\xi$. The solid green line shows the level of $0$. We set the distribution of $\vd_x$ to be \textbf{(a)}: uniform on $[1,3]$; \textbf{(b)}:two point masses on $1$ and $3$ with half and half probability.    
}
\label{fig:sign_of_parameter_appendix}
\end{figure}

\end{document}